%% file: main.tex
\documentclass[twoside]{article}

\usepackage[accepted]{aistats2022}
\usepackage[round]{natbib}

\usepackage{microtype}

\usepackage[dvipdfm]{graphicx}
\usepackage{subcaption}

\usepackage{booktabs} 

\usepackage[hyperfootnotes=false]{hyperref}

\usepackage{url}       
\usepackage{amsfonts, amssymb, amsmath, natbib}       

\usepackage{multirow}
\usepackage[table,xcdraw]{xcolor}
\usepackage{amsthm, tikz}
\usepackage{float}

\newcommand{\reals}{\mathbb{R}}
\newcommand{\E}{\mathbb{E}}

\newtheorem{theorem}{Theorem}

\newtheorem{lemma}{Lemma}
\newtheorem{corollary}{Corollary}
\newtheorem{definition}{Definition}
\newtheorem{assumption}{Assumption}

\newcommand{\youngsuk}[1]{{}}
\newcommand{\danielle}[1]{{}}
\newcommand{\bernie}[1]{{}}

\newcommand\blfootnote[1]{%
  \begingroup
  \renewcommand\thefootnote{}\footnote{#1}%
  \addtocounter{footnote}{-1}%
  \endgroup
}

\begin{document}

\runningtitle{Learning Quantile Function without Quantile Crossing}
\runningauthor{Park, Maddix, Aubet, Kan, Gasthaus, Wang}

\twocolumn[

\aistatstitle{Learning Quantile Functions without Quantile Crossing 
\\
for Distribution-free  Time Series Forecasting}


\aistatsauthor{
Youngsuk Park\textsuperscript{\dag}
\And  Danielle C. Maddix\textsuperscript{\dag}
\And 
Fran\c{c}ois-Xavier Aubet\textsuperscript{\dag}
}

\aistatsauthor{
Kelvin Kan 
\textsuperscript{\textbf{*}}
\textsuperscript{\ddag}
\And Jan Gasthaus\textsuperscript{\dag}
\And  Yuyang Wang\textsuperscript{\dag}
}
\aistatsaddress{\textsuperscript{\dag}Amazon Research
\\
\And \textsuperscript{\ddag}Emory University} 
] 

\begin{abstract}
\vspace{-.2cm}
Quantile regression is an effective technique to quantify uncertainty, fit challenging underlying distributions, and often provide full probabilistic predictions through joint learnings over multiple quantile levels.
A common drawback of these joint quantile regressions, however, is \textit{quantile crossing}, which violates the desirable monotone property of the conditional quantile function. 
In this work, we propose 
the Incremental (Spline) Quantile Functions I(S)QF, 
a flexible and efficient distribution-free quantile estimation framework that resolves quantile crossing with a simple neural network layer. Moreover, I(S)QF inter/extrapolate to predict arbitrary quantile levels that differ from the underlying training ones.   
Equipped with the analytical evaluation of the continuous ranked probability score of I(S)QF representations, we apply our methods to NN-based time series forecasting cases, where
the savings of the expensive re-training costs for non-trained quantile levels is particularly significant.
We also provide a generalization error analysis of our proposed approaches under the sequence-to-sequence setting. Lastly, extensive experiments demonstrate the improvement of consistency and accuracy errors over other baselines.

\end{abstract}

\vspace{-.5cm}
\section{
INTRODUCTION}
\label{sec:intr}

Probabilistic time series forecasting methods are increasingly replacing point prediction techniques in practical applications, as it is crucial for downstream decision-making processes to take the uncertainties in predictions into consideration \citep{park2019linear,park2020structured, kim2020optimal,petropoulos2022forecasting}. A popular approach to probabilistic forecasting is to combine a sequential model with a likelihood model that determines how to emit from the hidden or latent states to the observations. Examples include classical statistical models such as State-Space Models \citep{hyndman2008forecasting}, Gaussian processes \citep{rasmussen2006}, and more recently the neural network models like DeepAR / DeepVAR~\citep{salinas2020deepar,salinas2019high}, DeepState~\citep{de2020normalizing,rangapuram2018deep}, DeepFactor~\citep{wang2019deep}, etc. While full probabilistic predictions can be obtained from these models, a practical conundrum of which likelihood function to choose arises. As an example, DeepAR, a RNN-based probabilistic forecaster, offers likelihood choices of normal, student-$t$, negative binomial, etc.  Therefore, it is desirable in both theory and practice to have a method that requires no assumption of the data generating process.  
\blfootnote{\hspace{-.2cm}\textsuperscript{\textbf{*}}Work done as an intern at Amazon Research.}

Fortunately, quantile regression \citep{Koenker1978regression, koenker_2005}, which has been successfully used for robustly modeling probabilistic outputs, comes to rescue. The incorporation of the quantile regression component to various sequential neural network backbones has been shown to be particularly effective with recent advances in deep learning \citep{wen2017multi,gasthaus19sqf,  lim2019temporal, eisenach2020mqtransformer}. Obtaining a full probabilistic prediction (i.e., the ability to query a forecast at an arbitrary quantile) usually requires generating multiple quantiles at once.  
However, when modeling multiple quantiles simultaneously, \emph{quantile crossing}, i.e., the failure of the estimated conditional quantile function to obey the required monotonicity constraint, is a commonly observed concern beyond mere theoretical consistency. This phenomenon particularly stands out in the case of sequence-to-sequence (Seq2Seq) predictions, where the inconsistency can propagate over a sequence of predictions. Even though various strategies have been proposed to remedy quantile crossing (See Section~\ref{sec:related_works}), these techniques are not widely adopted in combination with recent deep probabilistic forecasting models, due to their perceived complexity and lack of solid principles. In particular, \citet{wen2017multi, lim2019temporal} do not explicitly address quantile crossing.

In addition to quantile crossing, another issue of typical quantile regression approaches in the time series context is that they require the set of predicted quantile levels to be fixed for training apriori. 
This restriction requires the expense of re-training the models when making a prediction at arbitrary quantile levels. This is a particular concern with the computational overhead for the class of heavy deep forecasting models with numerous training panels of time series.

In this paper, we propose a simple methodology of Incremental (Spline) Quantile Functions I(S)QF (in Sections~\ref{sec:iqf},~\ref{sec:isqf}), both of which quantify prediction uncertainty in a distribution-free manner. I(S)QF consist of a family of conditional quantile functions that interpolates between the given quantile levels and extrapolates in the parametric tail distribution beyond the extremal training quantiles.  
These inter/extrapolation strategies ultimately lead to resolving both quantile crossing and re-training cost issues, along with supporting analytical Continuous Ranked Probability Score
(CRPS)~\citep{gneiting2007strictly} evaluation in the training stage. 
We apply our methods to Seq2Seq time series forecasting, where both aforementioned issues are raised most frequently (in Section~\ref{sec:iqf_isqf_for_ts}).  
We analyze the generalization errors under multi-horizon and multi-quantile time series forecasting, characterized with quantities, such as the Rademacher complexity and discrepancy measure for stationarity (in Section~\ref{sec:generalization_error}).   Under the state-of-the-art Seq2Seq MQ-CNN model \citep{wen2017multi}, extensive experiments on real-world datasets demonstrate the consistency and accuracy improvement of our methodology with I(S)QF layers over various other baseline layers, e.g., default quantile \citep{wen2017multi}, Gaussian \citep{flunkert2017deepar}, and SQF \citep{gasthaus19sqf} (in Section~\ref{sec:experiment}). 

\vspace{-.3cm}
\subsection{Related Works}\label{sec:related_works}

\paragraph{Seq2Seq Quantile Forecasters.}
Recently, several Seq2Seq forecasting models have been proposed to jointly learn the quantile estimates over multiple quantiles without any explicit distribution assumptions.  
 \citet{wen2017multi} propose MQ-R(C)NN, which uses a RNN or dilated causal convolution (CNN) encoder, respectively, and a multilayer perceptron (MLP) decoder that outputs a set of quantile levels for the entire forecast horizon. 
A disadvantage is that the model needs to be re-trained if other quantile levels than the training ones are requested at inference time. 
~\citet{wen2019deep} extend the MQ-CNN model into a marginal quantile model with a generative quantile Gaussian copula.  This Gaussian copula component improves the forecast at the distribution tails, but the quantile crossing drawback that can occur with MQ-CNN still remains.
\citet{chen2019probabilistic} propose DeepTCN, another Seq2Seq model where the encoder is also a dilated causal convolution with residual blocks, and the decoder is simply a MLP with residual connections. Structure-wise, DeepTCN is almost the same as the basic structure of MQ-CNN, i.e., without the local MLP component that aims to model spikes and events. When the quantiles are learned jointly, all of these Seq2Seq models can still suffer from quantile crossing. 

\vspace{-0.2cm}
\paragraph{Quantile Regression.}
There are various approaches to resolve the issue of quantile crossing that occurs when quantile estimations over multiple quantiles are learned jointly. Most of these works are sorting-based post-processing at the end \citep{ chernozhukov2010quantile, kim2021deep} or or expensive constrained optimization \citep{liu2009stepwise}. \citet{schmidt2016quantile} propose learning on the non-negative increment between quantile estimates on pre-determined quantile levels, and then stacking them. In addition, some linear/non-linear interpolation between quantile estimates allows for a form of quantile function in some range of quantile levels. SQF \citep{gasthaus19sqf} is another method that gives a functional form based on the CRPS and its analytical evaluation, and results in no quantile crossing. However, this method is too flexible in choosing knot positions to optimize efficiently. Therefore, interpolation through linear splines leads to general performance degradation, especially on the tail regions.

\vspace{-.2cm}
\paragraph{Theoretical Analysis on Forecasting.}
There are several recent theoretical analyses of time series forecasting with the tools from learning theory. \citet{kuznetsov2015learning} investigate the theoretical analysis on the general scenario of non-stationary and non-mixing stochastic processes in terms of a data-dependent measure of sequential complexity and a discrepancy measure. \citet{zimin2017learning} study the learnability of stochastic processes with respect to the conditional risk, focusing on analyzing scenarios, where the pairwise discrepancy is controllable. \citet{mariet2019foundations} examine theoretical studies on general multivariate sequence-to-sequence settings in terms of discrepancy measure and mixing coefficient of the underlying stochastic process. Albeit the powerful intuitions and mathematical machinery provided, these works focus on general one-step prediction models, lacking more fruitful intuition for our use case of global multi-horizon and multi-quantile deep time series forecasting.

\vspace{-0.2cm}
\section{PRELIMINARIES}
\label{sec:background}
For a random variable $Z \in \reals$, we denote $F_Z(z)$ as its cumulative distribution function (CDF). Then, the $\alpha$-quantile of $Z$ is given as:
\[
q_Z(\alpha):=F^{-1}_Z(\alpha) = \inf\{z \in \reals : \alpha \leq F_Z(z)\},
\]
where $\alpha \in (0,1)$ denotes a quantile level. The quantile function $q(\cdot)$ is also called the percent-point function or inverse cumulative distribution function.
\subsection{Quantile Regression}

Let $(X,Z)\sim F_{(X,Z)}$ for a regression setting. Quantile regression seeks to estimate the $\alpha$-quantile of $Z \in \reals$ conditioned on $X=x$ for some $\alpha \in [0, 1]$, i.e., $q(\alpha \mid x)= F^{-1}_{Z\mid X=x}(\alpha)$. Equivalently, the $\alpha$-quantile is the solution of minimizing the expected quantile loss \citep{Koenker1978regression}:
\begin{equation}
q(\alpha \mid x) =
\underset{q\in \reals}{\mathrm{argmin}}~
\E_{Z\mid X=x}(\rho_\alpha (Z - q) ),
\label{eq:qr_expectation}
\end{equation}
where $\rho_\alpha (u) = u \times (\alpha - \mathbf{1}\{u < 0\})$ with $u=z-q$ denotes the quantile loss of $q\in \reals$ w.r.t. any $z \in \reals$.  
\paragraph{Continuous Ranked Probability Score
(CRPS).}  

The CRPS~\citep{gneiting2007strictly} is a  \textit{proper scoring rule}\footnote{$\E_{Z\sim F}L(F^{-1}, Z) \leq \E_{Z\sim F} L(G^{-1}, Z)$.} that averages the quantile losses over all quantiles, rather than optimizing a single quantile loss.  Formally, given a fixed target $z\in \reals$ and a family of quantile fuctions $\mathcal{Q}$, the CRPS $L:\mathcal{Q}, ~\reals \rightarrow \reals$ is given as:
\begin{align}
    L(q, z)
    = \int_{\alpha=0}^1
    2\rho_\alpha (z - q(\alpha)) d\alpha.
    \label{eq:crps}
\end{align} 
\subsection{Quantile Regression in Time Series Forecasting}
Quantile regression settings can be extended to the common case of multiple horizon and multiple quantile estimations in time series forecasting as follows. 
Suppose we have $m$ related time series data, each of which consists of observation $z_{i,t} \in \mathbb{R}$ with (optional) input covariates $\xi_{i,t} \in \mathbb{R}^d$ at time $t$. 
In $i$-th time series forecasting, given $T$ past target observations $z_{i, 1:T}$ and all (optional) \emph{future} covariates $\xi_{i, 1:T+\tau}$, we wish to make $\tau$ future quantile predictions at time $T$:
\[
\{\hat z_{i,T+1}^\alpha, \ldots, \hat z_{i,T+\tau}^\alpha\}_{\alpha \in \{\alpha\}} = F_\theta(z_{i,1:T}, \xi_{i,1:T + \tau}),
\]
where $\{\alpha\}$ denotes a set of quantile levels with $\alpha \in [0,1]$, and  $F_\theta$ denotes a global\footnote{Forecast models are the same across all time series $i$.} sequence-to-sequence quantile prediction. In short, we can express the $\alpha$-quantile estimate as:
\[
\hat z_{i,t}^\alpha= 
q^t_{\theta}(\alpha
\mid 
x_{i}
),
\]
where $x_{i}=(z_{i,1:T}, \xi_{i,1:T + \tau})$ denotes the input to the quantile function. 

Finally, Empirical Risk Minimization (ERM) with $L$ in \eqref{eq:crps} is formulated to find the best-fit parameters: 
\begin{align}
\underset{\theta}{\mathrm{minimize}}~ 
 \frac{1}{m\tau}\sum_{i=1}^m \sum_{t=T+1}^{T+\tau} L(q^t_\theta(\cdot\mid x_i), z_{i,t}).
 \label{eq:forecast_erm}
\end{align}

\subsection{Quantile Crossing}
When tackling the problem of quantile learning naively, 
inconsistencies that violate the non-decreasing property of quantile estimate on $\alpha$, called \textit{quantile crossing} may arise. In other words, for $\alpha_1>\alpha_2$, there exists $x$ s.t.
\[
    q(\alpha_1\mid x) < q(\alpha_2\mid x),
\]
where $q(\cdot\mid x)$ is the quantile estimate of the target conditioned on input $x$.
The quantile crossing issue occurs in many cases when quantile regression is applied separately over multiple quantiles $\{\alpha\}$, and/or ERM with CRPS in \eqref{eq:forecast_erm} are solved under naive parameterizations on quantile functions.   

\section{INCREMENTAL QUANTILE FUNCTIONS (IQF)}\label{sec:iqf}

Most forecast models output quantile estimates on a fixed finite set $\{\alpha_k\}_{k=1}^K$, called \textit{quantile knots}. In this section, we propose the Incremental Quantile Functions (IQF), a family of conditional quantile functions. IQF estimates targets on any quantile $\alpha\in (0,1)$, and overcomes quantile crossing. We first learn some basis quantile estimates on quantile knots, and then apply interpolation and extrapolation strategies around the knots to make it accessible on any query quantile.

\youngsuk{revisit} For notational simplicity, we omit index $i, t$. Let $h$ be the last hidden variable with $h={h}(x)$ where ${h}$ denotes a function that maps an input $x$ to a hidden variable $h$. We denote the last output layer as $q_\phi$ with parameter $\phi$ into which the last hidden variable $h$ is fed into, i.e., 
$q_\phi(\alpha \mid h):=q_{\theta}(\alpha \mid x)$. We often use $q(\alpha \mid h)$ by omitting parameters $\phi, \theta$ when obvious.

\subsection{Basis Estimates on Quantile Knots}\label{sec:basis}
On the quantile knots $\{\alpha_k\}_{k=1}^K$ with $\alpha_k < \alpha_{k+1}$, IQF first gives basis quantile estimates $\hat q$ by cumulatively adding non-negative increments as the quantile level $\alpha_k$ increases: 
\begin{equation}\label{eq:iqf_knots}
    \hat q_\phi(\alpha_k\mid h)= 
\begin{cases}
    \pi^\textrm{Linear}(h;\phi_1), & k=1,
    \\
    \hat q_\phi(\alpha_{k-1}\mid h) + \pi^\textrm{MLP}(h;\phi_k), & k \ne 1, 
\end{cases}
\end{equation}
where $\pi^\textrm{Linear}$
denotes a linear function, and 
$\pi^\textrm{MLP}$
denotes a multilayer perceptron (MLP) with a non-negative non-linear component, e.g., ReLu, Sigmoid. Here, $\phi= \{\phi_i\}_{i=1}^K$ are the associated parameters to learn.

\begin{figure}[t]
    \centering
        \includegraphics[width=0.6\columnwidth]{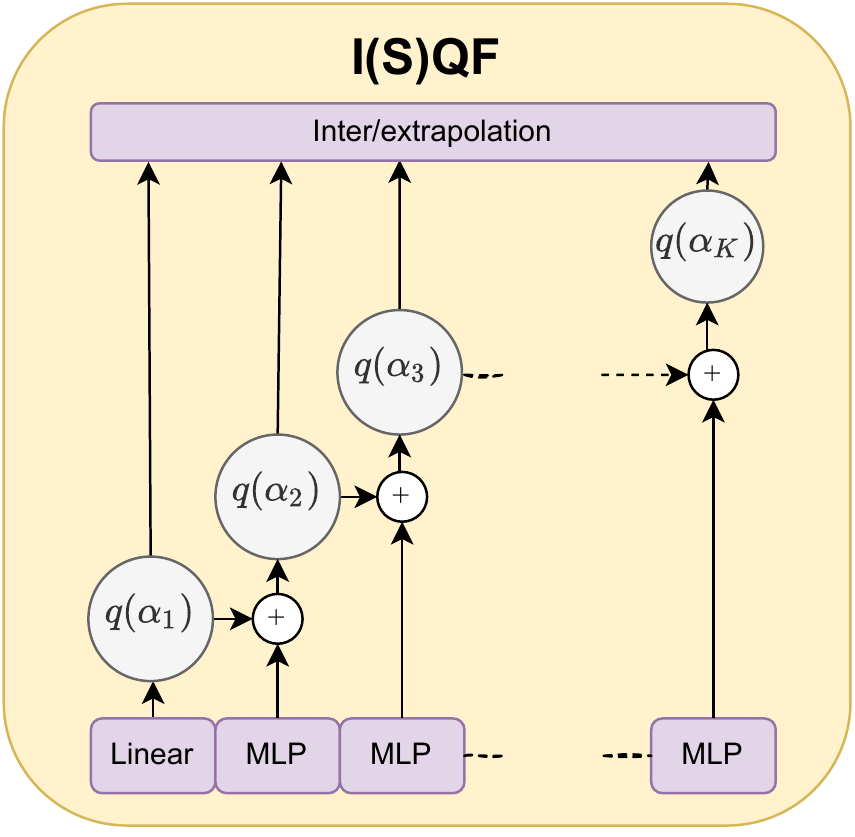}
\caption{Neural network design of I(S)QF representation. I(S)QF architecture never allows quantile crossing to happen to whichever (hidden) input $h$ is given.
\vspace{-0.3cm} 
}
        \label{fig:iqf}
\end{figure}

\subsection{Inter/Extrapolation beyond Knots}\label{sec:inter_extra} 
In order to provide full quantile estimates beyond basis estimates in \eqref{eq:iqf_knots}, i.e., $q(\alpha\mid h)$ on $\alpha \notin \{\alpha_k\}$, IQF takes the inter/extrapolation strategies into account: non-parametric interpolation in the middle and parametric extrapolation in the tails. It adopts a simple yet efficient (deterministic) linear interpolation and extrapolation with exponential tail distributions.

Assume the knot size $K \geq 2$, 
and use extremal knots to define tail regions\footnote{The tail region can be selected more arbitrarily.}, i.e., $\alpha_{\mathrm{tail}_{L}} = \alpha_1$ and  $\alpha_{\mathrm{tail}_{R}}=\alpha_{K}$.

\paragraph{Linear Interpolation.}
For the non-tail region $\alpha \in [\alpha_{\mathrm{tail}_{L}}, \alpha_{\mathrm{tail}_{R}}]$, each linear interpolation on $ \alpha_k \leq \alpha \leq \alpha_{k+1}$ interval gives: 
\begin{align}
    \label{eq:iqf_interpolation}
    \hspace{-.25cm}
    q(\alpha\mid h)
    = 
    w(\alpha)\hat q(\alpha_k\mid h) +  (1-w(\alpha)) \hat q(\alpha_{k+1}\mid h),
\end{align}
where $w(\alpha) = (\alpha_{k+1} - \alpha)/ (\alpha_{k+1} - \alpha_k)$.
Note that interpolation $q$ and basis $\hat q$ match at every knot $\alpha_k$.
 
\paragraph{Extrapolation with Exponential Tails.}
For the tail regions, we extrapolate the quantile estimate through exponential distribution at both ends:
\begin{align*}
    \alpha
    =
\begin{cases}
    \exp(\beta_L  (q(\alpha\mid h) - \gamma_L)),
    & \alpha \leq \alpha_{\mathrm{tail}_{L}},
    \\
    1 - \exp(-\beta_R  (q(\alpha\mid h) - \gamma_R)),
    & \alpha \geq \alpha_{\mathrm{tail}_{R}}.
\end{cases}
\end{align*} 
Here, parameters $(\beta_L, \gamma_L)$ and $(\beta_R, \gamma_R)$ are uniquely chosen so that extrapolation $q$ and basis $\hat q$ coincide on the two leftmost and rightmost quantile knots $\alpha_1$, $\alpha_2$ and $\alpha_{K-1}$, $\alpha_K$, respectively (See the supplementary materials).
 
Then, this is equivalent to extrapolating: 
\begin{align}
    q(\alpha\mid h) 
    =
\begin{cases}
    \frac{1}{\beta_L}
    \log \frac{\alpha}{\alpha_2}
    +
    \hat q(\alpha_{2}\mid h),
    & \hspace{-.2cm} \alpha \leq \alpha_{\mathrm{tail}_{L}},
    \\
    \frac{1}{\beta_R}
    \log \frac{1-\alpha_{K-1}}{1-\alpha}
    +
    \hat q(\alpha_{K-1}\mid h),
    & \hspace{-.2cm} \alpha \geq \alpha_{\mathrm{tail}_{R}}.
\end{cases}
\label{eq:iqf_extrapolation}
\end{align}
We add numeric safeguards to the denominators and log operations that enhance numerical stability while maintaining monotonicity. 
Finally, aggregating inter/extrapolation completes IQF:
\begin{equation*}
q(\alpha \mid h) =
    \begin{cases}
    \eqref{eq:iqf_interpolation},    & \alpha \in [\alpha_{\mathrm{tail}_{L}}, \alpha_{\mathrm{tail}_{R}}],
        \\
    \eqref{eq:iqf_extrapolation},   & \alpha \in (0, \alpha_{\mathrm{tail}_{L}}]\cup [\alpha_{\mathrm{tail}_{R}},1).
    \end{cases}
    \label{eq:iqf}
\end{equation*}
Figure~\ref{fig:iqf} displays the corresponding neural network architecture.

\vspace{-0.2cm}
\paragraph{Distribution Recovery.}

\begin{figure}[h!]
    \centering
    \begin{subfigure}[b]{0.32\textwidth} 
        \includegraphics[width=\columnwidth]{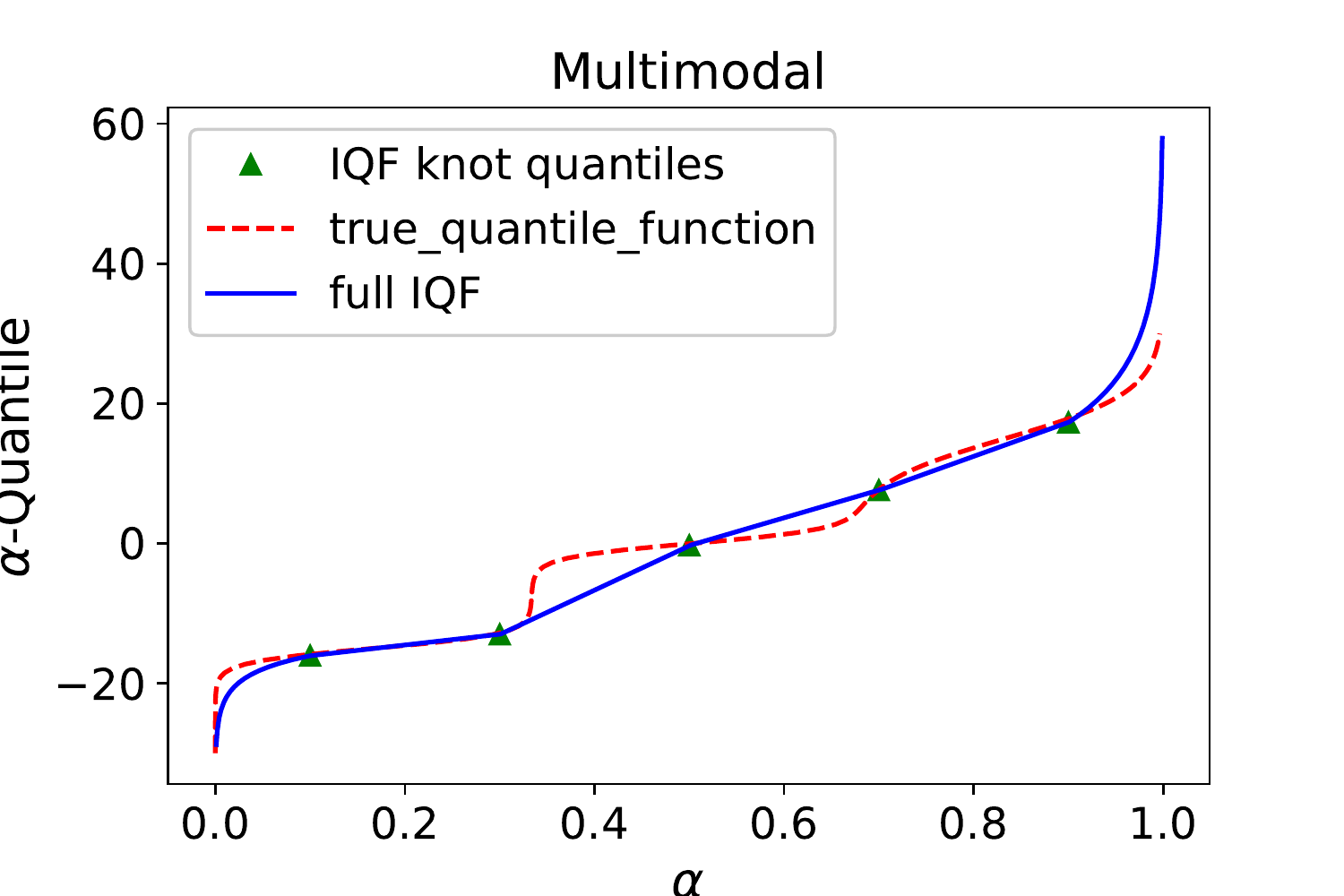}
        \caption{IQF with 5 knots.}
        
        \label{fig:mm_qfs_5}
    \end{subfigure}
    ~ 
    \begin{subfigure}[b]{0.32\textwidth}
        \includegraphics[width=\columnwidth]{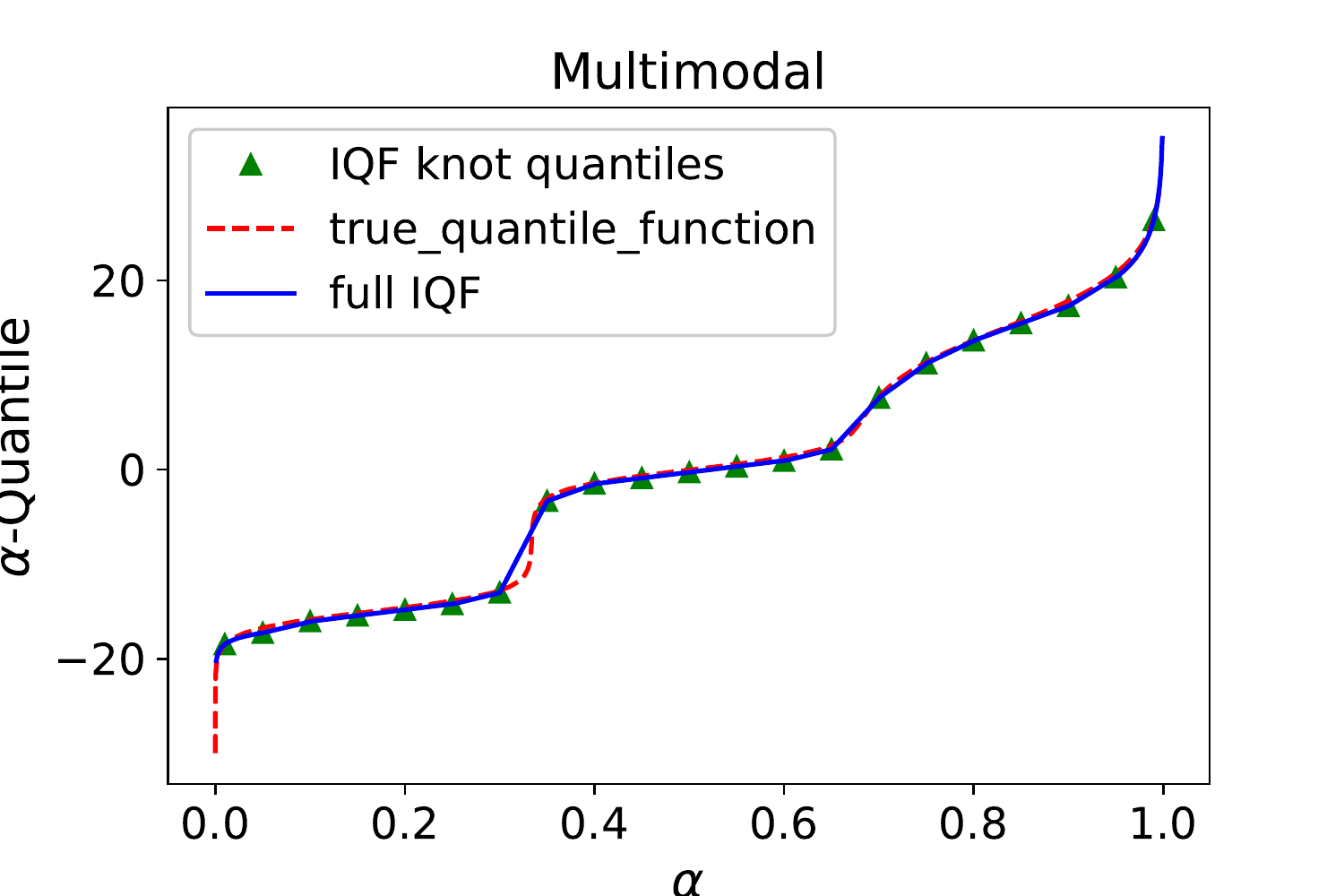}
        \caption{IQF with 20 knots.}
        \label{fig:mm_qfs_20}
    \end{subfigure}
\caption{IQF recovers the underlying multimodal distribution via inter/extrapolation beyond training knots with higher accuracy as the knot size $K$ increases. \vspace{-0.9cm}}
\label{fig:iqf_distribution_free_multi_modal}
\end{figure}

Figure~\ref{fig:iqf_distribution_free_multi_modal} shows the behaviors of IQF for a multi-modal distribution over different number of training knots without any distributional assumption. IQF also fits heavy tail distributions, e.g., Exponential and even Cauchy (although CRPS under true Cauchy distributions may not be well-defined). See the Section~\ref{sec:distribution_recovery} for the additional experiments.
Note that naive distributional assumptions, e.g. Gaussianity fail in these examples. Even though IQF is in a distribution-free class, any additional information regarding the tail distribution for IQF can be enforced based on prior knowledge on the data, e.g., non-negative targets in count data, and the choice of tail regions.

\vspace{-.2cm}
\section{INCREMENTAL SPLINE QUANTILE FUNCTIONS (ISQF)}\label{sec:isqf}
 
We extend IQF into a more flexible family, Incremental Spline Quantile Functions (ISQF). In particular, ISQF adopts learnable inter/extrapolation strategies.

\subsection{Learnable Inter/Extrapolation} 
While IQF takes fixed inter/extrapolation strategies into account, ISQF uses learnable versions of these strategies: linear spline interpolation in the middle and parametric extrapolation with exponential or Generalized Pareto Distribution (GPD) at the tails.
\paragraph{Piecewise-linear Interpolation.}
Given two fixed knot positions $\alpha_k$ and $\alpha_{k+1}$, we additionally adopt learnable linear splines, similar to SQF , to interpolate between them:
\begin{align}
    q(\alpha\mid h)
    &= 
    \sum_{s=0}^{S-1} \max \{ \min \{ \frac{\alpha - d_s}{d_{s+1}-d_s}, 1 \}, 0 \} (p_{s+1}-p_s)
    \nonumber
    \\
     \quad &+ 
    \hat q(\alpha_k\mid h).
    \label{eq:isqf_interpolation}
\end{align}
Here, the $S$ number of spline knots $\{d_s\}$ and associated quantile estimates $\{p_s\}$ are parameterized to be non-decreasing, and match on the basis points $\{\alpha_k\}_{k=1}^K$, i.e.,  $d_0=\alpha_k$, $d_S=\alpha_{k+1}$ and $p_0=\hat q(\alpha_k\mid h)$, $p_S=\hat q(\alpha_{k+1}\mid h)$. \eqref{eq:isqf_interpolation} is equivalent to SQF by a change of variables, and simpler to parameterize and learn. See more discussion in the supplementary materials.  

\paragraph{Extrapolation with Parametric Tails.} 
We generalize the exponential in IQF by requiring it to only pass through the tail quantile knot, and letting the parameters $\beta_L, \beta_R$ in \eqref{eq:iqf_extrapolation} be trainable. 
Another class of extrapolation other than exponential is Generalized Pareto Distribution (GPD). Similar to \cite{ehrlich2021spliced}, we model the tail distribution with GPD as:
\begin{align}
\alpha =
\begin{cases}
\alpha_{\mathrm{tail}_L}
 \left(1 
-
\eta\psi_\mu(\alpha, \alpha_{\mathrm{tail}_L})
\right)^{-1/\eta},
& \hspace{-.1cm} \alpha \leq \alpha_{\mathrm{tail}_L},
\\
\alpha_{\mathrm{tail}_R}(1- 
 \left(1 
+ 
\eta\psi_\mu(\alpha, \alpha_{\mathrm{tail}_R})
\right)^{-1/\eta}),
& \hspace{-.1cm} \alpha \geq \alpha_{\mathrm{tail}_R},
\end{cases}
\label{eq:isqf_extrapolation_gpd}
\end{align}
where $\psi_\mu(\alpha, \alpha') := [q(\alpha\mid h) - q(\alpha'\mid h)] / \mu$, 
and the shape and scale $(\eta, \mu)$ are learnable parameters. Note that the quantiles from extrapolation match on the extremal knots. Lastly, re-arranging the CDF \eqref{eq:isqf_extrapolation_gpd} into quantile form gives the desirable extrapolation.

\subsection{Analytical CRPS Evaluation}\label{sec:analytical_crps}
To this end, an analytic representation of the CRPS $L$ is preferable to an integral approximation because it saves computational cost, and is more accurate. We derive the analytical expression of CRPS integral for I(S)QF with both exponential and generalized Pareto tail distribution \citep{friederichs2012forecast}. 
See Section~\ref{sec:analytical_crps_appendix} for our derivations and final expressions. 





\section{FORECASTING WITH I(S)QF}\label{sec:iqf_isqf_for_ts} 


\begin{figure}[t]
    \centering
        \includegraphics[width=0.8\columnwidth]{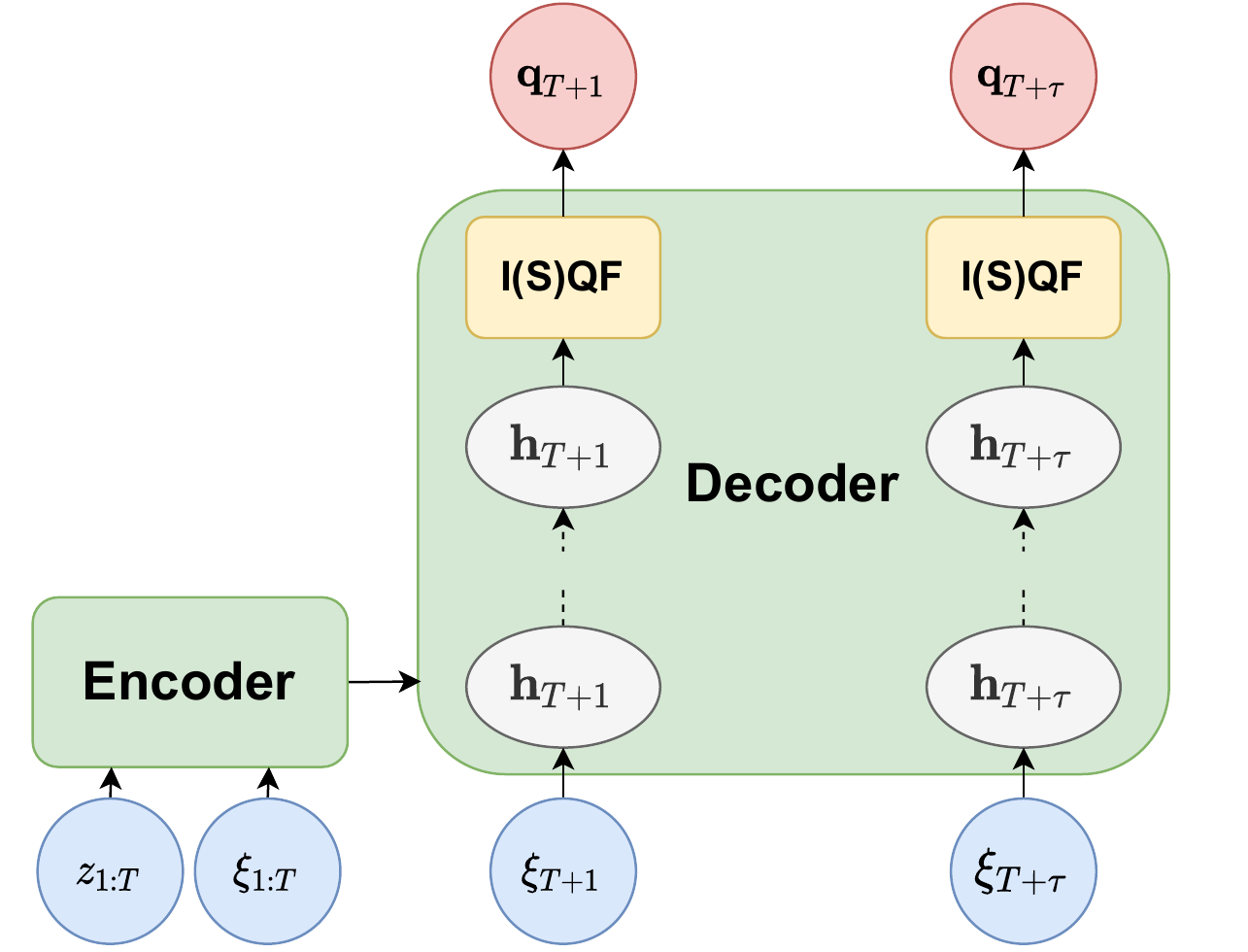}
\caption{An example of applying I(S)QF as the output layers in a Seq2Seq framework.}
        \label{fig:seq2seq}
\vspace{-0.3cm}
\end{figure}



\begin{table*}[t]
\centering
\scalebox{0.77}{
\begin{tabular}{|c|c|c|c|c|c|c|c|}
\hline
 &
  \begin{tabular}[c]{@{}c@{}}Eliminates \\ quantile crossing\end{tabular} &
  \begin{tabular}[c]{@{}c@{}}Eliminates re-training for \\ arbitrary quantile query\end{tabular} &
  \begin{tabular}[c]{@{}c@{}}Supports sample \\ path generation \end{tabular} &
  \begin{tabular}[c]{@{}c@{}}Inter-\\ polation\end{tabular} &
  \begin{tabular}[c]{@{}c@{}}Extra-\\ polation\end{tabular} &
  Knots &
  \begin{tabular}[c]{@{}c@{}}Number of\\ parameters\end{tabular} \\ \hline
QF &
  {
  No} &
  {
  No} &
  {
  No} &
  {
  No} &
  {
  No} &
  {
  Fixed} &
  {
  \textbf{K}} \\ \hline
IQF &
  {
  \textbf{Yes}} &
  {
  \textbf{Yes}} &
  {
  \textbf{Yes}} &
  {
  \textbf{Yes}} &
  {
  \textbf{Yes (Exp)}} &
  {
  Fixed} &
  {
  \textbf{K}} \\ \hline
ISQF &
  {
  \textbf{Yes}} &
  {
  \textbf{Yes}} &
  {
  \textbf{Yes}} &
  {
  \textbf{Yes}} &
  {
  \textbf{Yes (Exp, GPD)}} &
  {
  \textbf{Mixed}} &
  {
  K+2S} \\ \hline
SQF &
  {
  \textbf{Yes}} &
  {
  \textbf{Yes}} &
  {
  \textbf{Yes}} &
  {
  \textbf{Yes}} &
  {
  No} &
  {
  \textbf{Non-fixed}} &
  {
  2(K+S)} \\ \hline
\end{tabular}
}
\caption{Comparison of various quantile output layers. $K$ and $S$ denote the total number of quantile and spline knots for I(S)QF, respectively. In case of SQF, all the knots used for ISQF are treated as its spline knots to fit. \vspace{-.5cm}}
\label{tab:comparison_layers}
\end{table*}

\paragraph{Architecture Design for Seq2Seq Forecasting.}
We now walk through how to use I(S)QF for the conditional quantile function in the Seq2Seq 
time series forecasting setting. Under the general encoder and decoder architecture of  NN-based Seq2Seq models, we replace the last layers of the decoder with I(S)QF, as depicted in Figure~\ref{fig:seq2seq}.
Specifically, 
an input $x$ passes through an encoder and decoder to generate the last hidden variable $h=\mathbf{h}(x)$, which is fed into the I(S)QF layers, resulting in the desirable conditional quantile estimates. 
Finally, we learn the forecast parameters by minimizing the ERM with the CRPS in \eqref{eq:forecast_erm} with some training data. 
I(S)QF is also applicable to autoregressive models (See the supplementary materials for details.)



\paragraph{Sample Path Generation.} 
Based on the full conditional quantile function from I(S)QF, we draw $\alpha \sim \textrm{U}[0,1]$, and then generate a sample path:
\[
[\hat z_{T+1}, \ldots, \hat z_{T+\tau}] =
[
q^{T+1}(\alpha
\mid 
x),
\ldots,
q^{T+\tau}(\alpha
\mid 
x)
],
\]
for any input $x$. 

\paragraph{Comparison of Various Output Layers.}
IQF, ISQF and SQF support a monotonicity-preserving conditional quantile function on arbitrary quantiles and sample path generation. 
On the other hand, the naive MLP-based quantile function (QF) used in MQ-CNN \citep{wen2017multi} is not guaranteed to support these properties.  
In particular, its inability to support sample paths hinders QF from its applicability to autoregressive models, e.g., DeepAR \citep{flunkert2017deepar}.
The additional flexibility in choosing the knots with SQF results in more parameters to optimize over, which can be more difficult in the multi-horizon forecasting.
Table~\ref{tab:comparison_layers} summarizes these differences.

\section{GENERALIZATION ERRORS}\label{sec:generalization_error}
To analyze generalization errors, we first transform the available data $\mathcal{D}=\{z_{i, 1:T}, \xi_{i, 1:T+\tau}\}_{i=1}^m$ into some training data $\mathcal{D}_\mathrm{train}$ and test data $\mathcal{D}_\mathrm{test}$ suitable for defining empirical and generalized losses under Seq2Seq framework.   
First, in the training stage, we assume to forecast at $T-\tau$ with the next $\tau$ targets  $\{z_{i,t}\}_{t=T-\tau+1}^T$ accessible. Formally,  
the \textit{conditional empirical loss} $\hat{\mathcal{L}}(\theta; \mathcal{D}_\mathrm{train})$ under the CRPS $L$ is given as:  
\[
\hat{\mathcal{L}}(\theta; \mathcal{D}_\mathrm{train})
= \frac{1}{m\tau}\sum_{i=1}^m \sum_{t=T-\tau+1}^{T}  L (z_{i,t}, q^t_\theta(\cdot \mid x_i^{-\tau})).
\]
Here,
$\mathcal{D}_\mathrm{train} = \{z_{i,T-\tau+1:T}, x_i^{-\tau}\}_{i=1 }^{m}$ denotes the training data with $\tau$-shifted input $x_i^{-\tau}=(z_{i, 1:T-\tau}, \xi_{i, 1:T})$. 

Next, the \textit{conditional generalized loss}  $\mathcal{L}(\theta\mid \mathcal{D}_\mathrm{test})$ takes the expected loss over the next unknown $\tau$ targets $\{z_{i,t}\}_{t=T+1}^{T+\tau}$ from the actual forecast time $T$ as follows:  
\begin{align*}
&\mathcal{L}(\theta\mid \mathcal{D}_\mathrm{test})
=
\nonumber
\\
&
\frac{1}{m\tau}\sum_{i=1}^m \sum_{t=T+1}^{T+\tau} 
\E \big[L(Z_{i,t}, q^t_\theta(\cdot \mid X_i)) 
\bigm| X_i = x_i
\big],
\end{align*}
where $Z_{i,t}$ denotes a random variable for the future predictions.   
Here, $\mathcal{D}_\mathrm{test}= \{x_i\}_{i=1}^{m}$ denotes the test data
with input $x_i=(z_{i, 1:T}, \xi_{i, 1:T+\tau})$ to the quantile function.  
Then, optimal solution is denoted as:
\begin{align*}
\hat \theta = \mathrm{argmin}~\hat{\mathcal{L}}(\theta; \mathcal{D}_\mathrm{train}), 
~
 \theta^* = \mathrm{argmin}~\mathcal{L}(\theta \mid  \mathcal{D}_\mathrm{test}),
\end{align*}
for conditional empirical loss and generalized loss, respectively.

To begin our analysis, we make the following assumption. 
 
\begin{assumption}\label{assumption:boundedness_target}
The target time series $Z_{i,t}$ are bounded, i.e., $\|Z_{i,t}\|\leq D$ for some $D\geq 0$.
\end{assumption}
 
\subsection{Conditional Generalization Error} 
Let $\mathcal{F}$ be the set of functions $f$ where $f = L \circ q$ is the composition of CRPS loss $L$ in \eqref{eq:forecast_erm} and any quantile function $q$. 
\begin{definition} 
The Rademacher complexity of $\mathcal{F}$ is defined as:
\[
\mathfrak{R}_{N}(\mathcal{F}) = 
\E
\left[ \sup_{f\in \mathcal{F}}\left(\frac{1}{N}\sum_{i=1}^N \sigma_i f(Z_i)\right)\right],
\]
where $\sigma_1,\ldots, \sigma_N$ are independent random variables uniformly chosen from $\{-1, 1\}$. Here the expectation is taken over both $\sigma_i$ and data $Z_i$ for $i=1,\ldots, N$.  
\end{definition}

\begin{lemma}\label{lemma:rademacher}
For a class of quantile functions $\mathcal{Q}$ and CRPS loss $L$, the Rademacher complexity of a composited class $\mathcal{F} = L\circ \mathcal{Q}$ with sample size $N$ is upper bounded by that of $\mathcal{Q}$, i.e.,
\[
\mathfrak{R}_{N}(\mathcal{F}) \leq \mathfrak{R}_{N}(\mathcal{Q}).
\]
\end{lemma}

\begin{definition} 
The temporal discrepancy over $\tau$ time difference is defined as:
\[
\Delta_\mathrm{dis}^\tau (\mathcal{D}_\mathrm{test})
=
\sup_\theta[
\mathcal{L}(\theta\mid \mathcal{D}_\mathrm{test}) 
- 
\mathcal{L}^\mathrm{}(\theta\mid {\mathcal{D}}^{-\tau}_\mathrm{test})
].
\]
Here, 
${\mathcal{D}}^{-\tau}_\mathrm{test}= \{x_i^{-\tau}\}_{i=1}^{m}$
denotes the backtest data
with $\tau$-shifted input $x_i^{-\tau}=(z_{i, 1:T-\tau}, \xi_{i, 1:T})$.

\end{definition}

\begin{theorem}\label{thm:con_gen_error} 
For any $\delta > 0$, a conditional generalization error
\begin{align*}
   \mathcal{L}&(\hat \theta\mid \mathcal{D}_\mathrm{test})- \mathcal{L}(\theta^*\mid \mathcal{D}_\mathrm{test})
\leq 
2 \Delta_\mathrm{dis}^\tau(\mathcal{D}_\mathrm{test}) 
\\
&
+ 
2 \mathfrak{R}_{m\tau}(\mathcal{Q}\mid \mathcal{D}^{-\tau}_\mathrm{test}) 
+ \frac{4D}{\sqrt{m \tau}}\sqrt{\log\left(\frac{1}{\delta}\right)},
\end{align*}
holds with at least  $1-\delta$ probability.
Here,  
$\mathfrak{R}_{m\tau}(\mathcal{Q} \mid  \mathcal{D}^{-\tau}_\mathrm{test})$ denotes the Rademacher complexity of $\mathcal{Q}$, where the expectation on target data is conditioned on the backtest data $\mathcal{D}^{-\tau}_\mathrm{test}$.
\end{theorem}
Theorem~\ref{thm:con_gen_error} says that the error consists of three terms: First, the discrepancy $\Delta^\tau$ term captures errors from non-stationarity, which becomes zero if the data is stationary or periodic, and possibly nonzero otherwise. Note that this stationarity is not required in particular as a data generating process since the corresponding error would appear in the generalization errors. 
Second, Rademacher complexity $\mathfrak{R}_{m\tau}$ gets smaller under larger sample sizes or richer quantile function classes $\mathcal{Q}$. The last one is a high probability term but only with the logarithmic dependency. All of the terms except the discrepancy diminish to zero as the sample size $m\tau$ and knot size $K$ increase.

\textbf{Remark.} Even though a seemingly similar analysis exists \citep{kuznetsov2015learning, zimin2017learning, mariet2019foundations}, our theoretical results are arguably more sophisticated and informative, specifically targeted for the global Seq2Seq quantile forecasting setting. Note that the generalization errors are directly expressed with the complexity of quantile functionals $\mathcal{Q}$ by eliminating the CRPS loss dependency (in Lemma~\ref{lemma:rademacher}) and potential quantization errors through the analytical CRPS integral supported (in Section~\ref{sec:analytical_crps}). Moreover, we analyze the multi-horizon $\tau>1$ cases with exogenous variables $\xi_{i,t}$, applying $\tau$-shifted backtest, adopting $\tau$-shifted discrepancy with sophisticated telescoping and etc. On the other hand, existing works mainly depend on the complexity of both $\mathcal{F}$, analyze one-step prediction without exogenous variables, and can involve quantization errors.


\subsection{Unconditional Generalization Error}
Recall input time series $x_i$ to the quantile function consists of the $i$th observed target and input covariates. We assume that all the targets $Z_{i, 1:T+\tau}$ and the input $X_i$ are independent time series over $i$.
The unconditional generalization loss $\mathcal{L}$ is defined as:
\begin{align*}
\mathcal{L}(\theta)
= \frac{1}{\tau} \sum_{t=T+1}^{T+\tau} 
\E \big[L(Z_{i,t}, q^t_\theta(\cdot \mid X_i)) 
\big].
\end{align*}


The following Lemma explicitly connects the discrepancy quantity to stationarity and beyond. 
\begin{lemma}
Assume each input time series is stationary or periodic. Then the expected temporal discrepancy
$\Delta_\mathrm{dis}^\tau  = \sup_\theta\E[
\mathcal{L}(\theta\mid \mathcal{D}_\mathrm{test}) 
- 
\mathcal{L}^\mathrm{}(\theta\mid {\mathcal{D}}^{-\tau}_\mathrm{test})] = 0$. 
\end{lemma}
Finally, we have a generalization error below.
\begin{corollary}
For any $\delta >0$, a generalization error
\begin{align*}
\mathcal{L}(\hat \theta ) - \mathcal{L}(\theta^*) &\leq 
2 \Delta_\mathrm{dis}^\tau 
+ 
2 \mathfrak{R}_{m\tau}(\mathcal{Q})
+ \frac{2D}{\sqrt{m \tau}}\sqrt{\log\left(\frac{1}{\delta}\right)},  
\end{align*}
holds with at least  $1-\delta$ probability, where
$\Delta_\mathrm{dis}^\tau $
denotes the expected temporal discrepancy.
\end{corollary}



\section{EXPERIMENTS}\label{sec:experiment}


\begin{figure*}[t]
    \centering
    \begin{subfigure}[b]{0.32\textwidth}
        \includegraphics[width=\columnwidth]{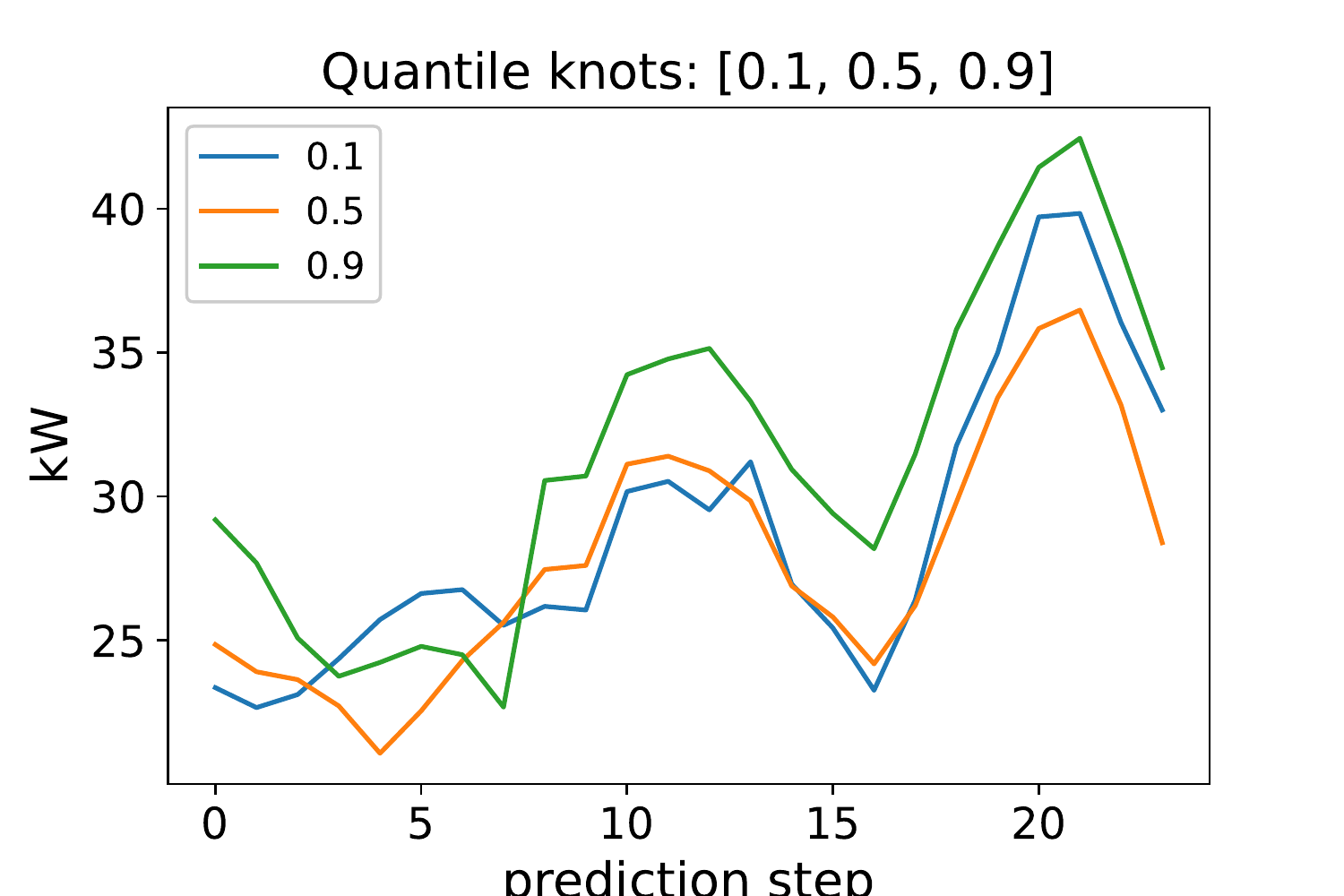}
        \caption{QF with 3 training knots.}
        \label{fig:elec_qf_3}
    \end{subfigure}
    ~ 
    \begin{subfigure}[b]{0.32\textwidth}
        \includegraphics[width=\columnwidth]{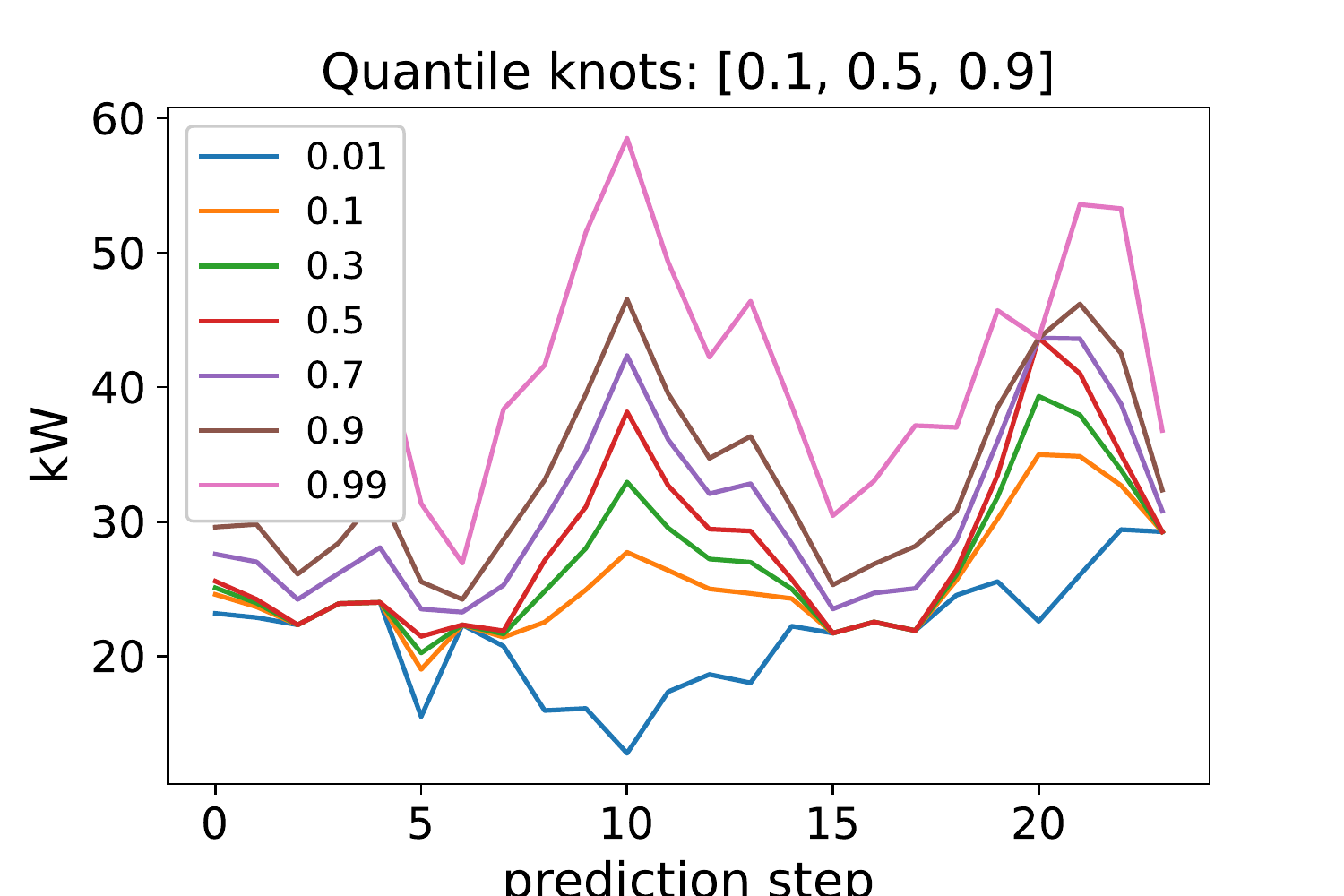}
        \caption{IQF with 3 training knots.}
        \label{fig:elec_iqf_3}
    \end{subfigure}
    ~ 
    \begin{subfigure}[b]{0.32\textwidth}
        \includegraphics[width=\columnwidth]{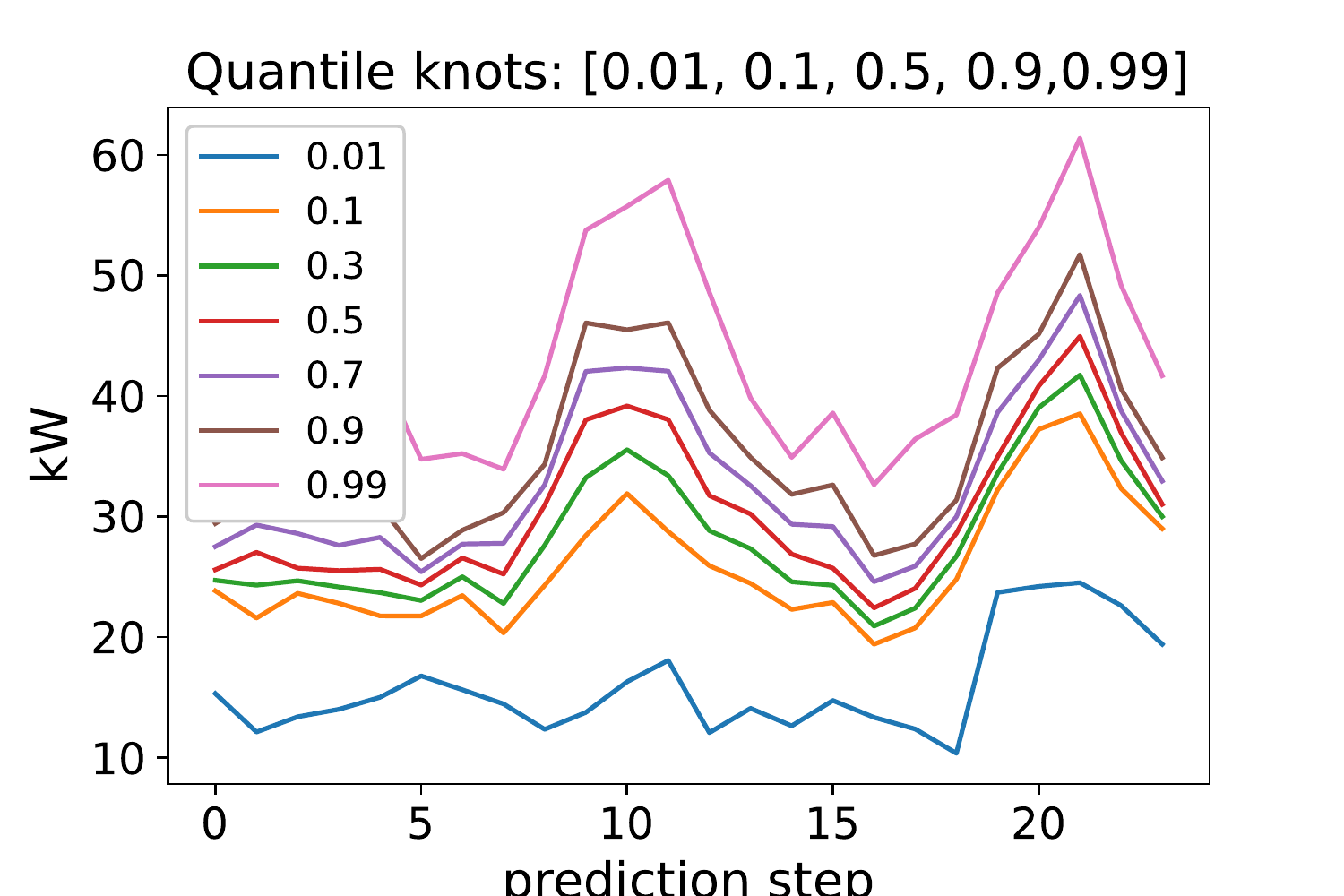}
        \caption{IQF with 5 training knots.}
        \label{fig:elec_iqf_5}
    \end{subfigure}
\caption{Experimental results for the \texttt{Elec} dataset under the MQ-CNN forecaster with $\tau = 24$ predicted time steps, and different quantile output layers and training quantile knots. The default quantile function (QF) is limited to query only on training quantiles, together with severe quantile crossing. On the other hand, IQF shows more consistent predictions with no quantile crossing over any quantile query.
}
\label{fig:quantile_crossing}
\end{figure*}

\input{accuracy_table}

In the experiments, we show the impact of I(S)QF on the state-of-the-art Seq2Seq forecasting model, MQ-CNN  \citep{wen2017multi}. We examine and compare several properties of I(S)QF including no quantile crossing, dynamic quantile query, and accuracy improvements. We implement the model in the open-source GluonTS\footnote{available in \url{https://github.com/awslabs/gluon-ts/blob/master/src/gluonts/model/seq2seq/.}} \citep{gluonts_jmlr} library. 
The experiments are done using AWS SageMaker~\citep{liberty2020elastic}. 

\subsection{Experimental Setup}

We conduct benchmark experiments on the \texttt{Elec} and \texttt{Traf} from the UCI data repository \citep{Dua:2017}, \texttt{Wiki} from Kaggle \citep{lai_dataset_2017}, and 6 different \texttt{M4} competition datasets \citep{makridakisM4concl}. We report the weighted quantile losses to measure the accuracy at various trained, interpolated and extrapolated quantile knots, the mean weighted quantile loss to approximate the CRPS, the mean scaled interval score (MSIS) \citep{gneiting2007strictly} to measure the sharpness and coverage of the distribution, and the crossing error to measure the severity of quantile crossing for our I(S)QF and various baselines: MLP-based QF, SQF, and Gaussian output layers. See the supplementary materials for more details on the setup and hyperparameters.

\vspace{-.2cm}
\subsection{Experimental Results}

\paragraph{Removal of Quantile Crossing Errors.}
The crossing errors in Table~\ref{tab:mq_cnn_iqf} show that QF suffers from quantile crossing, often severely with $42\%$ occurrence and  $0.5$ of averaged crossed distance on $[0,1]$ target domain on \texttt{Traf} dataset. On the other hand, all the other layers including our I(S)QF do not suffer from any quantile crossing.   
Figure~\ref{fig:quantile_crossing} visualizes the quantile crossing for the QF layers, and no quantile crossing for the IQF layers on the \texttt{Elec} data.

\paragraph{Supporting any Quantile Query and Saving Re-training Costs.} The original MQ-CNN model equipped with default QF can make inferences only on pre-determined quantile knots, resulting in N/A for other arbitary quantile queries shown in Table~\ref{tab:mq_cnn_iqf}. Therefore, inferences on different quantiles require an expensive re-training. I(S)QF shown in Figures~\ref{fig:elec_iqf_3}-\ref{fig:elec_iqf_5} and Table~\ref{tab:mq_cnn_iqf} provides the full conditional quantile function queryable on any quantile, only in seconds through inter/extrapolation. This results in re-training costs savings of approximately 10-60 minutes for each run depending on the dataset for the default QF layers.

\paragraph{Accuracy Improvements over Baselines.}
Table~\ref{tab:mq_cnn_iqf} shows that I(S)QF, especially IQF, do not compromise accuracy against QF, and often improve accuracy over other baselines. First, I(S)QF show often notable improvement compared to SQF.  IQF with 5 basis knots performed far better than SQF up to 10 times and 20 times in terms of meanQuantile and MSIS, respectively on most of the datasets. Clearly, SQF has difficulty in fitting the tail due to the lack of extrapolation based on the fact that the wQL[0.995] 10 of SQF in Tables 2 is often much larger than that for our I(S)QF.
This demonstrates that too many flexible knots are difficult to fit, and a moderate amount of fixed knot positions are helpful in practice. Second, the performance of the Gaussian output varies. It performs poorly on \texttt{Elec}, \texttt{Wiki}, and \texttt{M4-monthly}, and well on \texttt{Traf}, arguably due to whether the Gaussianity assumption matches the underlying distribution. Third, IQF performs similarly to QF on 4-6 out of 9 datasets, depending on the metric. For the remaining datasets, IQF generally outperforms QF on the mean weighted quantile loss
, but underperforms on the MSIS. 
We argue that QF sacrifices consistency, and instead gets tighter coverage. Note that some MSIS is not accessible for QF unless it is trained on the corresponding quantile.   
Fourth, ISQF outperforms SQF and Gaussian in general. Compared to QF/IQF, ISQF improves the performance on \texttt{Traf} and \texttt{Wiki} by large margins. It performs similarly to IQF on 4 out of 9 datasets, and worse especially on the remaining 3 datasets. Lastly, the wQL[0.995] metric shows the benefit of the extrapolation methods in I(S)QF.  See Section~\ref{sec:add_experiments} for the extensive results. 

\paragraph{Effects of Training Knots and Other Parameters.}
Through experiments under different number of training knots reported in Section~\ref{sec:add_experiments},
we first observe that the improvement gain for IQF becomes more significant under a larger number of quantile knots.
This result implies that a larger number of training knots tends to improve the accuracy since the CRPS of IQF better matches on that of the true underlying distribution.
In addition, as long as the extremal knots are beyond $0.1$ and $0.9$, I(S)QF behaves reasonably but the most efficient choices may depend on each dataset. Table~\ref{tab:all_benchmark_results} in Section~\ref{sec:all_benchmark_results} demonstrates these results, where 3 vs. 5 training knots are compared. 

Second, IQF and QF have the same number of parameters to fit, but ISQF has more parameters due to the additional spline knots. Therefore, ISQF generally requires more hyperparameter tuning, 
 e.g., different number of epochs and flexible spline knots.  
The performance of ISQF improves with more epochs (200 vs. 100), and in general we observe a (one sided part of) U-shaped curve  depending on the level of difficulty of fitting on each dataset as the number of knots increases. See additional discussion and figures in  Section~\ref{sec:all_benchmark_results}.

\vspace{-.15cm}
\section{DISCUSSION}
\vspace{-.15cm}
In this paper, we propose a distribution-free methodology that infers quantile estimates without quantile crossing. Our approach, the Incremental (Spline) Quantile Functions I(S)QF, is capable of inter/extrapolating to form a conditional quantile function that is accessible at any quantile query, and saves re-training costs accordingly. 
We apply our method to Seq2Seq time series forecasting, analyze generalization bounds, and validate the superiority of our methods through empirical studies. 
While Seq2Seq time series forecasting is mainly covered, our contributions open up our framework to new potential applications including autoregressive probabilistic time series forecasting or other applications like domain adaptation \citep{jin2022domain} where multiple quantile regressions are widely-used, providing additional benefits to future research. We leave for the future work the development of a variant of I(S)QF for multi-quantile prediction on multivariate time series alongside its theoretical analysis and suitable faster training schemes \citep{pmlr-v139-lu21d}.


\subsubsection*{Acknowledgements}
The authors would like to thank Syama Sundar Rangapuram, Lorenzo Stella, Hilaf Hasson, and Anoop Deoras in Amazon research for providing their fruitful feedback at the early stage of this work.


\bibliography{ts}

\newpage
\onecolumn
\appendix
\aistatstitle{
Supplementary Material for
``Learning Quantile Functions without Quantile Crossing 
for Distribution-free  Time Series Forecasting" }

\input{appendix}

\end{document}

%% file: accuracy_table.tex
\begin{table*}[!ht]
\centering
\scalebox{0.68}{
\begin{tabular}{
|c|c|c|c|c|c|c|c|c|c|
 }
\hline
Dataset &
  Strategy &
  mean\_wQL &
  crossing \% &
  wQL{[}0.5{]} &
  wQL{[}0.7{]} (I) &
  wQL{[}0.9{]} &
  wQL{[}0.995{]} (E) &
  MSIS{[}0.1{]} &
  MSIS{[}0.02{]} \\
  \hline
&
Gaussian &
0.129 ±0.03 &
\textbf{0.0} ±0.0 &
0.234 ±0.05 &
0.224 ±0.04 &
0.159 ±0.04 &
0.068 ±0.04 &
42.8 ±12.0 &
79.8 ±22.9
\\ \cline{2-10} 
&
SQF &
0.379 ±0.18 &
\textbf{0.0} ±0.0 &
0.506 ±0.2 &
0.556 ±0.25 &
0.542 ±0.28 &
0.457 ±0.27 &
157.8 ±84.1 &
369.1 ±204.1
\\ \cline{2-10} 
&
QF &
\textbf{0.047} ±0.0 &
1.271 ±0.21 &
\textbf{0.104} ±0.0 &
N/A &
\textbf{0.054} ±0.0 &
N/A &
N/A &
\textbf{17.3} ±0.2
\\ \cline{2-10} 
&
IQF &
0.055 ±0.01 &
\textbf{0.0} ±0.0 &
0.12 ±0.01 &
\textbf{0.101} ±0.01 &
0.056 ±0.0 &
\textbf{0.009} ±0.0 &
\textbf{12.2} ±0.3 &
17.6 ±0.5
\\ \cline{2-10} 
\multirow{-5}{*}{ \texttt{Elec}} &
ISQF &
0.198 ±0.06 &
\textbf{0.0} ±0.0 &
0.378 ±0.06 &
0.37 ±0.08 &
0.253 ±0.11 &
0.069 ±0.05 &
66.9 ±34.8 &
109.1 ±71.3
\\ \hline  \hline 
&
Gaussian &
\textbf{0.154} ±0.01 &
\textbf{0.0} ±0.0 &
\textbf{0.299} ±0.02 &
\textbf{0.3} ±0.02 &
\textbf{0.205} ±0.03 &
0.075 ±0.03 &
\textbf{14.5} ±1.3 &
\textbf{25.1} ±4.0
\\ \cline{2-10} 
&
SQF &
0.193 ±0.01 &
\textbf{0.0} ±0.0 &
0.331 ±0.02 &
0.339 ±0.02 &
0.274 ±0.05 &
0.152 ±0.05 &
22.2 ±2.1 &
45.4 ±4.2
\\ \cline{2-10} 
&
QF &
1.483 ±1.44 &
42.227 ±5.39 &
1.257 ±1.07 &
N/A &
1.035 ±0.76 &
N/A &
N/A &
1128.7 ±1569.1
\\ \cline{2-10} 
&
IQF &
1.31 ±0.74 &
\textbf{0.0} ±0.0 &
2.046 ±1.11 &
1.705 ±1.23 &
1.132 ±1.07 &
0.917 ±1.36 &
166.5 ±150.9 &
373.6 ±354.0
\\ \cline{2-10} 
\multirow{-5}{*}{ \texttt{Traf}} &
ISQF &
0.187 ±0.06 &
\textbf{0.0} ±0.0 &
0.355 ±0.08 &
0.377 ±0.12 &
0.295 ±0.12 &
\textbf{0.043} ±0.01 &
18.8 ±7.5 &
29.4 ±18.1
\\ \hline  \hline 
&
Gaussian &
0.36 ±0.2 &
\textbf{0.0} ±0.0 &
0.575 ±0.23 &
0.634 ±0.33 &
0.58 ±0.38 &
0.38 ±0.32 &
52.2 ±35.3 &
109.1 ±81.8
\\ \cline{2-10} 
&
SQF &
0.176 ±0.01 &
\textbf{0.0} ±0.0 &
0.323 ±0.02 &
0.355 ±0.04 &
0.291 ±0.04 &
0.122 ±0.01 &
21.6 ±2.2 &
40.0 ±3.0
\\ \cline{2-10} 
&
QF &
0.136 ±0.0 &
0.052 ±0.02 &
0.247 ±0.01 &
N/A &
0.214 ±0.0 &
N/A &
N/A &
\textbf{34.5} ±0.3
\\ \cline{2-10} 
&
IQF &
0.135 ±0.0 &
\textbf{0.0} ±0.0 &
0.246 ±0.0 &
0.278 ±0.0 &
0.213 ±0.0 &
0.087 ±0.0 &
\textbf{19.0} ±0.3 &
34.7 ±0.3
\\ \cline{2-10} 
\multirow{-5}{*}{ \texttt{Wiki}} &
ISQF &
\textbf{0.051} ±0.09 &
\textbf{0.0} ±0.0 &
\textbf{0.098} ±0.17 &
\textbf{0.087} ±0.15 &
\textbf{0.064} ±0.11 &
\textbf{0.026} ±0.04 &
21.2 ±0.0 &
41.9 ±0.0
\\ \hline  \hline 
&
Gaussian &
0.017 ±0.0 &
\textbf{0.0} ±0.0 &
0.034 ±0.0 &
0.029 ±0.0 &
0.017 ±0.0 &
0.005 ±0.0 &
43.8 ±6.6 &
74.0 ±15.3
\\ \cline{2-10} 
&
SQF &
0.017 ±0.0 &
\textbf{0.0} ±0.0 &
0.031 ±0.0 &
0.028 ±0.0 &
0.017 ±0.0 &
0.009 ±0.0 &
48.9 ±5.7 &
95.2 ±17.0
\\ \cline{2-10} 
&
QF &
0.015 ±0.0 &
0.329 ±0.27 &
0.029 ±0.0 &
N/A &
0.013 ±0.0 &
N/A &
N/A &
\textbf{52.2} ±1.6
\\ \cline{2-10} 
&
IQF &
0.017 ±0.0 &
\textbf{0.0} ±0.0 &
0.033 ±0.0 &
0.026 ±0.0 &
0.017 ±0.0 &
0.006 ±0.0 &
54.3 ±17.7 &
93.1 ±34.8
\\ \cline{2-10} 
\multirow{-5}{*}{ \texttt{M4-daily}} &
ISQF &
\textbf{0.014} ±0.01 &
\textbf{0.0} ±0.0 &
\textbf{0.025} ±0.01 &
\textbf{0.023} ±0.01 &
\textbf{0.013} ±0.01 &
\textbf{0.003} ±0.0 &
\textbf{42.5} ±0.9 &
154.1 ±59.0
\\ \hline  \hline 
&
Gaussian &
0.043 ±0.0 &
\textbf{0.0} ±0.0 &
0.086 ±0.0 &
0.087 ±0.0 &
0.059 ±0.0 &
0.017 ±0.0 &
\textbf{60.9} ±6.8 &
\textbf{81.3} ±9.0
\\ \cline{2-10} 
&
SQF &
0.046 ±0.0 &
\textbf{0.0} ±0.0 &
0.085 ±0.0 &
0.084 ±0.0 &
0.063 ±0.0 &
0.022 ±0.0 &
64.6 ±4.7 &
96.8 ±18.0
\\ \cline{2-10} 
&
QF &
\textbf{0.038} ±0.0 &
0.048 ±0.0 &
\textbf{0.067} ±0.0 &
N/A &
\textbf{0.057} ±0.0 &
N/A &
N/A &
86.8 ±2.4
\\ \cline{2-10} 
&
IQF &
0.042 ±0.0 &
\textbf{0.0} ±0.0 &
0.069 ±0.0 &
\textbf{0.072} ±0.0 &
0.058 ±0.0 &
0.019 ±0.01 &
70.1 ±4.3 &
103.7 ±8.3
\\ \cline{2-10} 
\multirow{-5}{*}{ \texttt{M4-weekly}} &
ISQF &
0.047 ±0.0 &
\textbf{0.0} ±0.0 &
0.084 ±0.0 &
0.084 ±0.0 &
0.063 ±0.0 &
\textbf{0.012} ±0.0 &
65.8 ±5.3 &
157.6 ±31.2
\\ \hline  \hline 
&
Gaussian &
0.149 ±0.01 &
\textbf{0.0} ±0.0 &
0.188 ±0.01 &
0.179 ±0.01 &
0.153 ±0.01 &
0.115 ±0.01 &
37.0 ±3.0 &
80.4 ±7.2
\\ \cline{2-10} 
&
SQF &
0.12 ±0.01 &
\textbf{0.0} ±0.0 &
0.179 ±0.01 &
0.168 ±0.01 &
0.128 ±0.01 &
0.079 ±0.01 &
25.3 ±2.0 &
50.8 ±5.8
\\ \cline{2-10} 
&
QF &
0.087 ±0.0 &
0.239 ±0.04 &
0.131 ±0.0 &
N/A &
\textbf{0.09} ±0.0 &
N/A &
N/A &
\textbf{31.5} ±0.7
\\ \cline{2-10} 
&
IQF &
\textbf{0.081} ±0.0 &
\textbf{0.0} ±0.0 &
\textbf{0.131} ±0.0 &
\textbf{0.121} ±0.0 &
0.093 ±0.0 &
\textbf{0.038} ±0.02 &
\textbf{19.3} ±1.2 &
33.4 ±5.4
\\ \cline{2-10} 
\multirow{-5}{*}{ \texttt{M4-monthly}} &
ISQF &
0.115 ±0.0 &
\textbf{0.0} ±0.0 &
0.18 ±0.0 &
0.174 ±0.0 &
0.14 ±0.01 &
0.042 ±0.0 &
28.0 ±1.0 &
51.3 ±3.3
\\ \hline  
\end{tabular}
}
\caption{
Comparison of the accuracy for our I(S)QF and various baselines with 5 training quantile knots [0.01, 0.1, 0.5, 0.9, 0.99]. The mean and standard deviation are computed over 4 runs, and the winning method is shown in bold. (I) and (E) indicate the quantiles, where interpolation and extrapolation are performed, respectively.  The MSIS[0.1] measures the 90\% prediction interval using the interpolated $95^{\text{th}}$ and $5^{\text{th}}$ quantiles, if defined. Similarly, the MSIS[0.02] measures the 98\% prediction interval using the $99^{\text{th}}$ and $1^{\text{st}}$ training quantiles. 
\vspace{-.3cm}
}
\label{tab:mq_cnn_iqf}
\end{table*}

%% file: appendix.tex
\section{Extension to Auto-Regressive (AR) Models}
A similar procedure done for Seq2Seq models can be applied to autoregressive models. There is an additional intermediate stage, where the prediction at each time to be fed to the next time-step is sampled from I(S)QF for the next prediction in the sequence. 
In case of sample path generation, we sample $\hat z_{t}=q^t(\alpha_t
\mid 
x)$ by drawing $\alpha_t$ 
uniformly at random at prediction time $t=T+1,\ldots, T+\tau$, recursively.

\section{Extrapolation}

\subsection{Extrapolation with Exponential Tails for IQF}
The fixed parameters $\beta_L, \beta_R$ in the exponential extrapolation for IQF are given as:
\begin{align}
\beta_L&= \frac{\log \left[(\alpha_2+\epsilon)/(\alpha_1+\epsilon)+\epsilon \right]}{\hat  q(\alpha_{2}\mid h) -  \hat q(\alpha_{1}\mid h)}, 
\hspace{.5cm} \beta_R =  \frac{\log \left[(1-\alpha_{K-1}+\epsilon)/(1-\alpha_{K}+\epsilon)+\epsilon \right]} { \hat q(\alpha_{K}\mid h) -  \hat q(\alpha_{K-1}\mid h)}\label{eq:exponential_parameters},
\end{align}
where $\epsilon$ is a tolerance parameter, set to half of machine precision, to prevent numerical errors when dividing by these parameters. Note that this choice also still enforces the desired monotonicity.

\subsection{Extrapolation with Exponential Tails for ISQF}
In IQF, the left and right exponenital tails are required to coincide with the two leftmost and rightmost quantile knots, respectively. In ISQF, this is generalized by restricting the tails to pass through only the leftmost/rightmost quantile knots and letting the free variables to be trainable. In particular, the exponential tails are given as
\begin{align}
    q(\alpha \mid h)
&= \begin{cases}
a_L \log(\alpha) + b_L & =:  q_{\text{tail}_L} (\alpha \mid h),  \quad  0 < \alpha \leq \alpha_{\text{tail}_L} , \\
a_R \log(1-\alpha) + b_R & =:  q_{\text{tail}_R} (\alpha \mid h), \quad  \alpha_{\text{tail}_R} \leq \alpha <1,
\end{cases} \label{eq:tails_formulation}
\end{align}
where $a_L = 1 / \beta_L, a_R = 1 /\beta_R, b_L = -a_L \log({\alpha_{\text{tail}_L}})+\hat q(\alpha_{\text{tail}_L}\mid h), b_R = a_R \log(1-\alpha_{\text{tail}_R}) + \hat q(\alpha_{\text{tail}_R}\mid h)$.

\section{Analytical CRPS for I(S)QF}\label{sec:analytical_crps_appendix}

In this section, we provide the derivation for the analytical CRPS of I(S)QF. For notational simplicity, we denote $q(\alpha):=q(\alpha \mid h)$ and $\hat q(\alpha):=\hat q(\alpha \mid h)$. The CRPS $L$ of I(S)QF is given by
\begin{align}
    L(q, z) &= \int_{0}^1
    2\rho_\alpha (z - q(\alpha)) d\alpha 
    = \left(\int_{0}^{\alpha_{\text{tail}_L}}+\int_{\alpha_{\text{tail}_L}}^{\alpha_{\text{tail}_R}}+\int_{\alpha_{\text{tail}_R}}^{1}\right) 2\rho_\alpha (z - q(\alpha)) d\alpha.
    \label{eq:CRPS}
\end{align}
\subsection{Exponential Tails}
\paragraph{Left Tail CRPS}
The first term of \eqref{eq:CRPS} which corresponds to the left tail can be evaluated as
\begin{equation*}
\begin{aligned}
    \int_{0}^{\alpha_{\text{tail}_L}} 2\rho_\alpha (z - q_{\text{tail}_L}(\alpha)) d\alpha &= \int_{0}^{\alpha_{\text{tail}_L}} 2 (\alpha-\mathbf{1}\{z \leq q_{\text{tail}_L}(\alpha)\})(z-q_{\text{tail}_L}(\alpha)) d\alpha \\
    &= \int_{0}^{\alpha_{\text{tail}_L}} 2 \alpha(z-q_{\text{tail}_L}(\alpha)) d\alpha - \int_{\tilde{\alpha}_{\text{tail}_L}}^{\alpha_{\text{tail}_L}} 2 (z-q_{\text{tail}_L}(\alpha)) d\alpha.
\end{aligned}
\end{equation*}

\newpage
Here, the quantile level $\tilde{\alpha}_{\text{tail}_L}$ denotes the quantile, where the indicator function changes value and is given by
$$
\tilde{\alpha}_{\text{tail}_L}=
\begin{cases}
\exp\big( (z-b_L)/a_L \big), & \text{if } z < \hat q(\alpha_{\text{tail}_L}), \\
\alpha_{\text{tail}_L}, & \text{otherwise}.
\end{cases}
$$
Evaluating the integrals using~\eqref{eq:tails_formulation}, we obtain the following 
\begin{equation}
\begin{aligned}
    \int_{0}^{\alpha_{\text{tail}_L}} 2\rho_\alpha (z - q_{\text{tail}_L}(\alpha)) d\alpha 
    &= (z-b_L)(\alpha_{\text{tail}_L}^2-2\alpha_{\text{tail}_L}+2\tilde{\alpha}_{\text{tail}_L})\\
    \quad &+ a_L \Big[ \alpha_{\text{tail}_L}^2(-\log(\alpha_{\text{tail}_L})+\frac{1}{2}) +2\alpha_{\text{tail}_L}(\log(\alpha_{\text{tail}_L})-1)-2\tilde{\alpha}_{\text{tail}_L}(\log(\tilde{\alpha}_{\text{tail}_L})-1) \Big].
\label{eq:left_crps}
\end{aligned}
\end{equation}
\paragraph{Right Tail CRPS}
By using the symmetry of the two tails, we compute the right tail CRPS using the results of the left tail in~\eqref{eq:left_crps} as 
\begin{align*}
    \int^{1}_{\alpha_{\text{tail}_R}} 2\rho_\alpha (z - q_{\text{tail}_R}(\alpha))  d\alpha  
    &= - \int_{0}^{\alpha_{\text{tail}_L}} 2\rho_{\alpha'} (z - q_{\text{tail}_L}(\alpha'))  d\alpha' \\
    &= -(z-b_R)(1+\alpha_{\text{tail}_R}^2-2\tilde{\alpha}_{\text{tail}_R})
 - a_R \Big[(1-\alpha_{\text{tail}_R}^2) \log(1-\alpha_{\text{tail}_R})\\
 \quad &+ \frac{1}{2}+\frac{\alpha_{\text{tail}_R}^2}{2}+\alpha_{\text{tail}_R}
-2(1-\tilde{\alpha}_{\text{tail}_R})\log(1-\tilde{\alpha}_{\text{tail}_R})-2\tilde{\alpha}_{\text{tail}_R} \Big],
\end{align*}
 with the change of variables: $\alpha'=1-\alpha$, $\alpha_{\text{tail}_R}=1-\alpha_{\text{tail}_L}$, $\beta_R=-\beta_L$, $\hat{q}(\alpha_{\text{tail}_R})=\hat{q}(\alpha_{\text{tail}_L})$, $a_R=a_L$, $b_R=b_L$, $\tilde{\alpha}_{\text{tail}_R}=1-\tilde{\alpha}_{\text{tail}_L}$, and  $q_{\text{tail}_R}(\alpha)=q_{\text{tail}_L}(\alpha')$. 


\subsection{GPD Tails}
The analytical CPRS for the Generalized Pareto (GPD) right tail is given in \cite{friederichs2012forecast} as: 
\danielle{check how to handle $ \alpha_{\text{tail}_R}$ multiplier} 
\[
    \begin{aligned}
        \text{CRPS}_{\text{GPD}_{\text{tail}_R}}(\alpha) &=
 \frac{\mu}{\eta} \big[1+\eta \psi_\mu(\alpha, \alpha_{\text{tail}_R})\big]\big[2\text{GPD}_{\text{tail}_R}(\alpha) -1\big] \\
        &- \frac{2\mu}{\eta(\eta-1)}\bigg[\frac{1}{\eta-2}+\big[1-\text{GPD}_{\text{tail}_R}(\alpha)\big]\big[1+\eta
        \psi_\mu(\alpha, \alpha_{\text{tail}_R}) \big]
        \bigg],
    \end{aligned}
\]
where $\eta \ne 0$, and $\text{GPD}_{\text{tail}_R}(\alpha)$ is the expression in Equation 8 in the main body for $\alpha \ge \alpha_{\text{tail}_R}$ divided by $\alpha_{\text{tail}_R}$. The $\text{CRPS}_{\text{GPD}_{\text{tail}_L}}(\alpha)$ has a similar form with $\alpha_{\text{tail}_L}$.

\subsection{Spline CRPS}
The middle term of \eqref{eq:CRPS} corresponds to the non-tail region (spline), and the integral is given by
\begin{equation}
    \int_{\alpha_k}^{\alpha_{k+1}} 2\rho_\alpha (z - q_{\text{spline}}(\alpha)) d\alpha = \int_{\alpha_k}^{\alpha_{k+1}} 2 \alpha(z-q_\text{spline}(\alpha)) d\alpha - \int_{\tilde{\alpha}}^{\alpha_{k+1}} 2 (z-q_\text{spline}(\alpha)) d\alpha,
    \label{eq:CRPS_spline}
\end{equation}
 where $q_\text{spline}$ denotes a linear spline with $S$ knots defined in $[\alpha_k,\alpha_{k+1}]$ as:
\begin{equation}
    q_\text{spline}(\alpha)
    = 
    \sum_{s=0}^{S-1} \max \{ \min \{ \frac{\alpha - d_s}{d_{s+1}-d_s}, 1 \}, 0 \} (p_{s+1}-p_s) + 
    \hat q(\alpha_k),
    \label{eq:ISQF_spline}
\end{equation}
with $\{d_s\}$ and $\{p_s\}$ spline knots and associated quantile estimates, respectively.
Note that since $p_0=\hat q(\alpha_k)$ and $p_S=\hat q(\alpha_{k+1})$, \eqref{eq:ISQF_spline} reduces to the linear interpolation of IQF when $S=1$. 
Here, the quantile level $\tilde{\alpha}$ is where the indicator function changes value and given by
$$
\tilde{\alpha}=
\begin{cases}
\alpha_k, & \text{if } z \leq \hat q(\alpha_k), \\
d_{s_0}+(z-q(\alpha_k)-\sum_{s=0}^{s_0-1} (p_{s+1}-p_s)) \left(\frac{d_{s_0+1}-d_{s_0}}{p_{s_0+1}-p_{s_0}} \right), & \text{if } \hat q(\alpha_k)<z<\hat q(\alpha_{k+1}), \\
\alpha_{k+1}, & \text{if } \hat q(\alpha_{k+1})\leq z,
\end{cases}
$$
where $s_0=\max \{ s\mid q_\text{spline}(d_{s})<z, 0\leq s \leq S \}$. The quantile level $\tilde{\alpha}$ can be computed in $\mathcal{O}(S)$ time by evaluating the spline at the knots. Since the knots are in increasing order, and $q_\text{spline}$ is a non-decreasing function, the time complexity can be reduced to $\mathcal{O}(\log(S))$ if binary search is used. 

We compute the first term in~\eqref{eq:CRPS_spline} as follows: 
\begin{equation}
    \begin{aligned}
 \int_{\alpha_k}^{\alpha_{k+1}} 2 \alpha(z-q_\text{spline}(\alpha)) d\alpha &= (\alpha_{k+1}^2-\alpha_k^2)(z-\hat q(\alpha_k)) -  2\sum_{s=0}^{S-1} (p_{s+1}-p_s)\int_{\alpha_k}^{\alpha_{k+1}} \max \{ \min \{ \frac{\alpha - d_s}{d_{s+1}-d_s}, 1 \}, 0 \}  d\alpha.
    \end{aligned}
    \label{eqn:spline_int}
\end{equation}

To evaluate the integral in the second term of~\eqref{eqn:spline_int}, we break up the integral at the spline knots, where $\alpha_k \le d_s \le d_{s+1} \le \alpha_{k+1}$:
\[
\begin{aligned}
  \int_{\alpha_k}^{\alpha_{k+1}} 2\alpha \max \{ \min \{ \frac{\alpha - d_s}{d_{s+1}-d_s}, 1 \}, 0 \}  d\alpha &=\int_{\alpha_k}^{d_s} 0 d\alpha +2\int_{d_s}^{d_{s+1}} \alpha \frac{\alpha - d_s}{d_{s+1}-d_s} d\alpha 
  +2\int_{d_{s+1}}^{\alpha_{k+1}}  \alpha d\alpha \\
& = \frac{1}{d_{s+1}-d_s} \bigg(\frac{2d_{s+1}^3}{3}-d_sd_{s+1}^2
    +\frac{d_s^3}{3} \bigg ) +\alpha_{k+1}^2-d_{s+1}^2.
    \end{aligned}
\]
Substituting this into~\eqref{eqn:spline_int} gives:
\begin{equation}
    \int_{\alpha_k}^{\alpha_{k+1}} 2\alpha(z-q_\text{spline}(\alpha)) d\alpha =(\alpha_{k+1}^2-\alpha_k^2)(z-\hat q(\alpha_k)) -\sum_{s=0}^{S-1}(p_{s+1}-p_s)\bigg[\frac{1}{d_{s+1}-d_s} \bigg(\frac{2d_{s+1}^3}{3}-d_sd_{s+1}^2
    +\frac{d_s^3}{3} \bigg ) +\alpha_{k+1}^2-d_{s+1}^2\bigg].
\label{eqn:spline_term1}
\end{equation}

Similarly, we compute the second term in~\eqref{eq:CRPS_spline} as follows: 
\begin{equation}
    \begin{aligned}
 \int_{\tilde \alpha}^{\alpha_{k+1}} 2(z-q_\text{spline}(\alpha)) d\alpha &= 2(\alpha_{k+1}-\tilde \alpha)(z-\hat q(\alpha_k)) -  2\sum_{s=0}^{S-1}(p_{s+1}-p_s) \int_{\tilde \alpha}^{\alpha_{k+1}} \max \{ \min \{ \frac{\alpha - d_s}{d_{s+1}-d_s}, 1 \}, 0 \} d\alpha. 
    \end{aligned}
    \label{eqn:spline_int_2}
\end{equation}
Here, we have the following three cases: 1. $\tilde \alpha \le d_s \le d_{s+1}$, 2. $d_s \le \tilde \alpha \le d_{s+1}$, 3. $d_s \le d_{s+1} \le \tilde \alpha$. Hence, we can succinctly write the integral in the second term of~\eqref{eqn:spline_int_2} as: 
\[
\begin{aligned}
  \int_{\tilde \alpha}^{\alpha_{k+1}} 2\max \{ \min \{ \frac{\alpha - d_s}{d_{s+1}-d_s}, 1 \}, 0 \}  d\alpha &= 2\int_{r_s}^{d_{s+1}} \frac{\alpha - d_s}{d_{s+1}-d_s} d\alpha 
  +2\int_{\max(\tilde{\alpha},d_{s+1})}^{\alpha_{k+1}}  d\alpha \\
& = d_{s+1}^2-2d_sd_{s+1} - r_s^2-2d_sr_s + 2(\alpha_{k+1}-\max(\tilde{\alpha},d_{s+1})),
    \end{aligned}
\]
where $r_s = \max \{ \min \{ \tilde{\alpha}, d_{s+1}\}, d_s \}$.

Substituting this into~\eqref{eqn:spline_int_2} gives:
\begin{equation}
\begin{aligned}
    \int_{\tilde \alpha}^{\alpha_{k+1}} 2(z-q_\text{spline}(\alpha)) d\alpha =2(\alpha_{k+1}-\tilde \alpha)(z-\hat q(\alpha_k)) &-\sum_{s=0}^{S-1}(p_{s+1}-p_s)\bigg[\frac{1}{d_{s+1}-d_s} \bigg(d_{s+1}^2-2d_sd_{s+1} - r_s^2+2d_sr_s \bigg)\\ &+2(\alpha_{k+1}-\max(\tilde{\alpha},d_{s+1}))\bigg].
\label{eqn:spline_term2}
\end{aligned}
\end{equation}

In summary, substituting the integrals in~\eqref{eqn:spline_term1} and~\eqref{eqn:spline_term2} into \eqref{eq:CRPS_spline} gives the following final expression:
\begin{align*}
\int_{\alpha_k}^{\alpha_{k+1}} 2\rho_\alpha (z - q_{\text{spline}}(\alpha)) d\alpha 
    &= (\alpha_{k+1}^2-\alpha_k^2-2(\alpha_{k+1}-\tilde{\alpha}))(z-\hat q(\alpha_k)) \\
\quad &+ \sum_{s=0}^{S-1} (p_{s+1}-p_s)\bigg[\frac{1}{d_{s+1}-d_s}
\bigg( d_{s+1}^2 (-\frac{2}{3} d_{s+1}+d_s+1) - d_s (\frac{d_s^2}{3}+2d_{s+1}) \\
\quad & - r_s (r_s-2d_s)  \bigg)
 -\alpha_{k+1}^2+d_{s+1}^2+2\alpha_{k+1}-2\max(\tilde{\alpha},d_{s+1})\bigg].
\end{align*}

\section{Equivalence of the Splines of ISQF and SQF}
We show that the spline of ISQF is equivalent to SQF. We first recall for $\alpha \in [\alpha_k, \alpha_{k+1}]$, SQF is given by
\begin{equation}
    \text{SQF}(\alpha)
    = \hat q(\alpha_k) + \sum_{s=0}^{S-1}  c_s \max \{ \alpha-d_s, 0\},
    \label{eq:SQF_equation}
\end{equation}
where $\{d_s\}$ denote the $x$-coordinate of the spline knots as in ISQF, and $\{c_s\}$ denote the parameters determining the slopes of the pieces. Since $\{d_s\}$ is in an increasing order, the following holds:
\begin{equation}
    \sum_{s=0}^{S-1} c_s \max \{ \alpha-d_s, 0\} = \sum_{s=0}^{S-1} m_s \max \{ \min \{ \alpha-d_s, d_{s+1}-d_s\}, 0\},
    \label{eq:SQF_cumsubtract}
\end{equation}
where $m_s=\sum_{i=0}^s c_i$ denotes the slope of the $(s+1)$-th piece. Substituting~\eqref{eq:SQF_cumsubtract} into~\eqref{eq:SQF_equation}, we have 
\begin{equation}
    \text{SQF}(\alpha)
    = \hat q(\alpha_k) + \sum_{s=0}^{S-1} m_s \max \{ \min \{ \alpha-d_s, d_{s+1}-d_s\}, 0\}.
    \label{eq:SQF_intermediate}
\end{equation}
Letting $m_s = (p_{s+1}-p_s)/(d_{s+1}-d_s)$ and substituting it into~\eqref{eq:SQF_intermediate}, we obtain
\begin{align*}
    \text{SQF}(\alpha)
    &= \hat q(\alpha_k) + \sum_{s=0}^{S-1} \frac{p_{s+1}-p_s}{d_{s+1}-d_s} \max \{ \min \{ \alpha-d_s, d_{s+1}-d_s\}, 0\} \\
    &= \hat q(\alpha_k) + \sum_{s=0}^{S-1} (p_{s+1}-p_s) \max \{ \min \{ \frac{\alpha-d_s}{d_{s+1}-d_s}, 1\}, 0\},
\end{align*}
which is equivalent to the spline of ISQF~\eqref{eq:ISQF_spline}.

\section{Proofs of Generalization Errors}\label{sec:main_proof}
\subsection{Proof of Lemma 1}

\begin{lemma}\label{lemma:boundedness_L}
Let $z_1, z_2\in \mathcal{Z}_D$ where $\mathcal{Z}_D \subset \mathbb{B}_D:=\{z\mid \|z\|\leq D\}$, then  
\[
|L(z_1, q) - L(z_2, q)| \leq 2D.
\]
\end{lemma}

\begin{proof}
Without loss of generality, let $z_1 \leq z_2$. Then, for a fixed $\alpha$ with $q:=q(\alpha)$,
\begin{align*}
    \rho_\alpha(q-z_1) - \rho_\alpha(q-z_2)
    =
    \begin{cases}
        \alpha(z_2-z_1), & q\geq z_2, \\
        q - z_2 + \alpha(z_2-z_1), & z_2 \leq q \leq z_1, \\
        (1-\alpha) (z_2 - z_1), & q\leq z_1.
    \end{cases}
\end{align*}
Since 
\[
(1-\alpha)(z_1-z_2) \leq q - z_2 + \alpha(z_2-z_1) \leq \alpha(z_2-z_1),
\]
under $z_2 \leq q \leq z_1$, 
\[
|\rho_\alpha(q-z_1) - \rho_\alpha(q-z_2)|\leq \max(\alpha, 1-\alpha) |z_2-z_1| \leq |z_2-z_1|,
\]
holds. 
Thus
\begin{align*}
    \int \rho_\alpha(q(\alpha)-z_1) - \rho_\alpha(q(\alpha)-z_2) d\alpha
    &\leq 
    \int |\rho_\alpha(q(\alpha)-z_1) - \rho_\alpha(q(\alpha)-z_2)| d\alpha
    \\
    &\leq |z_2 - z_1|
    \\
    &\leq 2D.
\end{align*}
\end{proof}

\begin{lemma}\label{lemma:crps_lipschitz}
CRPS $L(q, z)$ is 1-Lipschitz continuous w.r.t any quantile function $q \in \mathcal{Q}$ in $L_1$ norm, i.e., for any $z\in \mathcal{Z}_D$ and $q_1, q_2\in \mathcal{Q}$,
\[
|L(z, q_1) - L(z, q_2)| \leq \|q_1  - q_2 \|_{1}.
\] 
\end{lemma}

\begin{proof}
Similar to the proof of Lemma~\ref{lemma:boundedness_L}, for a fixed $\alpha$ and $z$, 
\[
|\rho_a(q_1(\alpha)-z) - \rho_a(q_2(\alpha)-z)| \leq |q_2(\alpha)-q_1(\alpha)|,
\]
holds. 
Thus
\begin{align*}
    \left | \int \rho_\alpha(q_1(\alpha)-z) - \rho_\alpha(q_2(\alpha)-z) d\alpha \right | 
    &\leq 
    \int |\rho_\alpha(q_1(\alpha)-z) - \rho_\alpha(q_2(\alpha)-z)| d\alpha
    \\
    &\leq 
    \int |q_2(\alpha) - q_1(\alpha)| d\alpha
    \\
    &= \|q_2 - q_1\|_{1}.
\end{align*}

\end{proof}

\begin{lemma}[Lemma 1 in main body]\label{lemma:R_F}
The Rademacher complexity on $L\circ q \in \mathcal{F}$ is upper-bounded by
\[
\mathfrak{R}_N(\mathcal{F}) \leq \mathfrak{R}_N(\mathcal{Q}),
\]
\end{lemma}
where $q \in \mathcal{Q}$.
\begin{proof}
The proof is immediate from Lemma~\ref{lemma:crps_lipschitz} and the property of Rademacher complexity.
\end{proof}

\subsection{Proof of Theorem 1}

Recall that 
$\mathcal{D}_\mathrm{test} = \{x_i\}$ with  $x_{i}=(z_{i, 1:T}, \xi_{i, 1:T+\tau})$,  
$\mathcal{D}_\mathrm{train}= \{z_{i,T-\tau+1:T}, x_i^{-\tau}\}$, and ${\mathcal{D}}^{-\tau}_\mathrm{test}= \{x_i^{-\tau}\}_{i=1}^{m}$ with $x^{-\tau}_{i}=(z_{i, 1:T-\tau}, \xi_{i, 1:T})$.

\begin{lemma}\label{lemma:boundedness_phi}
Let $\{\bar z_{i,t}\}_{i,t}$ be the same as $\{z_{i,t}\}_{i,t}$ for all $1 \leq t\leq T$ and $i\in [m]$ except for one element, i.e.,  $\bar z_{i, t}\neq z_{i, t}$ for some $T-\tau +1\leq t\leq T$ and some $i\in [m]$, and $\bar z_{i, t}= z_{i, t}$ otherwise. 
Then, for $\phi(\{z_{i,T-\tau+1:T}, x_i^{-\tau}\} ):= \sup_\theta [ 
    \mathcal{L}(\theta \mid  \{x_i^{-\tau}\})
    -
    \hat{\mathcal{L}}(\theta; \{z_{i,T-\tau+1:T}, x_i^{-\tau}\})]$,
\[
|\phi(\{z_{i,T-\tau+1:T}, x_i^{-\tau}\} )
-\phi(\{\bar z_{i,T-\tau+1:T}, x_i^{-\tau}\} )|
\leq 
\frac{2D}{m\tau},
\]
holds.
\end{lemma}
\begin{proof}
\begin{align*}
|\phi(\{z_{i,T-\tau+1:T}, x_i^{-\tau}\} )
-\phi(\{\bar z_{i,T-\tau+1:T}, x_i^{-\tau}\} )|
&\overset{(a)}{\leq} 
\sup_\theta |\hat{\mathcal{L}}(\theta; \{\bar z_{i,T-\tau+1:T}, x_i^{-\tau}\})
-
\hat{\mathcal{L}}(\theta; \{z_{i,T-\tau+1:T}, x_i^{-\tau}\})|
\\
&\overset{(b)}{\leq} 
\frac{1}{m\tau} \sum_{i=1}^m \sum_{t=T-\tau + 1}^{T} |
L (\bar z_{i,t}, q^t_{\tilde \theta}(\cdot \mid x_i^{-\tau}))
-
L (z_{i,t}, q^t_{\tilde \theta}(\cdot \mid x_i^{-\tau}))|\\
&\overset{(c)}{\leq} 
\frac{2D}{m\tau},
\end{align*}
where (a) holds due to $|\sup A - \sup B| \leq \sup |A-B|$, (b) holds due to sub-additivity of the supremum and the definition of the supremum attainer $\tilde \theta$, and (c) holds due to the assumption on $\{z_{i,t}\}$, and $\{\bar z_{i,t}\}$ and Lemma~\ref{lemma:boundedness_L}. 
\end{proof}

\begin{lemma}\label{lemma:concentration_phi}
For $\phi(\mathcal{D}_\mathrm{train}) = \phi(\{z_{i,T-\tau+1:T}, x_i^{-\tau}\} ):= \sup_\theta [ 
    \mathcal{L}(\theta \mid  \{x_i^{-\tau}\})
    -
    \hat{\mathcal{L}}(\theta; \{z_{i,T-\tau+1:T}, x_i^{-\tau}\})]=\sup_\theta[
    \mathcal{L}(\theta\mid  \mathcal{D}^{-\tau}_\mathrm{test})
    -
    \hat{\mathcal{L}}(\theta; \mathcal{D}_\mathrm{train})
    ]$,
\[
\Pr
\left( 
\phi(\mathcal{D}_\mathrm{train})
- 
\E 
\phi(\mathcal{D}_\mathrm{train})
\ge \epsilon
\right)
\le
e^{-m\tau\epsilon^2/2D^2},
\]
holds where $\E$ is the expectation over $Z_{i, T-\tau+1:T }$ 
conditioned on $x_i^{-\tau}$ for $i\in [m]$ appeared in
$\mathcal{L}(\theta\mid  \mathcal{D}^{-\tau}_\mathrm{test})$
.
\end{lemma}
\begin{proof}
The result is immediate through applying Lemma~\ref{lemma:boundedness_phi} into 
McDiarmid's inequality.
\end{proof}





\begin{assumption}\label{assumption:boundedness_target}
The target time series $Z_{i,t}$ are bounded, i.e., $\|Z_{i,t}\|\leq D$ for some $D\geq 0$.
\end{assumption}

\begin{definition}
The temporal discrepancy over $\tau$ time difference is defined as:
\[
\Delta_\mathrm{dis}^\tau (\mathcal{D}_\mathrm{test})
=
\sup_\theta[
\mathcal{L}(\theta\mid \mathcal{D}_\mathrm{test}) 
- 
\mathcal{L}^\mathrm{}(\theta\mid {\mathcal{D}}^{-\tau}_\mathrm{test})
].
\]
Here, 
${\mathcal{D}}^{-\tau}_\mathrm{test}= \{x_i^{-\tau}\}_{i=1}^{m}$
denotes the backtest data
with $\tau$-shifted input $x_i^{-\tau}=(z_{i, 1:T-\tau}, \xi_{i, 1:T})$.
\end{definition}



\begin{theorem}[Theorem 1 in the main body]
\label{thm:con_gen_error}
For any $\delta > 0$, a conditional generalization error
\begin{align*}
   \mathcal{L}&(\hat \theta\mid \mathcal{D}_\mathrm{test})- \mathcal{L}(\theta^*\mid \mathcal{D}_\mathrm{test})
\leq 
2 \Delta_\mathrm{dis}^\tau(\mathcal{D}_\mathrm{test}) 
+ 
2 \mathfrak{R}_{m\tau}(\mathcal{Q}\mid \mathcal{D}^{-\tau}_\mathrm{test}) 
+ \frac{4D}{\sqrt{m \tau}}\sqrt{\log\left(\frac{1}{\delta}\right)},
\end{align*}
holds with at least  $1-\delta$ probability.  
Here,  
$\mathfrak{R}_{m\tau}(\mathcal{Q} \mid  \mathcal{D}^{-\tau}_\mathrm{test})$ denotes the Rademacher complexity of $\mathcal{Q}$, where the expectation on target data is conditioned on the backtest data.
\end{theorem}

\begin{proof}
Let $\mathcal{L}(\theta) :=\mathcal{L}(\theta\mid \mathcal{D}_\mathrm{test})$ and $\hat{\mathcal{L}}(\theta)= \hat{\mathcal{L}}(\theta;\mathcal{D}_\mathrm{train})$,
together with
$\hat \theta = \mathrm{argmin}~\hat{\mathcal{L}}(\theta )$
and 
$\theta^* = \mathrm{argmin}~\mathcal{L}(\theta)$.  
Then 
\begin{align}
    \mathcal{L}(\hat \theta)- \mathcal{L}(\theta^*)
    &=
    \mathcal{L} (\hat \theta) - \hat{\mathcal{L}}(\hat \theta) 
    + 
    \hat{\mathcal{L}}(\hat \theta) - \hat{\mathcal{L}} (\theta^*)
    +
    \hat{\mathcal{L}}(\theta^*) - \mathcal{L}(\theta^*) 
    \nonumber \\ 
    &\leq
    \mathcal{L} (\hat \theta) - \hat{\mathcal{L}} (\hat \theta) 
    +
    \hat{\mathcal{L}}(\theta^*) - \mathcal{L}(\theta^*) 
    \nonumber \\  
    &\leq
    2 \sup_\theta |\mathcal{L}(\theta) - \hat{\mathcal{L}}(\theta)|.
    \label{eq:gen_error_sup}
\end{align}

The sub-additivity of the supremum gives 
\begin{align}
    \sup[
    \mathcal{L}(\theta \mid  \mathcal{D}_\mathrm{test}) - \hat{\mathcal{L}}(\theta; \mathcal{D}_\mathrm{train}) 
    ]
    & \leq 
    \underbrace{
    \sup[
    \mathcal{L}(\theta\mid  \mathcal{D}_\mathrm{test}) 
    - 
    \mathcal{L}(\theta\mid \mathcal{D}^{-\tau}_\mathrm{test})
    ]
    }_{\Delta^{\tau}_\mathrm{dis}(\mathcal{D}_\mathrm{test})}
    +
    \underbrace{
    \sup[
    \mathcal{L}(\theta\mid  \mathcal{D}^{-\tau}_\mathrm{test})
    -
    \hat{\mathcal{L}}(\theta; \mathcal{D}_\mathrm{train})
    ]
    }_{\phi(\mathcal{D}_\mathrm{train})},
    \label{eq:1}
\end{align}
where 
\begin{align*}
    L(\theta \mid \mathcal{D}_\mathrm{test})
&= 
\frac{1}{m\tau}\sum_{i=1}^m \left[\sum_{t=T+1}^{T+\tau} \E \left[L(Z_{i,t}, q^t_\theta(\cdot \mid X_i)) \bigm| X_i =x_i\right]
\right],
\\
L(\theta \mid \mathcal{D}^{-\tau}_\mathrm{test})
&= 
\frac{1}{m\tau}\sum_{i=1}^m \left[\sum_{t=T-\tau+1}^{T} \E \left[L(Z_{i,t}, q^t_\theta(\cdot \mid X_i^{-\tau})) \bigm| X_i^{-\tau} =x_i^{-\tau}\right]
\right],
\\
\hat{\mathcal{L}}(\theta; \mathcal{D}_\mathrm{train})
&=  
\frac{1}{m\tau}\sum_{i=1}^m \sum_{t=T-\tau+1}^{T}  L (z_{i,t}, q^t_\theta(\cdot \mid x_i^{-\tau}))
.
\end{align*}

The first term on RHS follows from the definition 
of $\Delta^\tau_\mathrm{dis}(\mathcal{D}_\mathrm{test})$.

For the second term $\phi(\mathcal{D}_\mathrm{train})$, let's define 
\[
L^{t}(h )
= 
\frac{1}{m\tau}\sum_{i=1}^m \left[\sum_{\gamma=T-\tau+1}^{t} \E \left[L(Z_{i,t}, q^{\gamma}_\theta(\cdot \mid X_i^{-\tau})) \bigm| X_i^{-\tau} =x_i^{-\tau}\right]
+
\sum_{\gamma=t+1}^{T} L(z_{i,t}, q^{\gamma}_\theta(\cdot \mid x_i^{-\tau}))
\right],
\] 
for any $t=T -\tau+1, \ldots, T$.  
Then telescoping the term gives
\begin{align}
    \mathcal{L}(\theta\mid  \mathcal{D}^{-\tau}_\mathrm{test})
    -
    \hat{\mathcal{L}}(\theta; \mathcal{D}_\mathrm{train})
    &=
    \mathcal{L}(h\mid  \{x_i^{-\tau}\})
    -
    \hat{\mathcal{L}}(h; \{z_{i,T-\tau+1:T}, x_i^{-\tau}\})  \nonumber 
    \\ \nonumber &=
    L^{T }(h)-L^{T-\tau}(h) \nonumber
    \\\nonumber
    &=
    \sum_{t=T-\tau+1}^{T} L^{t}(h )-L^{t-1}(h)
    \\
    &=\sum_{i=1}^m \frac{1}{m\tau} \bigg[\sum_{t=T-\tau+1}^{T} 
    \E \left[L(Z_{i,t}, q^{t}_\theta(\cdot \mid X_i^{-\tau})) \bigm| X_i^{-\tau} =x_i^{-\tau} \right]
    -
    L(z_{i,t}, q^{t}_\theta(\cdot \mid x_i^{-\tau})) \bigg] \nonumber.
\end{align}

Now the expectation of $\phi$ conditioned on 
$\{x_i^{-\tau}\}$ 
is
\begin{align}
    \E \phi(\mathcal{D}_\mathrm{train}) 
&= 
\E \sup[
\mathcal{L}(\theta\mid \mathcal{D}^{-\tau}_\mathrm{test})
-
\hat{\mathcal{L}}(\theta; \mathcal{D}_\mathrm{train}) 
] \nonumber
\\
&
\overset{(a)}{\leq} 
\E 
\sup
\frac{1}{m\tau}
\sum_{i, t}
|\E f(\theta;i, t)-f(\theta;i, t)| \nonumber
\\
&\overset{(b)}{\leq}
\mathfrak{R}_{m\tau}(\mathcal{F}\mid \{x_i^{-\tau}\}), 
\label{eq:E_phi_R_F}
\end{align}
where (a) holds by defining 
\[
\begin{aligned}
\E f(\theta;i, t) &= \E_{Z_{i,t}|X_i^{-\tau}} \left[L(Z_{i,t}, q^t_\theta(\cdot \mid X_i^{-\tau})) \bigm| X_i^{-\tau}=x_i^{-\tau} \right], \\
 f(\theta;i, t) &= L(z_{i,t}, q^t_\theta(\cdot \mid x_i^{-\tau})),
\end{aligned}
\]
and (b) holds due to the standard symmetrization trick followed by the definition of Rademacher complexity on $f=L \circ q\in \mathcal{F}$. 

Aggregating~\eqref{eq:1} and \eqref{eq:E_phi_R_F} gives 
\begin{align*}
G_{m\tau}:=\sup_\theta |\mathcal{L}(\theta) - \hat{\mathcal{L}}(\theta)|
- \left(
    \Delta_\mathrm{dis}^\tau (\mathcal{D}_\mathrm{test})
    + \E \phi(\mathcal{D}_\mathrm{train})
    \right)
    &\leq 
    \phi(\mathcal{D}_\mathrm{train})- \E \phi(\mathcal{D}_\mathrm{train}).
\end{align*}

Writing $\phi:= \phi(\mathcal{D}_\mathrm{train})$ in shorthand, we have 
\begin{align*}
    \Pr(G_{m\tau} \ge \epsilon) 
    &\leq 
    \Pr 
    \left(
    \phi - \E \phi
    \geq \epsilon
    \right)
    \\ & \overset{(c)}{\leq} 
    e^{-m\tau\epsilon^2/2D^2},
\end{align*}
where (c) hold due to Lemma~\ref{lemma:concentration_phi}.
Choosing $\delta > 0$ and setting
\[
\epsilon = 
 \sqrt{\frac{2D^2}{m \tau}\log\left(\frac{1}{\delta}\right)},
\]
guarantees the following
w.p. $1-\delta$:
\begin{align}
    \sup |\mathcal{L}(\theta)- \hat{\mathcal{L}}(\theta)| 
    &\leq 
    \Delta_\mathrm{dis}^\tau (\mathcal{D}_\mathrm{test})
    + \E \phi
    + \sqrt{\frac{2D^2}{m \tau}\log\left(\frac{1}{\delta}\right)} \nonumber
    \\
    &\overset{(d)}{\leq}
    \Delta_\mathrm{dis}^\tau (\mathcal{D}_\mathrm{test})
    +
    \mathfrak{R}_{m\tau}(\mathcal{Q}\mid \mathcal{D}^{-\tau}_\mathrm{test})
    + \sqrt{\frac{2D^2}{m \tau}\log\left(\frac{1}{\delta  }\right)} \nonumber,
\end{align}
where (d) holds due to Equation~\eqref{eq:E_phi_R_F} and Lemma~\ref{lemma:R_F}. Finally, combining with~\eqref{eq:gen_error_sup} gives the desired result.
\end{proof}
\subsection{Proof of Lemma 2}

\begin{lemma}[Lemma 2 in the main body]
Assume each input time series is stationary or periodic. Then the expected temporal discrepancy
$\Delta_\mathrm{dis}^\tau  = \sup_\theta\E[
\mathcal{L}(\theta\mid \mathcal{D}_\mathrm{test}) 
- 
\mathcal{L}^\mathrm{}(\theta\mid {\mathcal{D}}^{-\tau}_\mathrm{test})] = 0$. 
\end{lemma}

\begin{proof}

Recall the temporal discrepancy over $\tau$ time difference
$\Delta_\mathrm{dis}^\tau (\mathcal{D}_\mathrm{test})
=
\sup_\theta[
\mathcal{L}(\theta\mid \mathcal{D}_\mathrm{test}) 
- 
\mathcal{L}^\mathrm{}(\theta\mid {\mathcal{D}}^{-\tau}_\mathrm{test})$.

Equivalently, 
\begin{align}
\Delta^{\tau}_\mathrm{dis}(\mathcal{D}_\mathrm{test})
&=
\sup\bigg[
\frac{1}{m\tau}\sum_{i=1}^m \sum_{t=T+1}^{T+\tau} \E \left[L(Z_{i,t}, q^t_\theta(\cdot \mid x_i))  \right] 
- 
\E \left[ L(Z_{i,t-\tau}, q^{t-\tau}_\theta(\cdot \mid x_i^{-\tau})) \right]
\bigg] \nonumber
\\
&=
\sup\bigg[
\frac{1}{m\tau}\sum_{i=1}^m 
\sum_{t=T+1}^{T+\tau} 
\sum_{\gamma=t-\tau+1}^{t}
\E \left[L(Z_{i,\gamma}, q^\gamma_\theta(\cdot \mid x^{\gamma}_i))  \right] 
- 
\E \left[ L(Z_{i,\gamma-1}, q^{\gamma-1}_\theta(\cdot \mid x_i^{\gamma-1})) \right]
\bigg], \nonumber
\end{align}
where we denote $x_i^\gamma = (z_{i,1:\gamma-\tau}, \xi_{1:\gamma})$, $x_i^{T+\tau} := x_i  = (z_{i,1:T}, \xi_{1:T+\tau})$ and $x_i^{T} :=x_i^{-\tau} = (z_{i,1:T-\tau}, \xi_{1:T})$.

Note that, after swapping second and third summation and taking the expectation, applying sub-additivity of supremum gives
\[
\Delta^{\tau}_\mathrm{dis} \leq \sum_{\gamma=T+1}^{T+\tau} \Delta_\mathrm{dis}^\gamma,
\]
where $ \Delta_\mathrm{dis}^\gamma=\sup[\E \mathcal{L}(\theta\mid  \{x_i^{\gamma}\}) - \E \mathcal{L}(\theta\mid  \{x_i^{\gamma-1}\})]$.

The rest of the proof is similar to \cite{mariet2019foundations}. It is immediate to see $ \Delta_\mathrm{dis}^\gamma=0$ as $\Pr (Z_{i, \gamma-\tau+1:\gamma}\mid X_i^\gamma) = \Pr (Z_{i, \gamma-\tau:\gamma-1}\mid X_i^{\gamma-1})$ 
under stationarity 
or periodicity,
which gives the desired result.
\end{proof}

\subsection{Proof of Corollary 1}

\begin{corollary}[Corollary 1 in the main body]
For any $\delta >0$, a generalization error
\begin{align*}
\mathcal{L}(\hat \theta ) - \mathcal{L}(\theta^*) &\leq 
2 \Delta_\mathrm{dis}^\tau 
+ 
2 \mathfrak{R}_{m\tau}(\mathcal{Q})
+ \frac{4D}{\sqrt{m \tau}}\sqrt{\log\left(\frac{1}{\delta}\right)},  
\end{align*}
holds with at least  $1-\delta$ probability, where
$\Delta_\mathrm{dis}^\tau $
denotes the expected temporal discrepancy.
\end{corollary}

\begin{proof}
We begin with expected version $L(\theta)$ on Equation~\eqref{eq:gen_error_sup} and ~\eqref{eq:1} and its empirical version under i.i.d. setup. Then the rest of the proof is basically the same. 
\end{proof}


\section{Metrics}
We use the \texttt{Evaluator} class from GluonTS\footnote{\url{https://github.com/awslabs/gluon-ts/blob/master/src/gluonts/evaluation/.}} to report the following probabilistic error metrics.

\subsection{Weighted Quantile Losses (wQL[$\alpha$])}

For a given quantile $\alpha\in(0,1)$, a target value $z_t$, input $x$, and $\alpha$-quantile prediction $q^t(\alpha \mid x)$, the $\alpha$-quantile 
loss is defined as:
\begin{align*}
\rho_{\alpha}(z_t -q^t(\alpha \mid x)) &= (z_t - q^t(\alpha \mid x))(\alpha - \mathbf{1}\{z_t - q^t(\alpha \mid x) < 0\}).
\end{align*}
We report the normalized sum of quantile losses, 
\[
\text{wQL}[\alpha] = 
2\frac{\sum_{i,t} \rho_\alpha(z_{i,t} - q^t(\alpha \mid x_i))} {\sum_{i,t} |z_{i,t}|},
\]

to compute the weighted quantile losses for a given time span $t = T,, \dots T+\tau$ across all time series.
We include results for the training quantiles $\alpha=0.5, 0.9$  and interpolated quantile $\alpha = 0.7$, and the extrapolated quantile $\alpha = 0.995$. 

\subsection{Mean Weighted Quantile Loss (mean\_wQL)}
We also report the mean weighted quantile loss by averaging the weighted quantile losses over all of the $k = 1, ..., K$ quantiles $\alpha_k$ as:
\[
    \text{mean}\_\text{wQL} = \frac{1}{K}\sum_{k=1}^K \text{wQL}[\alpha_k],
\]
as an approximation to the CRPS.

\subsection{Percent Quantile Crossing}
We quantify the effect of quantile crossing by introducing the perecent quantile crossing metric as the proportion of adjacent quantiles that are crossing for time series at every predicted time step as:


\[
    \frac{1}{m \tau (K-1)} \sum_{\substack{i, t, k}} 
    \mathbf{1}\{ q^t(\alpha_k\mid x_i) > q^t(\alpha_{k+1}\mid x_i), ~ \alpha_k < \alpha_{k+1}\}  \times 100\%,
\]
for some input $x_i$, where $i = 1, \dots, m$, $t = T + 1, \dots, T + \tau$, and $k = 1, \dots, K-1$.
Note that this metric only measures the number of quantile crossing between adjacent quantiles, and is not reflective of the potential total amount of quantile crossings. For example in Figure 4a, we observe extreme crossing with the QF output layer between the 0.1 and 0.9 non-adjacent quantile levels on the \texttt{Elec} dataset, which is not counted under this metric. Hence, this metric is a lower bound on the total number of quantile crossings exhibited by QF.

\subsection{Mean Scaled Interval Score (MSIS)}
The mean scaled interval score (MSIS) \citep{gneiting2007strictly} is another probabilistic metric to measure the sharpness and coverage of the prediction interval. It is also reported in \cite{gasthaus19sqf, kim2021deep}, and used as a metric in the M4 forecasting competition\footnote{ \url{https://www.m4.unic.ac.cy/wp-content/uploads/2018/03/M4-Competitors-Guide.pdf.}}. The $\text{MSIS}[\zeta]$ is defined as follows: 
\[ 
\begin{aligned}
\text{MSIS}[\zeta] =
   \frac{1}{\text{SE}(z)}\Big( \frac{1}{m\tau} \sum_{i,t}(q^t(\alpha^U \mid x_i) - q^t(\alpha^L \mid x_i) + &\frac{2}{\zeta}[(q^t(\alpha^L \mid x_i)-z_{i,t})\mathbf{1}\{z_{i,t} < q^t(\alpha^L \mid x_i)\} \\ &+ (z_{i,t}-q^t(\alpha^U \mid x_i))\mathbf{1}\{z_{i,t}>q^t(\alpha^U \mid x_i)\}]\Big),
\end{aligned}
\]
where $i = 1, \dots m$, $t = T+1, \dots T + \tau$, the upper quantile $\alpha^U = 1-\zeta/2$, and the lower quantile $\alpha^L = \zeta/2$. The seasonal error SE for time series frequency $f$ is given as:
\[
\text{SE}(z) = \frac{1}{m(T-f)}\sum_{i,t'} |z_{i,t'} - z_{i, t'+f}|,
\]
where $t' = 1, \dots T-f$.

\section{Additional Experiments}\label{sec:add_experiments}
\subsection{Setup}\label{sec:exp_setup}

\paragraph{Datasets.}
We use the open-source versions of the benchmarking datasets available in the  GluonTS dataset repository\footnote{\url{https://github.com/awslabs/gluon-ts/blob/master/src/gluonts/dataset/repository/datasets.py.}} as referenced in Table \ref{tab:datasets}.  

\begin{table}[h]
\centering
\begin{tabular}{|l|l|c|c|c|c|c|c|c|c}
\hline
\textsc{domain} & \textsc{name} & \textsc{support} & \textsc{freq} & \textsc{no. ts} & \textsc{avg. len} &  \textsc{pred. len} & \textsc{no. covariates} 

\\
\hline
 electrical load & \texttt{Elec} & $\mathbb{R}^+$ & H & 321 & 21044 & 24 & 4\\ \hline 
road traffic & \texttt{Traf} & $[0, 1]$ & H & 862 & 14036 & 24 & 4\\\hline 
\begin{tabular}[c]{@{}l@{}}visit counts of \\ wikipedia pages \\  \end{tabular} &
    \texttt{Wiki}    & $\mathbb{N}$ & D   & 9535                                                      & 762  & 30   & 3                             \\ \hline
\multirow{5}{*}{\begin{tabular}[c]{@{}l@{}}M4 forecasting \\ competition \\   \end{tabular}} & \texttt{M4-daily} & $\mathbb{R}^+$ & D & 4227  & 2357 &14 & 3\\
 & \texttt{M4-weekly} & $\mathbb{R}^+$ &  W & 359 & 1022 & 13 & 2\\
  & \texttt{M4-monthly} & $\mathbb{R}^+$ & M & 48000 & 216 & 18&1\\
   & \texttt{M4-quarterly} & $\mathbb{R}^+$ & Q  & 24000 & 92 & 8& 1\\
    & \texttt{M4-yearly} & $\mathbb{R}^+$ & Y & 23000 & 31 & 6 & 0\\
\hline
\end{tabular}
\caption{Summary of dataset statistics, where \texttt{Elec} and \texttt{Traf} are dervied from the UCI data repository \citep{Dua:2017}, \texttt{Wiki} from Kaggle \citep{lai_dataset_2017}, and 6 different \texttt{M4} competition datasets \citep{makridakisM4concl}.}
\label{tab:datasets}
\end{table}


\paragraph{Forecast Models and Parameters.} 
We select the MQ-CNN \citep{wen2017multi} model, which is a state-of-the-art time series quantile forecasting model in the sequence-to-sequence framework. The \texttt{MQCNNEstimator}\footnote{\url{https://github.com/awslabs/gluon-ts/blob/master/src/gluonts/model/seq2seq/_mq_dnn_estimator.py.}} is available in the open-source GluonTS \citep{gluonts_jmlr} package. For each dataset, we use the default time features $\mathbf{\xi}_t \in \mathbb{R}^d$, e.g. day of the week and hour of the day, as covariates, where $d$ is determined by the time series frequency.

We use the default hyperparameters optimized for the default QF layer from GluonTS. In our experiments, we only vary the training quantiles from 3 to 5 knots for all layers, and the number of spline knots and epochs for ISQF.  In our experiments, we can simply toggle between the QF and IQF output layers using the \texttt{is\_iqf} hyper-parameter. Similarly, to test the other three output layers, e.g ISQF, SQF and Gaussian, we pass the corresponding \texttt{DistributionOutput} class to the \texttt{distr\_output} hyperparameter. Table \ref{tab:hps} summarizes the hyper-parameter settings for the various output layers.

\begin{table}[h]
\centering
\begin{tabular}{|l|l|c|c|c|c|c}
\hline
Strategy & \texttt{is\_iqf} & \texttt{distr\_output} & \texttt{epochs} & \texttt{no. spline knots (S)} 
\\
\hline
Gaussian & N/A & \texttt{GaussianOutput} & 100 & N/A \\ \hline
SQF & N/A & \texttt{PiecewiseLinearOutput} & 100 & 10 \\ \hline
QF & False & N/A & 100 & N/A \\ \hline
IQF & True & N/A & 100 & N/A \\ \hline
ISQF & N/A & \texttt{ISQFOutput} & 200 & 3 \\
\hline
\end{tabular}
\caption{Hyperparameter settings for the \texttt{MQCNNEstimator} for our I(S)QF and the various other output layer baselines with the training \texttt{quantiles} set to [0.1, 0.5, 0.9] (3) and [0.01, 0.1, 0.5, 0.9, 0.99] (5) in the experiments.}
\label{tab:hps}
\end{table}

\subsection{All Benchmark Results}\label{sec:all_benchmark_results}
\begin{figure}[H]
  \centering
    \begin{subfigure}[b]{0.49\columnwidth}
        \centering
        \includegraphics[width=0.99\columnwidth]{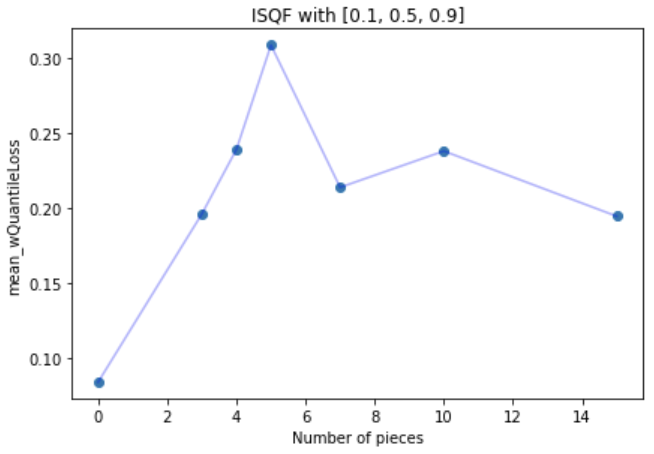}
        \caption{3 quantile knots on \texttt{Elec}.}
        \label{fig:elect_3_knots_spline_evolution}
    \end{subfigure}
    \begin{subfigure}[b]{0.49\columnwidth}
        \centering
        \includegraphics[width=0.99\columnwidth]{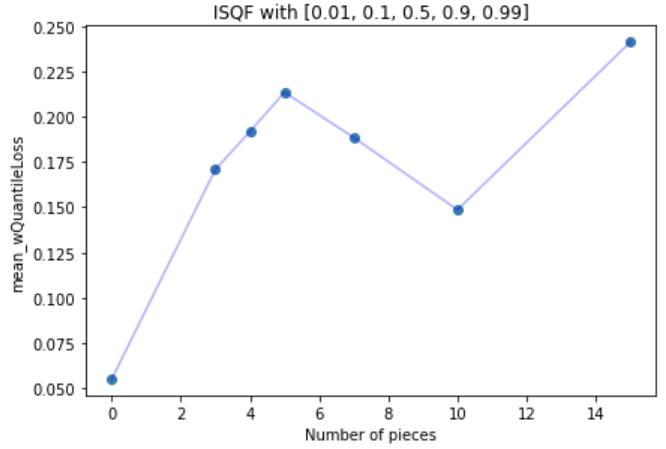}
        \caption{5 quantile knots on \texttt{Elec}.}
        \label{fig:elect_5_knots_spline_evolution}
    \end{subfigure}
    \centering
    \begin{subfigure}[b]{0.49\columnwidth}
        \centering
        \includegraphics[width=0.99\columnwidth]{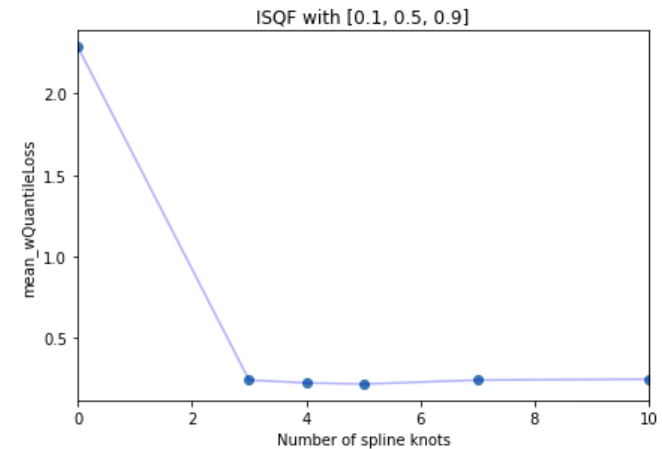}
        \caption{3 quantile knots on \texttt{Traf}.}
        \label{fig:traffic_3_knots_spline_evolution}
    \end{subfigure}
    \begin{subfigure}[b]{0.49\columnwidth}
        \centering
        \includegraphics[width=0.99\columnwidth]{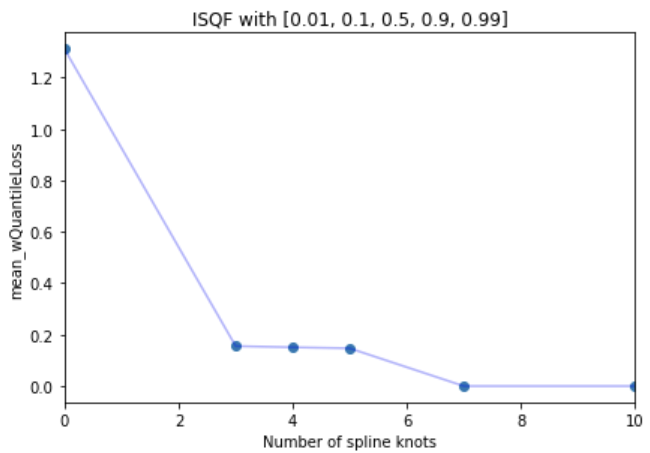}
        \caption{5 quantile knots on \texttt{Traf}.}
        \label{fig:traffic_5_knots_spline_evolution}
    \end{subfigure}
\caption{Experiments that vary the number of spline knots for ISQF on \texttt{Elec} (top) and \texttt{Traf} (bottom) datasets with various quantile knots [0.1, 0.5, 0.9] (3) (left) and [0.01, 0.1, 0.5, 0.9, 0.99] (5) (right).}
\label{fig:spline_evolution}
\end{figure}

 Table \ref{tab:all_benchmark_results} shows the complete benchmarking results on the 9 real-world datasets. 
 \paragraph{ISQF spline knots experiments}
 Figure \ref{fig:spline_evolution} shows the mean weighted quantile loss of ISQF when varying the number of spline knots, both for 3 and 5 quantile knots.
 We observe that the added flexibility and complexity of using spline knots can improve the performance drastically up to a certain point at which it start to slowly deteriorate, forming a U-shaped curve where the left side is much steeper than the right one. The point at which the performance starts to slowly deteriorate depends on the dataset. On some datasets, the added flexibility is needed as we see for the \texttt{Traf} dataset on the left side of the Figures \ref{fig:traffic_3_knots_spline_evolution}-\ref{fig:traffic_5_knots_spline_evolution}.
 On the \texttt{Elec} dataset in Figures \ref{fig:elect_3_knots_spline_evolution}-\ref{fig:elect_5_knots_spline_evolution}, it appears that the added flexibility is hurting the performance, and we see that the mean weighted quantile loss slowly increases as the number of spline knots increase.
 This is a special case where the center of the U-shaped curve is at 0, which corresponds to IQF.
 Similarly, in Figure \ref{fig:traffic_3_knots_spline_evolution}, the performance slowly degrades after 5 spline knots, showing the beginning of the right part of a U-shaped curve.

\paragraph{DeepAR ISQF results}

Table~\ref{tab:results_deepar} shows the results of applying the auto-regressive probabilistic DeepAR model with different benchmark layers and our ISQF. We observe our I(S)QF performing strongly on these datasets.  The experiments are ran using the PyTorch version of the GluonTS~\citep{gluonts_jmlr} package. 
\input{accuracy_table_appendix_deepar}

\begin{figure}[H]
    \centering
    \begin{subfigure}[b]{0.9\columnwidth}
        \centering
        \includegraphics[width=0.85\textwidth]{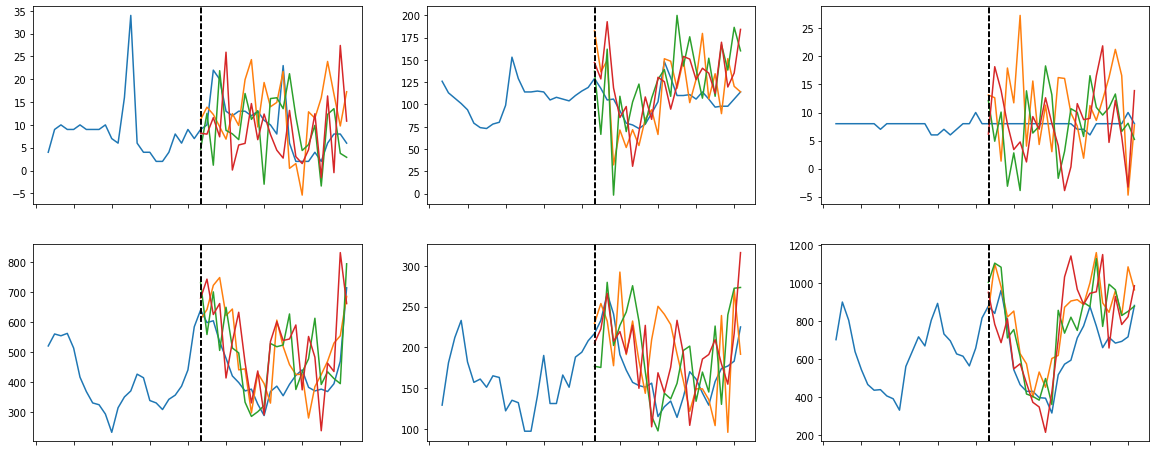}
        \caption{Gaussian}
    \end{subfigure}
    \hfill
    \begin{subfigure}[b]{0.9\columnwidth}
        \centering
        \includegraphics[width=0.85\textwidth]{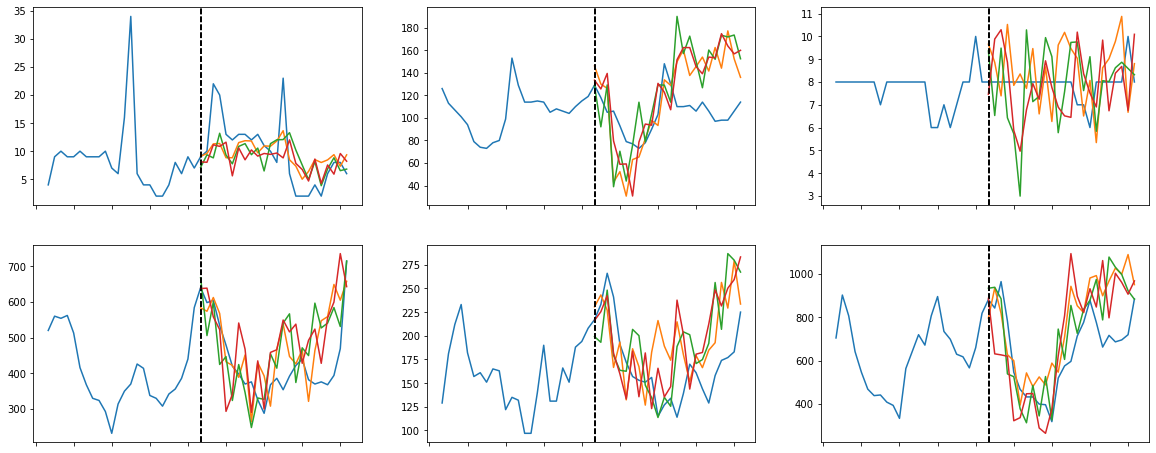}
        \caption{SQF}
    \end{subfigure}
    \hfill
    \begin{subfigure}[b]{0.9\columnwidth}
        \centering
        \includegraphics[width=0.85\textwidth]{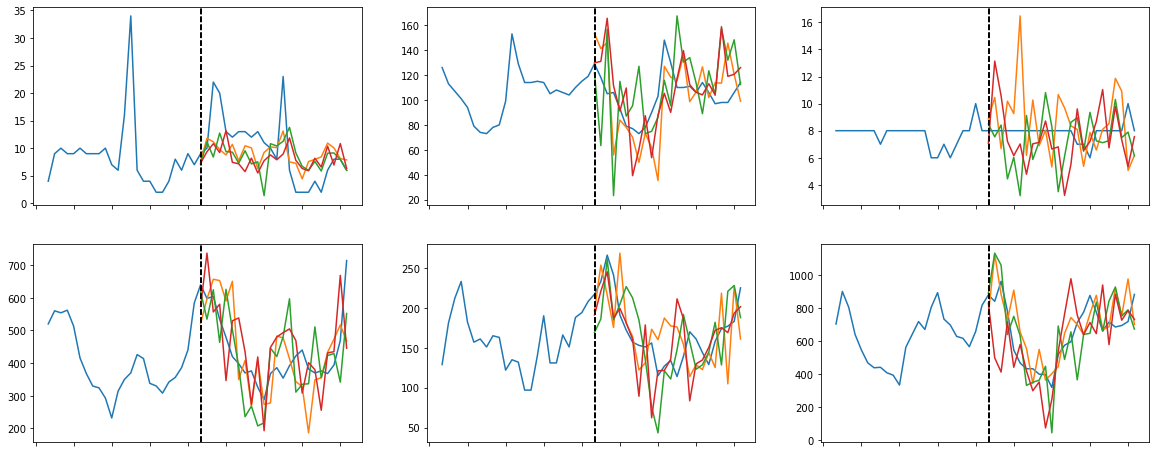}
        \caption{IQF}
    \end{subfigure}
    \hfill
    \begin{subfigure}[b]{0.9\columnwidth}
        \centering
        \includegraphics[width=0.85\textwidth]{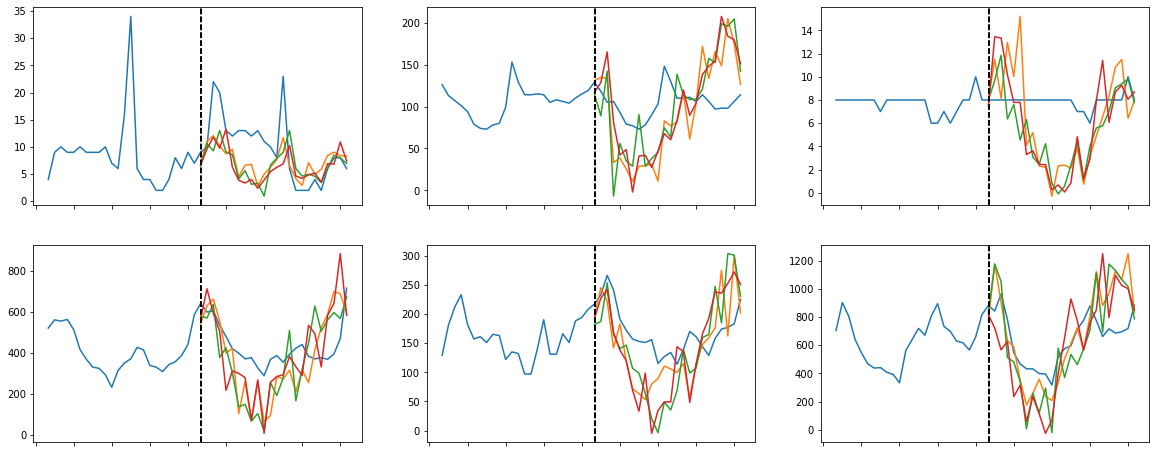}
        \caption{ISQF}
    \end{subfigure}
\caption{Sample paths generated by using MQ-CNN with our I(S)QF and various benchmark layers. Three sample paths are generated for each of the 6 time series from the \texttt{Elec} dataset. The dashed vertical lines represent the start of the prediction horizon. 
}
\label{fig:sample_path}
\end{figure}

\paragraph{Sample path generation}
Figure~\ref{fig:sample_path} visualizes the sample paths of applying the MQ-CNN model with different benchmark layers and our I(S)QF.

\input{accuracy_table_appendix}

\section{Distribution Recovery with IQF}\label{sec:distribution_recovery}
\subsection{Multimodal Distribution}

\begin{figure}[h]
    \centering
    \begin{subfigure}[b]{0.32\textwidth}
        \includegraphics[width=\columnwidth]{Multimodal_quantile_functions_5.pdf}
        \caption{IQF with 5 knots.}
        
        \label{fig:mm_qfs_5}
    \end{subfigure}
    ~ 
    \begin{subfigure}[b]{0.32\textwidth}
        \includegraphics[width=\columnwidth]{Multimodal_quantile_functions_21.pdf}
        \caption{IQF with 20 knots.}
        \label{fig:mm_qfs_20}
    \end{subfigure}
    ~ 
    \\
    \begin{subfigure}[b]{0.32\textwidth}
        \includegraphics[width=\columnwidth]{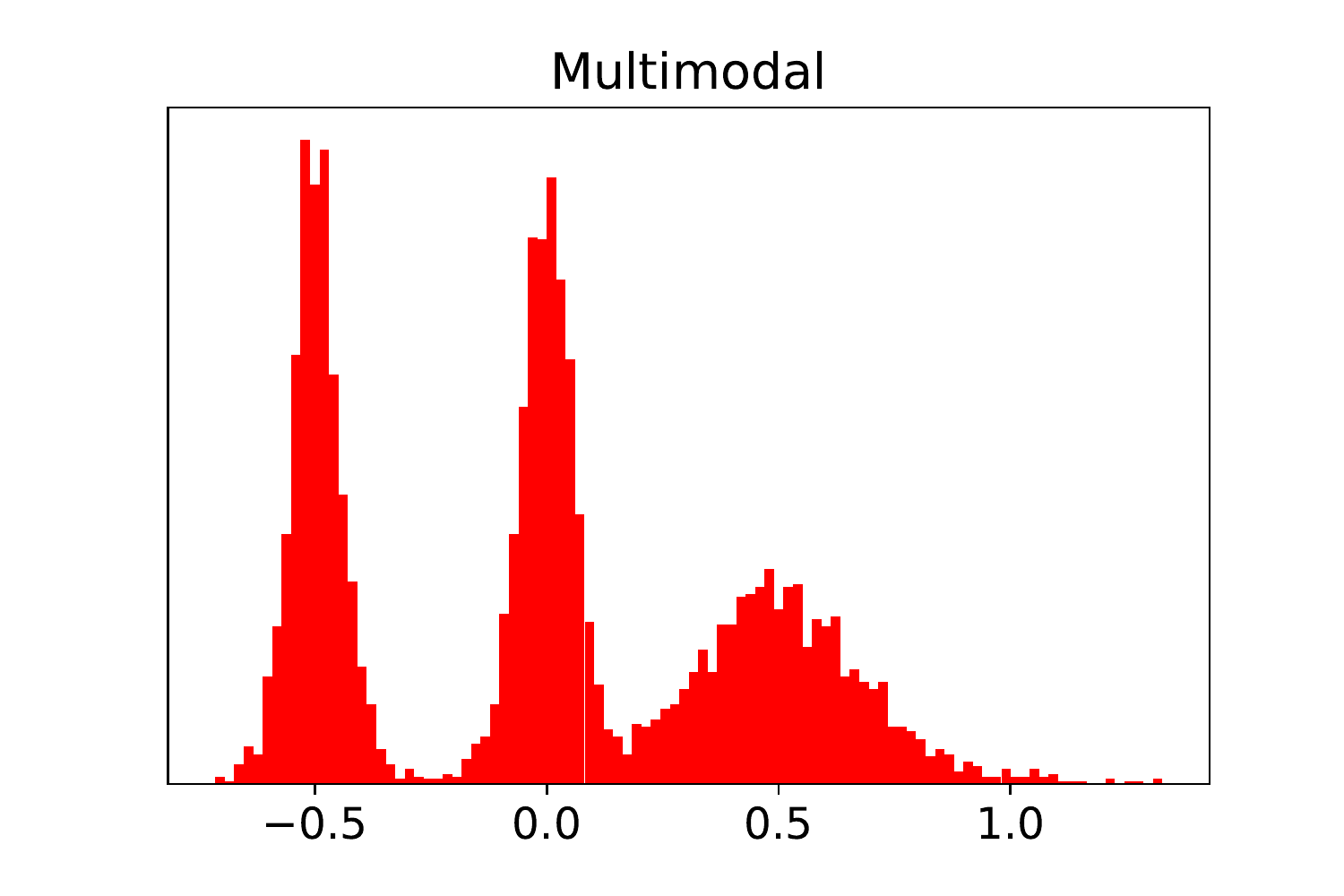}
        \caption{Empirical PDF: true QF.}
        \label{fig:mm_true_emp}
    \end{subfigure}
    \begin{subfigure}[b]{0.32\textwidth}
        \includegraphics[width=\columnwidth]{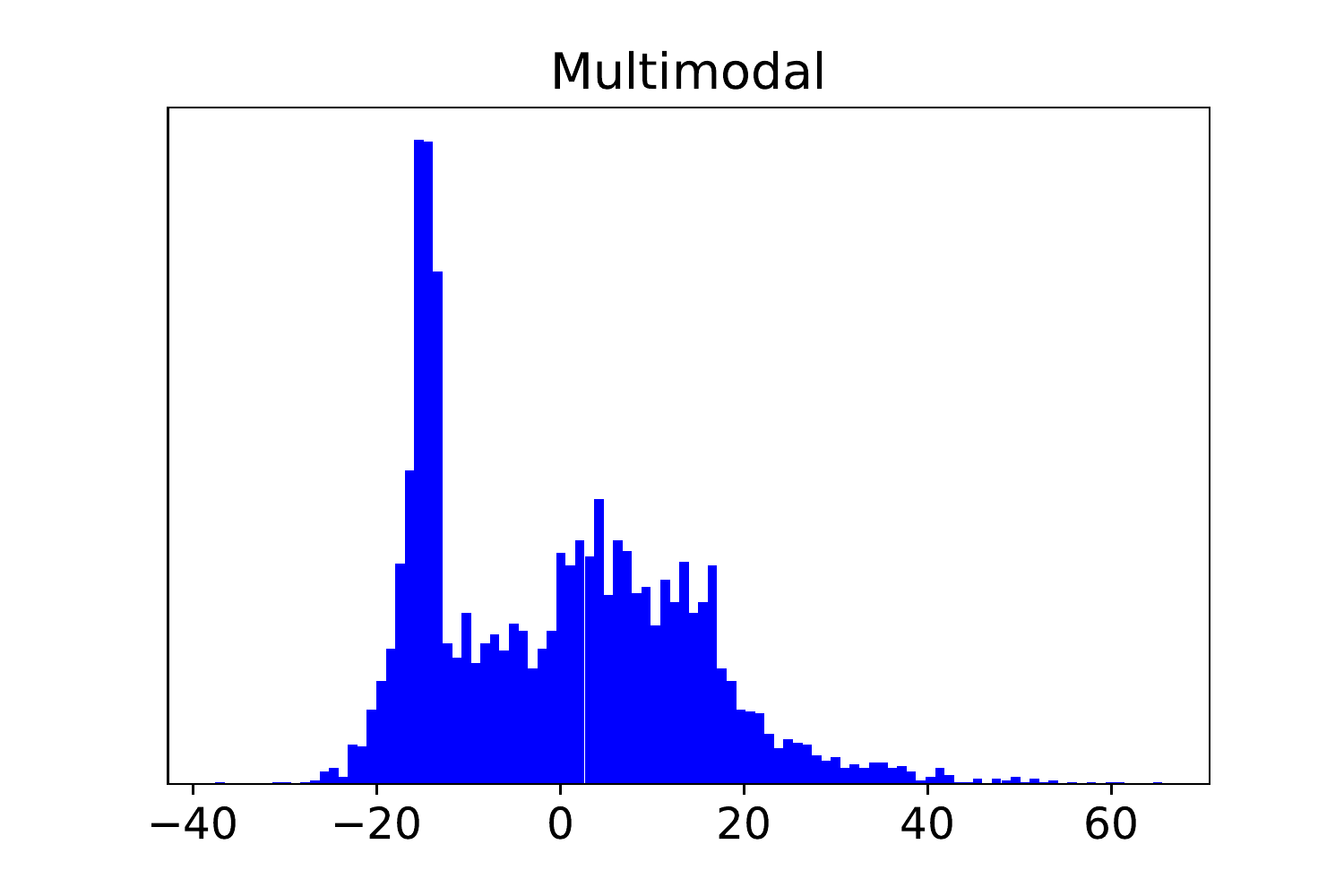}
        \caption{Empirical PDF: IQF with 5 knots.}
        \label{fig:mm_appx_emp_5}
    \end{subfigure}
    \begin{subfigure}[b]{0.32\textwidth}
        \includegraphics[width=\columnwidth]{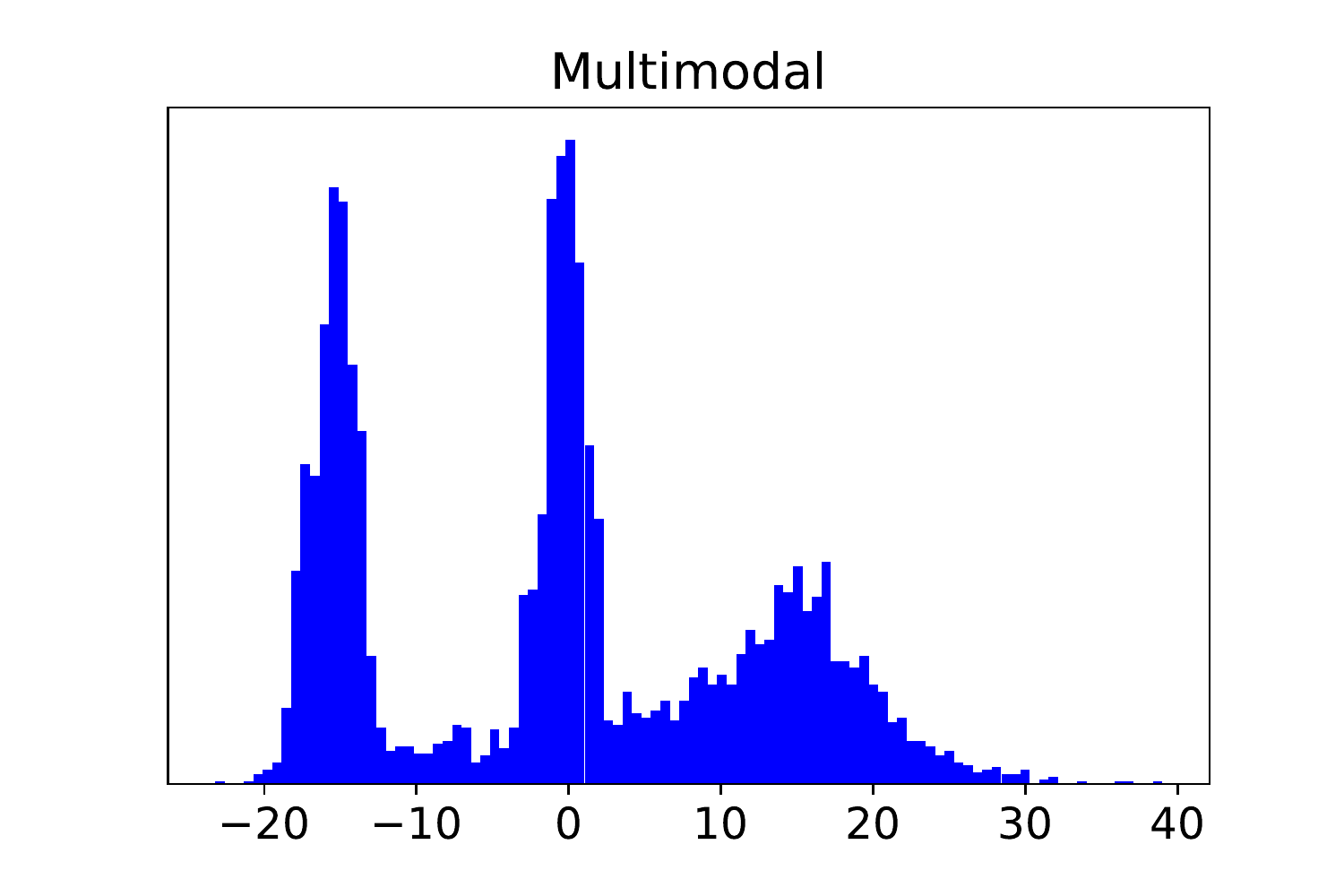}
        \caption{Empirical PDF: IQF with 20 knots.}
        \label{fig:mm_appx_emp_20}
    \end{subfigure}
\caption{Toy Multimodal distribution with 3 peaks: IQF interpolates and extrapolates around training knots to approximate a Multimodal distribution (c) without any distributional assumption, even more accurately as the number of knots increases from 5 ((a), (d)) to 20 ((b), (e)). }

\label{fig:iqf_distribution_free_multi_modal}
\end{figure}

\subsection{Cauchy Distribution}

\begin{figure}[H]
    \centering
    \begin{subfigure}[b]{0.32\textwidth}
        \includegraphics[width=\columnwidth]{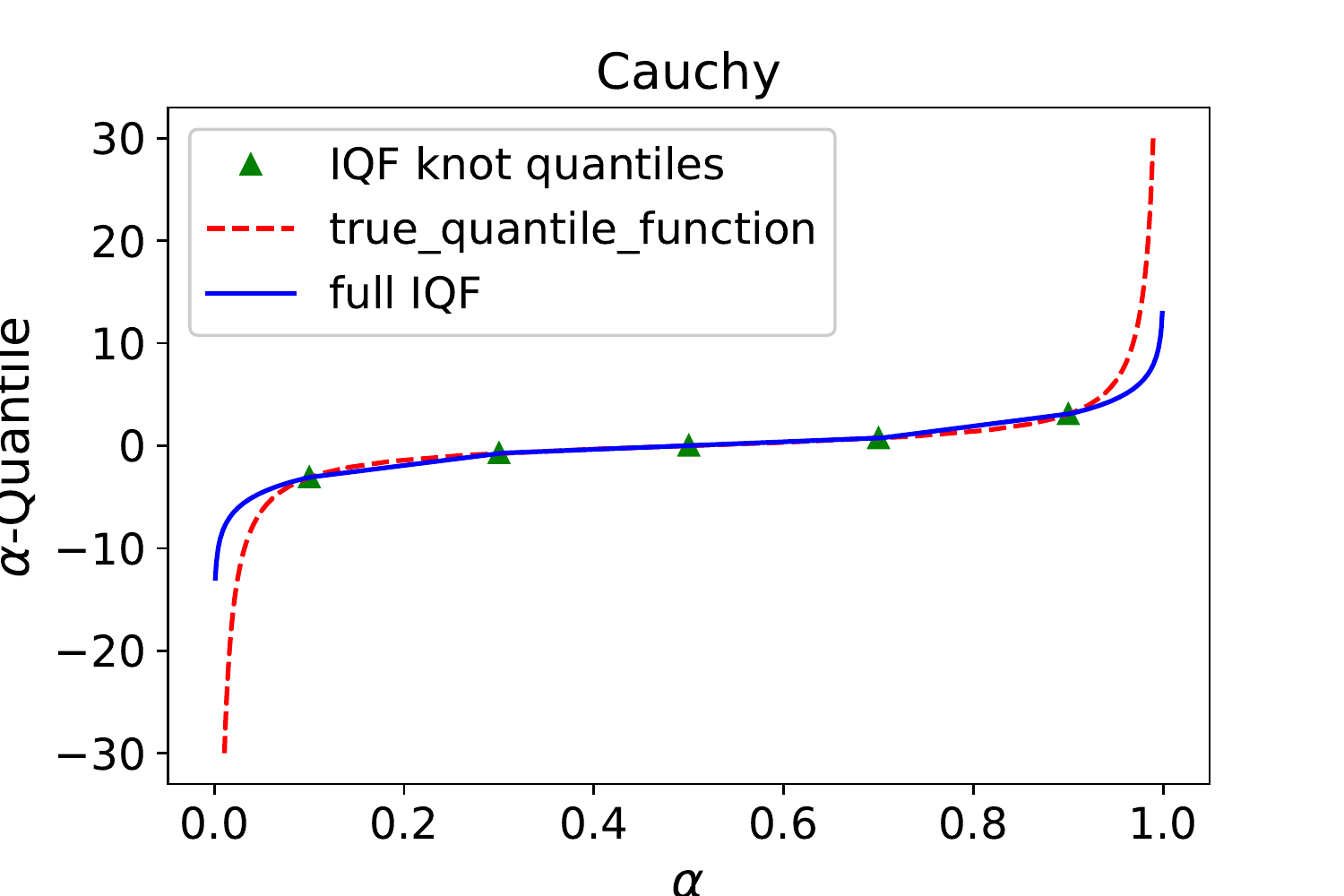}
        \caption{IQF with 5 knots.}
        
        \label{fig:gull}
    \end{subfigure}
    ~ 
    \begin{subfigure}[b]{0.32\textwidth}
        \includegraphics[width=\columnwidth]{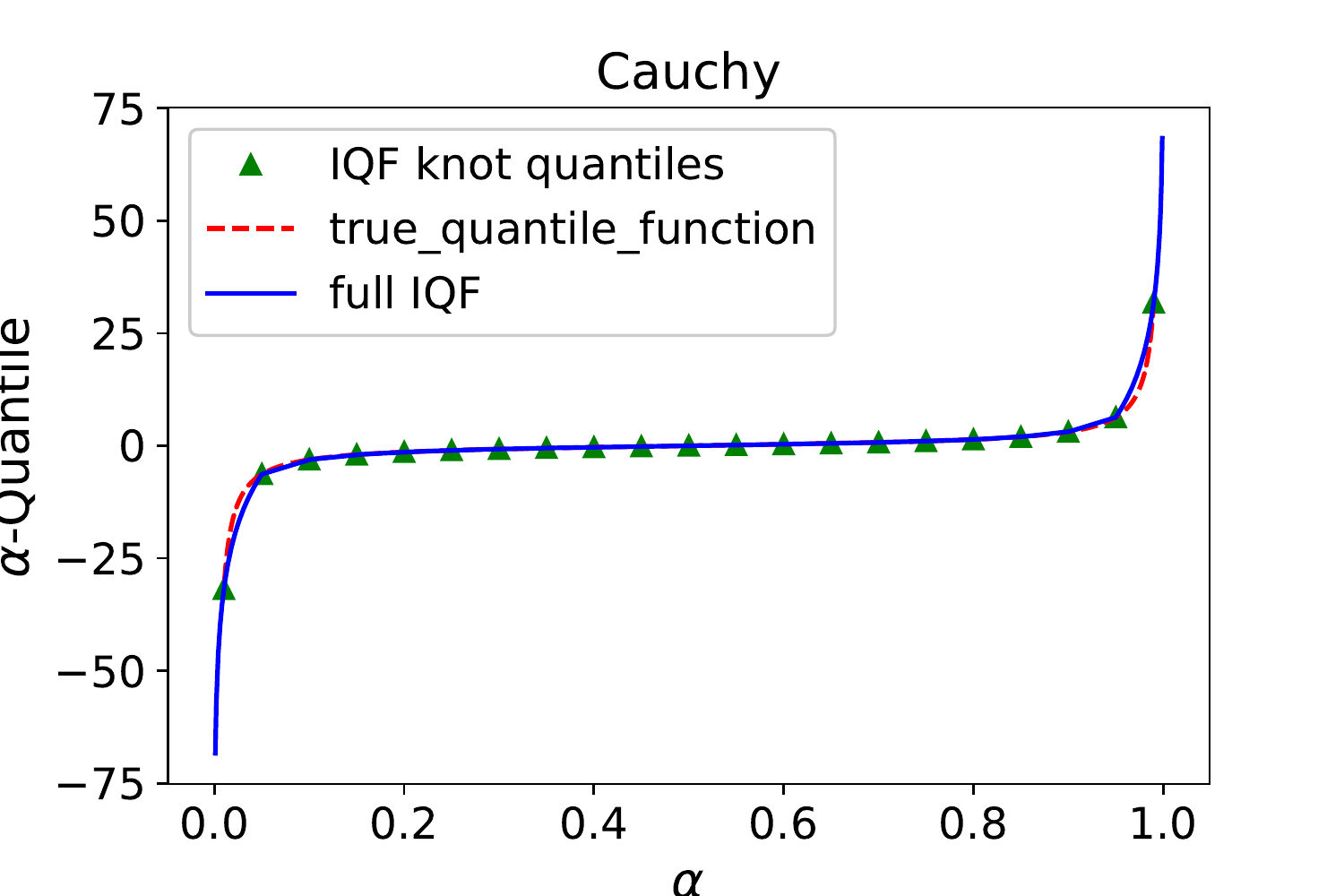}
        \caption{IQF with 20 knots.}
        \label{fig:gull}
    \end{subfigure}
    ~ 
    \\
    \begin{subfigure}[b]{0.32\textwidth}
        \includegraphics[width=\columnwidth]{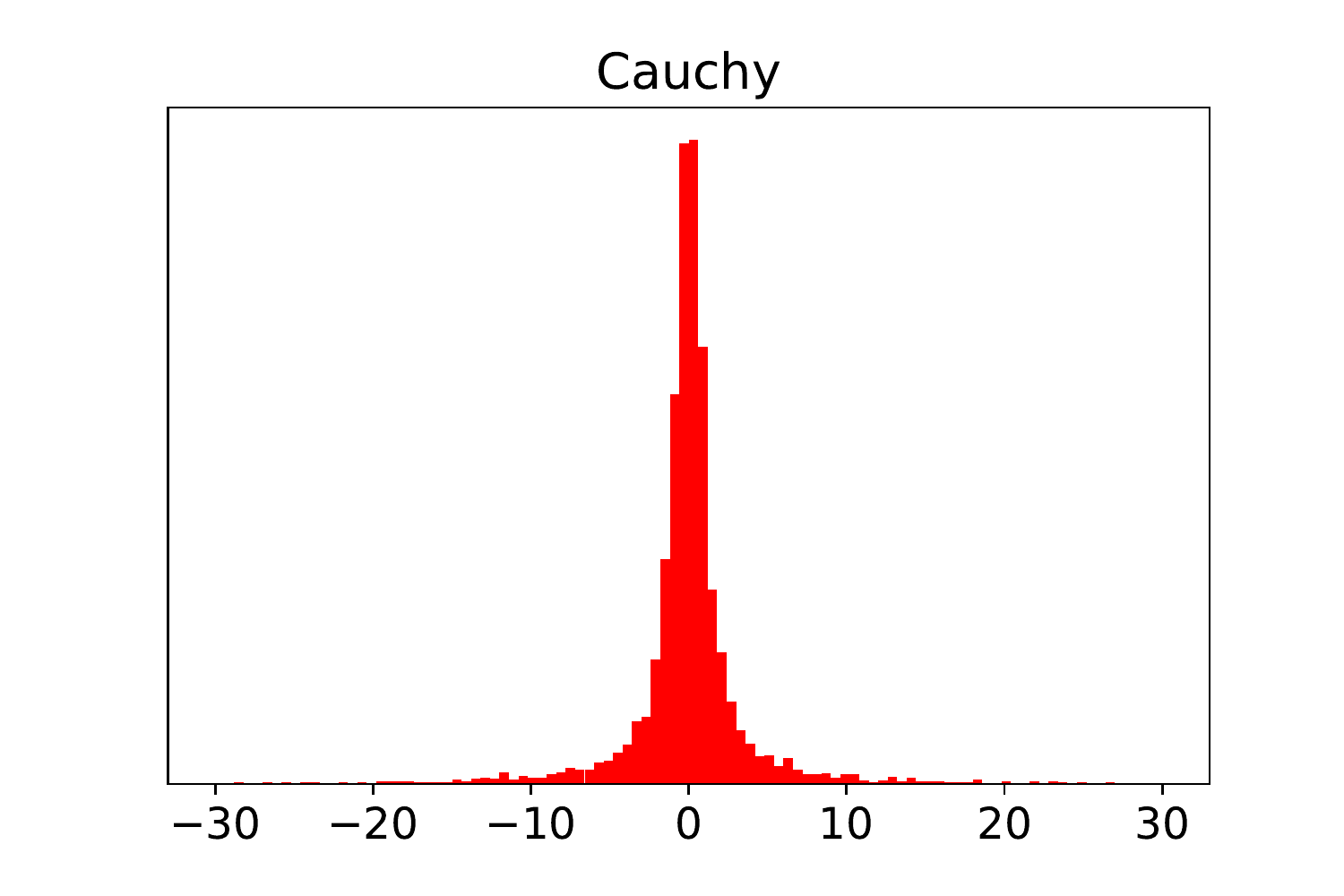}
        \caption{True QF Empirical PDF.}
        \label{fig:tiger}
    \end{subfigure}
    \begin{subfigure}[b]{0.32\textwidth}
        \includegraphics[width=\columnwidth]{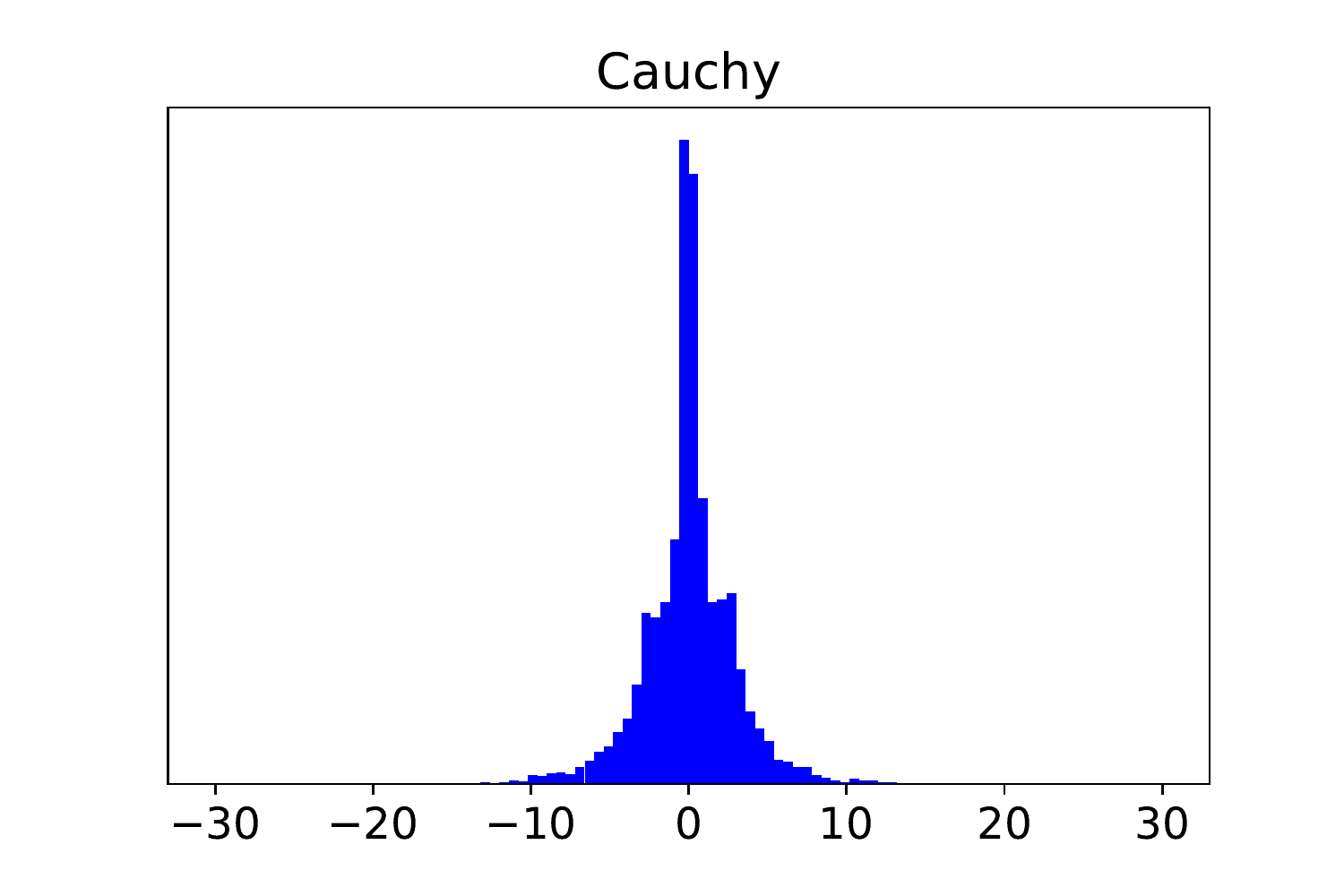}
        \caption{IQF Empirical PDF with 5 knots.}
        \label{fig:mouse}
    \end{subfigure}
    \begin{subfigure}[b]{0.32\textwidth}
        \includegraphics[width=\columnwidth]{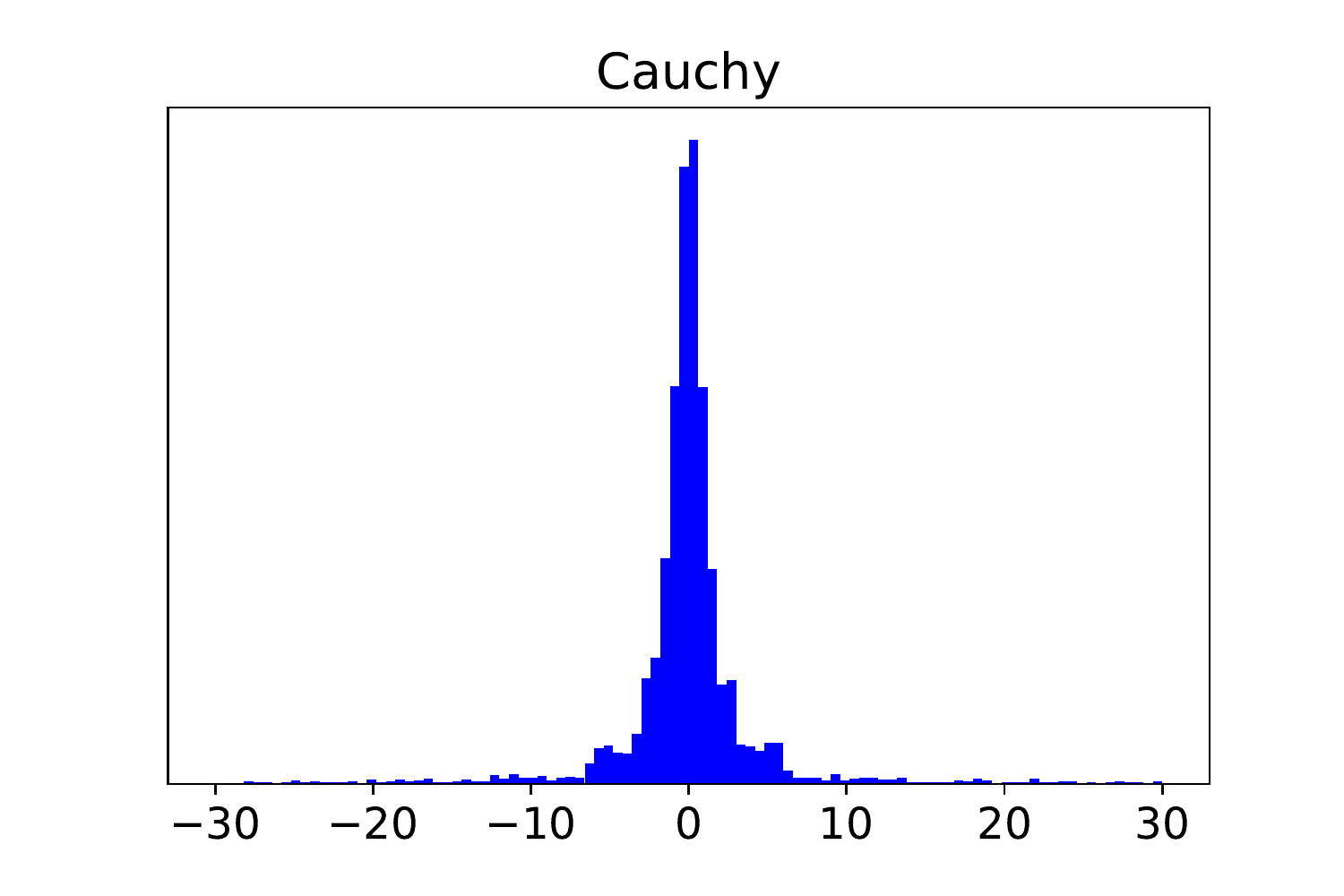}
        \caption{IQF Empirical PDF with 20 knots.}
        \label{fig:mouse}
    \end{subfigure}
\caption{IQF fits a Cauchy distribution as a distribution-free approach with a various number of training knots.}

\label{fig:iqf_distribution_free_cauchy}
\end{figure}















%% file: accuracy_table_appendix_deepar.tex
\begin{table}[H] 
\centering
\scalebox{0.68}{
\begin{tabular}{
|c|c|c|c|c|c|c|c|c|c|
 }
\hline
Dataset &
  Strategy &
  mean\_wQL &
  crossing \% &
  wQL{[}0.5{]} &
  wQL{[}0.7{]} (I) &
  wQL{[}0.9{]} &
  wQL{[}0.995{]} (E) &
  MSIS{[}0.1{]} &
  MSIS{[}0.02{]} \\
  \hline
&
Gaussian &
0.053 ± 0.006 &
\textbf{0.0} ±0.0 &
0.082 ±0.008 &
0.076 ±0.008 &
0.047 ±0.006 &
0.010 ±0.003 &
\textbf{5.685} ±0.430 &
\textbf{10.254} ±1.198
\\ \cline{2-10} 
&
SQF &
0.054 ±0.008 &
\textbf{0.0} ±0.0 &
0.084 ±0.015 &
0.078 ±0.011 &
0.046 ±0.009 &
\textbf{0.008} ±0.004 &
6.424 ±1.488 &
12.485 ±3.211
\\ \cline{2-10} 
&
IQF &
\textbf{0.048} ±0.005 &
\textbf{0.0} ±0.0 &
\textbf{0.073} ±0.006 &
\textbf{0.067} ±0.007 &
\textbf{0.037} ±0.004 &
0.015 ±0.006 &
6.177 ±0.607 &
15.887 ±2.533
\\ \cline{2-10} 
\multirow{-4}{*}{ \texttt{Elec}}  &
ISQF &
0.055 ±0.010 &
\textbf{0.0} ±0.0 &
0.081 ±0.013 &
0.079 ±0.016 &
0.049 ±0.013 &
0.012 ±0.002 &
6.299 ±0.656 &
14.413 ±1.617
\\ 
\hline  \hline 
&
Gaussian &
0.126 ±0.004 &
\textbf{0.0} ±0.0 &
0.176 ±0.004 &
0.179 ±0.008 &
0.123 ±0.005 &
\textbf{0.025} ±0.000 &
5.599 ±0.244 &
\textbf{10.870} ±0.247
\\ \cline{2-10} 
&
SQF &
\textbf{0.111} ±0.002 &
\textbf{0.0} ±0.0 &
0.157 ±0.004 &
0.152 ±0.003 &
0.105 ±0.001 &
0.032 ±0.002 &
\textbf{4.889} ±0.071 &
12.020 ±0.353
\\ \cline{2-10} 
&
IQF &
0.119 ±0.014 &
\textbf{0.0} ±0.0 &
0.171 ±0.026 &
0.164 ±0.022 &
0.107 ±0.008 &
0.034 ±0.002 &
5.174 ±0.443 &
12.569 ±0.458
\\ \cline{2-10} 
\multirow{-4}{*}{ \texttt{Traf}} &
ISQF &
\textbf{0.111} ±0.003 &
\textbf{0.0} ±0.0 &
\textbf{0.154} ±0.005 &
\textbf{0.150} ±0.005 &
\textbf{0.103} ±0.003 &
0.035 ±0.002 &
5.059 ±0.225 &
12.608 ±0.713
\\ \hline
\end{tabular}
}
\caption{
Comparison of the accuracy for using DeepAR with Gaussian, SQF and our ISQF with 5 training quantile knots [0.01, 0.1, 0.5, 0.9, 0.99]. The mean and standard deviation are computed over 4 runs, and the winning method is shown in bold. (I) and (E) indicate the quantiles, where interpolation and extrapolation are performed, respectively.  The MSIS[0.1] measures the 90\% prediction interval using the interpolated $95^{\text{th}}$ and $5^{\text{th}}$ quantiles, if defined. Similarly, the MSIS[0.02] measures the 98\% prediction interval using the $99^{\text{th}}$ and $1^{\text{st}}$ training quantiles. }
\label{tab:results_deepar}
\end{table}

%% file: accuracy_table_appendix.tex
\begin{table}[H]
\centering
\scalebox{0.65}{
\begin{tabular}{
|c|c|c|c|c|c|c|c|c|c|c
 }
\hline
Dataset &
  Strategy &
  mean\_wQL &
  crossing \% &
  wQL{[}0.5{]} &
  wQL{[}0.7{]} &
  wQL{[}0.9{]} &
  wQL{[}0.995{]} &
  MSIS[0.1] &
  MSIS[0.02] \\
  \hline
&
Gaussian &
0.129 ±0.028 &
\textbf{0.0} ±0.0 &
0.234 ±0.052 &
0.224 ±0.042 &
0.159 ±0.042 &
0.068 ±0.038 &
42.8 ±12.0 &
79.8 ±22.9
\\ \cline{2-10} 
&
SQF &
0.379 ±0.182 &
\textbf{0.0} ±0.0 &
0.506 ±0.203 &
0.556 ±0.248 &
0.542 ±0.283 &
0.457 ±0.273 &
157.8 ±84.1 &
369.1 ±204.1
\\ \cline{2-10} 
&
QF 3 &
0.078 ±0.004 &
0.009 ±0.001 &
0.117 ±0.007 &
N/A &
0.055 ±0.001 &
N/A &
N/A &
20.4 ±1.1
\\ \cline{2-10} 
&
QF 5 &
\textbf{0.047} ±0.001 &
0.013 ±0.002 &
\textbf{0.104} ±0.004 &
N/A &
\textbf{0.054} ±0.001 &
N/A &
N/A &
\textbf{17.3} ±0.2
\\ \cline{2-10} 
&
IQF 3 &
0.084 ±0.007 &
\textbf{0.0} ±0.0 &
0.114 ±0.007 &
0.101 ±0.007 &
0.064 ±0.01 &
0.026 ±0.017 &
15.5 ±3.1 &
26.9 ±9.3
\\ \cline{2-10} 
&
IQF 5 &
0.055 ±0.006 &
\textbf{0.0} ±0.0 &
0.12 ±0.012 &
\textbf{0.101} ±0.008 &
0.056 ±0.004 &
\textbf{0.009} ±0.0 &
\textbf{12.2} ±0.3 &
17.6 ±0.5
\\ \cline{2-10} 
&
ISQF 3 &
0.305 ±0.083 &
\textbf{0.0} ±0.0 &
0.421 ±0.099 &
0.412 ±0.125 &
0.305 ±0.159 &
0.032 ±0.014 &
47.4 ±20.1 &
62.0 ±31.6
\\ \cline{2-10} 
\multirow{-8}{*}{ \texttt{Elec}} &
ISQF 5 &
0.198 ±0.063 &
\textbf{0.0} ±0.0 &
0.378 ±0.057 &
0.37 ±0.085 &
0.253 ±0.113 &
0.069 ±0.051 &
66.9 ±34.8 &
109.1 ±71.3
\\ \hline  \hline 
&
Gaussian &
\textbf{0.154} ±0.01 &
\textbf{0.0} ±0.0 &
\textbf{0.299} ±0.022 &
\textbf{0.3} ±0.024 &
\textbf{0.205} ±0.027 &
0.075 ±0.026 &
\textbf{14.5} ±1.3 &
\textbf{25.1} ±4.0
\\ \cline{2-10} 
&
SQF &
0.193 ±0.008 &
\textbf{0.0} ±0.0 &
0.331 ±0.02 &
0.339 ±0.024 &
0.274 ±0.049 &
0.152 ±0.052 &
22.2 ±2.1 &
45.4 ±4.2
\\ \cline{2-10} 
&
QF 3 &
1.814 ±1.703 &
0.499 ±0.306 &
1.278 ±0.479 &
N/A &
3.641 ±4.802 &
N/A &
N/A &
2572.1 ±2716.9
\\ \cline{2-10} 
&
QF 5 &
1.483 ±1.439 &
0.422 ±0.054 &
1.257 ±1.073 &
N/A &
1.035 ±0.763 &
N/A &
N/A &
1128.7 ±1569.1
\\ \cline{2-10} 
&
IQF 3 &
2.285 ±2.278 &
\textbf{0.0} ±0.0 &
3.18 ±3.041 &
3.157 ±3.458 &
2.761 ±3.289 &
0.75 ±0.792 &
188.3 ±145.7 &
282.8 ±184.4
\\ \cline{2-10} 
&
IQF 5 &
1.31 ±0.738 &
\textbf{0.0} ±0.0 &
2.046 ±1.11 &
1.705 ±1.23 &
1.132 ±1.071 &
0.917 ±1.355 &
166.5 ±150.9 &
373.6 ±354.0
\\ \cline{2-10} 
&
ISQF 3 &
0.269 ±0.03 &
\textbf{0.0} ±0.0 &
0.366 ±0.035 &
0.37 ±0.07 &
0.27 ±0.077 &
0.055 ±0.018 &
16.4 ±1.8 &
25.2 ±3.5
\\ \cline{2-10} 
\multirow{-8}{*}{ \texttt{Traf}} &
ISQF 5 &
0.187 ±0.057 &
\textbf{0.0} ±0.0 &
0.355 ±0.078 &
0.377 ±0.116 &
0.295 ±0.121 &
\textbf{0.043} ±0.008 &
18.8 ±7.5 &
29.4 ±18.1
\\ \hline  \hline 
&
Gaussian &
0.36 ±0.196 &
\textbf{0.0} ±0.0 &
0.575 ±0.233 &
0.634 ±0.327 &
0.58 ±0.38 &
0.38 ±0.325 &
52.2 ±35.3 &
109.1 ±81.8
\\ \cline{2-10} 
&
SQF &
0.176 ±0.013 &
\textbf{0.0} ±0.0 &
0.323 ±0.022 &
0.355 ±0.039 &
0.291 ±0.044 &
0.122 ±0.014 &
21.6 ±2.2 &
40.0 ±3.0
\\ \cline{2-10} 
&
QF 3 &
0.183 ±0.001 &
0.001 ±0.001 &
0.243 ±0.002 &
N/A &
0.213 ±0.001 &
N/A &
N/A &
35.7 ±0.3
\\ \cline{2-10} 
&
QF 5 &
0.136 ±0.002 &
0.001 ±0.0 &
0.247 ±0.006 &
N/A &
0.214 ±0.001 &
N/A &
N/A &
\textbf{34.5} ±0.3
\\ \cline{2-10} 
&
IQF 3 &
0.184 ±0.004 &
\textbf{0.0} ±0.0 &
0.244 ±0.003 &
0.277 ±0.004 &
0.213 ±0.001 &
0.103 ±0.001 &
19.1 ±1.4 &
38.2 ±3.7
\\ \cline{2-10} 
&
IQF 5 &
0.135 ±0.001 &
\textbf{0.0} ±0.0 &
0.246 ±0.004 &
0.278 ±0.004 &
0.213 ±0.001 &
0.087 ±0.0 &
\textbf{19.0} ±0.3 &
34.7 ±0.3
\\ \cline{2-10} 
&
ISQF 3 &
{0.052} ±0.09 &
\textbf{0.0} ±0.0 &
\textbf{0.071} ±0.123 &
\textbf{0.076} ±0.131 &
\textbf{0.063} ±0.109 &
{0.027} ±0.047 &
19.2 ±0.0 &
37.0 ±0.0
\\ \cline{2-10} 
\multirow{-8}{*}{ \texttt{Wiki}} &
ISQF 5 &
\textbf{0.051} ±0.089 &
\textbf{0.0} ±0.0 &
0.098 ±0.171 &
0.087 ±0.151 &
0.064 ±0.112 &
\textbf{0.026} ±0.045 &
21.2 ±0.0 &
41.9 ±0.0
\\ \hline  \hline 
&
Gaussian &
\textbf{0.027} ±0.005 &
\textbf{0.0} ±0.0 &
0.065 ±0.011 &
\textbf{0.057} ±0.008 &
\textbf{0.031} ±0.003 &
\textbf{0.003} ±0.0 &
\textbf{33.6} ±11.4 &
\textbf{41.0} ±14.7
\\ \cline{2-10} 
&
SQF &
0.039 ±0.007 &
\textbf{0.0} ±0.0 &
0.077 ±0.009 &
0.069 ±0.008 &
0.046 ±0.01 &
0.007 ±0.001 &
66.7 ±8.1 &
80.5 ±8.0
\\ \cline{2-10} 
&
QF 3 &
0.041 ±0.003 &
0.038 ±0.024 &
0.061 ±0.004 &
N/A &
0.039 ±0.003 &
N/A &
N/A &
146.0 ±9.6
\\ \cline{2-10} 
&
QF 5 &
0.03 ±0.003 &
0.038 ±0.007 &
\textbf{0.051} ±0.008 &
N/A &
0.037 ±0.005 &
N/A &
N/A &
111.5 ±5.4
\\ \cline{2-10} 
&
IQF 3 &
0.047 ±0.005 &
\textbf{0.0} ±0.0 &
0.057 ±0.01 &
0.07 ±0.006 &
0.039 ±0.001 &
0.007 ±0.0 &
142.0 ±85.1 &
267.1 ±233.2
\\ \cline{2-10} 
&
IQF 5 &
0.031 ±0.005 &
\textbf{0.0} ±0.0 &
0.052 ±0.006 &
0.07 ±0.008 &
0.041 ±0.005 &
0.007 ±0.001 &
89.7 ±30.0 &
146.6 ±85.0
\\ \cline{2-10} 
&
ISQF 3 &
0.061 ±0.004 &
\textbf{0.0} ±0.0 &
0.081 ±0.01 &
0.071 ±0.009 &
0.044 ±0.006 &
0.012 ±0.002 &
124.7 ±25.2 &
162.6 ±25.0
\\ \cline{2-10} 
\multirow{-8}{*}{ \texttt{M4-hourly}} &
ISQF 5 &
0.047 ±0.006 &
\textbf{0.0} ±0.0 &
0.09 ±0.002 &
0.076 ±0.008 &
0.047 ±0.008 &
0.015 ±0.006 &
63.5 ±6.9 &
160.8 ±32.1
\\ \hline  \hline 
&
Gaussian &
0.017 ±0.002 &
\textbf{0.0} ±0.0 &
0.034 ±0.001 &
0.029 ±0.0 &
0.017 ±0.001 &
0.005 ±0.001 &
43.8 ±6.6 &
74.0 ±15.3
\\ \cline{2-10} 
&
SQF &
0.017 ±0.001 &
\textbf{0.0} ±0.0 &
0.031 ±0.001 &
0.028 ±0.001 &
0.017 ±0.001 &
0.009 ±0.002 &
48.9 ±5.7 &
95.2 ±17.0
\\ \cline{2-10} 
&
QF 3 &
0.023 ±0.0 &
0.001 ±0.001 &
0.035 ±0.002 &
N/A &
0.014 ±0.001 &
N/A &
N/A &
53.0 ±2.2
\\ \cline{2-10} 
&
QF 5 &
0.015 ±0.001 &
0.003 ±0.003 &
0.029 ±0.003 &
N/A &
\textbf{0.013} ±0.001 &
N/A &
N/A &
\textbf{52.2} ±1.6
\\ \cline{2-10} 
&
IQF 3 &
0.023 ±0.002 &
\textbf{0.0} ±0.0 &
0.031 ±0.001 &
0.027 ±0.003 &
0.019 ±0.006 &
0.01 ±0.008 &
50.5 ±16.5 &
99.5 ±44.0
\\ \cline{2-10} 
&
IQF 5 &
0.017 ±0.001 &
\textbf{0.0} ±0.0 &
0.033 ±0.002 &
0.026 ±0.001 &
0.017 ±0.002 &
0.006 ±0.003 &
54.3 ±17.7 &
93.1 ±34.8
\\ \cline{2-10} 
&
ISQF 3 &
0.026 ±0.003 &
\textbf{0.0} ±0.0 &
{0.037} ±0.005 &
{0.032} ±0.004 &
{0.019} ±0.003 &
{0.004} ±0.0 &
{42.9} ±3.4 &
{65.4} ±3.5
\\ \cline{2-10} 
\multirow{-8}{*}{ \texttt{M4-daily}} &
ISQF 5 &
\textbf{0.014} ±0.008 &
\textbf{0.0} ±0.0 &
\textbf{0.025} ±0.014 &
\textbf{0.023} ±0.013 &
\textbf{0.013} ±0.008 &
\textbf{0.003} ±0.002 &
\textbf{42.5} ±0.9 &
154.1 ±59.0
\\ \hline  \hline 
&
Gaussian &
0.043 ±0.002 &
\textbf{0.0} ±0.0 &
0.086 ±0.004 &
0.087 ±0.005 &
0.059 ±0.002 &
0.017 ±0.001 &
\textbf{60.9} ±6.8 &
81.3 ±9.0
\\ \cline{2-10} 
&
SQF &
0.046 ±0.003 &
\textbf{0.0} ±0.0 &
0.085 ±0.002 &
0.084 ±0.002 &
0.063 ±0.002 &
0.022 ±0.003 &
64.6 ±4.7 &
96.8 ±18.0
\\ \cline{2-10} 
&
QF 3 &
0.055 ±0.002 &
0.001 ±0.001 &
0.071 ±0.004 &
N/A &
0.058 ±0.003 &
N/A &
N/A &
\textbf{73.7} ±1.3
\\ \cline{2-10} 
&
QF 5 &
\textbf{0.038} ±0.0 &
0.0 ±0.0 &
\textbf{0.067} ±0.002 &
N/A &
\textbf{0.057} ±0.001 &
N/A &
N/A &
86.8 ±2.4
\\ \cline{2-10} 
&
IQF 3 &
0.056 ±0.003 &
\textbf{0.0} ±0.0 &
0.069 ±0.004 &
0.072 ±0.003 &
0.058 ±0.002 &
0.024 ±0.004 &
64.9 ±6.6 &
102.9 ±17.3
\\ \cline{2-10} 
&
IQF 5 &
0.042 ±0.001 &
\textbf{0.0} ±0.0 &
0.069 ±0.004 &
\textbf{0.072} ±0.003 &
0.058 ±0.004 &
0.019 ±0.01 &
70.1 ±4.3 &
103.7 ±8.3
\\ \cline{2-10} 
&
ISQF 3 &
0.068 ±0.003 &
\textbf{0.0} ±0.0 &
0.088 ±0.003 &
0.087 ±0.003 &
0.067 ±0.004 &
0.014 ±0.003 &
90.8 ±20.5 &
131.1 ±36.1
\\ \cline{2-10} 
\multirow{-8}{*}{ \texttt{M4-weekly}} &
ISQF 5 &
0.047 ±0.001 &
\textbf{0.0} ±0.0 &
0.084 ±0.004 &
0.084 ±0.003 &
0.063 ±0.002 &
\textbf{0.012} ±0.0 &
65.8 ±5.3 &
157.6 ±31.2
\\ \hline  \hline 
&
Gaussian &
0.149 ±0.01 &
\textbf{0.0} ±0.0 &
0.188 ±0.01 &
0.179 ±0.01 &
0.153 ±0.01 &
0.115 ±0.009 &
37.0 ±3.0 &
80.4 ±7.2
\\ \cline{2-10} 
&
SQF &
0.12 ±0.009 &
\textbf{0.0} ±0.0 &
0.179 ±0.008 &
0.168 ±0.009 &
0.128 ±0.009 &
0.079 ±0.007 &
25.3 ±2.0 &
50.8 ±5.8
\\ \cline{2-10} 
&
QF 3 &
0.111 ±0.002 &
0.006 ±0.003 &
0.131 ±0.003 &
N/A &
0.093 ±0.003 &
N/A &
N/A &
37.4 ±2.1
\\ \cline{2-10} 
&
QF 5 &
0.087 ±0.001 &
0.002 ±0.0 &
0.131 ±0.001 &
N/A &
0.09 ±0.001 &
N/A &
N/A &
\textbf{31.5} ±0.7
\\ \cline{2-10} 
&
IQF 3 &
0.109 ±0.003 &
\textbf{0.0} ±0.0 &
0.132 ±0.002 &
0.123 ±0.001 &
0.095 ±0.004 &
0.052 ±0.007 &
20.0 ±1.5 &
37.2 ±4.6
\\ \cline{2-10} 
&
IQF 5 &
0.081 ±0.004 &
\textbf{0.0} ±0.0 &
0.131 ±0.001 &
0.121 ±0.001 &
0.093 ±0.001 &
0.038 ±0.023 &
\textbf{19.3} ±1.2 &
33.4 ±5.4
\\ \cline{2-10} 
&
ISQF 3 &
\textbf{0.081} ±0.082 &
\textbf{0.0} ±0.0 &
\textbf{0.102} ±0.103 &
\textbf{0.094} ±0.095 &
\textbf{0.067} ±0.067 &
\textbf{0.018} ±0.018 &
48.2 ±25.6 &
70.3 ±38.4
\\ \cline{2-10} 
\multirow{-8}{*}{ \texttt{M4-monthly}} &
ISQF 5 &
0.114 ±0.002 &
\textbf{0.0} ±0.0 &
0.181 ±0.001 &
0.174 ±0.0 &
0.141 ±0.009 &
0.042 ±0.003 &
28.7 ±0.9 &
52.6 ±1.8
\\ \hline  \hline 
&
Gaussian &
0.054 ±0.002 &
\textbf{0.0} ±0.0 &
0.12 ±0.006 &
0.108 ±0.005 &
0.064 ±0.003 &
0.012 ±0.003 &
\textbf{12.9} ±0.1 &
\textbf{16.7} ±1.0
\\ \cline{2-10} 
&
SQF &
0.065 ±0.002 &
\textbf{0.0} ±0.0 &
0.117 ±0.005 &
0.111 ±0.004 &
0.076 ±0.001 &
0.04 ±0.004 &
18.2 ±0.5 &
31.8 ±1.7
\\ \cline{2-10} 
&
QF 3 &
0.076 ±0.002 &
0.0 ±0.0 &
0.108 ±0.001 &
N/A &
0.063 ±0.002 &
N/A &
N/A &
19.1 ±1.9
\\ \cline{2-10} 
&
QF 5 &
0.052 ±0.001 &
0.003 ±0.002 &
0.112 ±0.004 &
N/A &
0.065 ±0.001 &
N/A &
N/A &
20.3 ±2.8
\\ \cline{2-10} 
&
IQF 3 &
0.077 ±0.004 &
\textbf{0.0} ±0.0 &
0.104 ±0.003 &
0.096 ±0.002 &
0.071 ±0.01 &
0.033 ±0.021 &
18.1 ±5.3 &
31.6 ±14.5
\\ \cline{2-10} 
&
IQF 5 &
0.054 ±0.002 &
\textbf{0.0} ±0.0 &
0.106 ±0.002 &
0.095 ±0.002 &
0.068 ±0.006 &
0.013 ±0.001 &
18.8 ±1.8 &
27.5 ±2.8
\\ \cline{2-10} 
&
ISQF 3 &
\textbf{0.045} ±0.045 &
\textbf{0.0} ±0.0 &
\textbf{0.061} ±0.061 &
\textbf{0.057} ±0.057 &
\textbf{0.038} ±0.038 &
\textbf{0.008} ±0.008 &
19.7 ±0.1 &
26.6 ±0.5
\\ \cline{2-10} 
\multirow{-8}{*}{ \texttt{M4-quarterly}} &
ISQF 5 &
0.06 ±0.001 &
\textbf{0.0} ±0.0 &
0.115 ±0.003 &
0.108 ±0.003 &
0.073 ±0.002 &
0.015 ±0.001 &
17.8 ±0.3 &
35.5 ±6.0
\\ \hline  \hline 
&
Gaussian &
0.085 ±0.007 &
\textbf{0.0} ±0.0 &
0.17 ±0.013 &
0.153 ±0.009 &
0.095 ±0.004 &
0.033 ±0.001 &
42.2 ±5.3 &
67.6 ±7.6
\\ \cline{2-10} 
&
SQF &
0.079 ±0.002 &
\textbf{0.0} ±0.0 &
0.142 ±0.001 &
0.135 ±0.002 &
0.092 ±0.002 &
0.045 ±0.006 &
40.4 ±1.7 &
76.4 ±6.0
\\ \cline{2-10} 
&
QF 3 &
0.1 ±0.001 &
0.001 ±0.001 &
0.134 ±0.001 &
N/A &
0.089 ±0.001 &
N/A &
N/A &
59.2 ±2.6
\\ \cline{2-10} 
&
QF 5 &
0.07 ±0.001 &
0.001 ±0.001 &
0.134 ±0.0 &
N/A &
0.09 ±0.001 &
N/A &
N/A &
\textbf{50.4} ±0.5
\\ \cline{2-10} 
&
IQF 3 &
0.098 ±0.002 &
\textbf{0.0} ±0.0 &
0.134 ±0.002 &
0.13 ±0.001 &
0.09 ±0.004 &
0.034 ±0.01 &
\textbf{36.3} ±3.6 &
58.6 ±10.4
\\ \cline{2-10} 
&
IQF 5 &
\textbf{0.069} ±0.002 &
\textbf{0.0} ±0.0 &
0.135 ±0.001 &
0.13 ±0.003 &
0.09 ±0.004 &
0.026 ±0.013 &
37.7 ±1.2 &
59.7 ±8.8
\\ \cline{2-10} 
&
ISQF 3 &
0.085 ±0.062 &
\textbf{0.0} ±0.0 &
\textbf{0.131} ±0.1 &
\textbf{0.114} ±0.086 &
\textbf{0.072} ±0.052 &
\textbf{0.015} ±0.011 &
42.0 ±1.4 &
61.0 ±1.2
\\ \cline{2-10} 
\multirow{-8}{*}{ \texttt{M4-yearly}} &
ISQF 5 &
0.075 ±0.003 &
\textbf{0.0} ±0.0 &
0.148 ±0.006 &
0.136 ±0.004 &
0.091 ±0.002 &
0.023 ±0.002 &
37.4 ±2.0 &
66.8 ±3.8
\\ \hline 
\end{tabular}
}
\caption{
Comparison of the accuracy for our I(S)QF and various baselines with 5 training quantile knots [0.01, 0.1, 0.5, 0.9, 0.99] and 3 training quantile knots [0.1, 0.5, 0.9]. The mean and standard deviation are computed over 4 runs, and the winning method is shown in bold. (I) and (E) indicate the quantiles, where interpolation and extrapolation are performed, respectively.  The MSIS[0.1] measures the 90\% prediction interval using the interpolated $95^{\text{th}}$ and $5^{\text{th}}$ quantiles, if defined. Similarly, the MSIS[0.02] measures the 98\% prediction interval using the $99^{\text{th}}$ and $1^{\text{st}}$ training quantiles.
}
\label{tab:all_benchmark_results}
\end{table}

%% file: main.bbl
\begin{thebibliography}{}

\bibitem[Alexandrov et~al., 2020]{gluonts_jmlr}
Alexandrov, A., Benidis, K., Bohlke-Schneider, M., Flunkert, V., Gasthaus, J.,
  Januschowski, T., Maddix, D.~C., Rangapuram, S., Salinas, D., Schulz, J.,
  Stella, L., Türkmen, A.~C., and Wang, Y. (2020).
\newblock {GluonTS: Probabilistic and Neural Time Series Modeling in Python}.
\newblock {\em Journal of Machine Learning Research}, 21(116):1--6.

\bibitem[Chen et~al., 2020]{chen2019probabilistic}
Chen, Y., Kang, Y., Chen, Y., and Wang, Z. (2020).
\newblock Probabilistic forecasting with temporal convolutional neural network.
\newblock {\em Neurocomputing}, 399:491--501.

\bibitem[Chernozhukov et~al., 2010]{chernozhukov2010quantile}
Chernozhukov, V., Fern{\'a}ndez-Val, I., and Galichon, A. (2010).
\newblock Quantile and probability curves without crossing.
\newblock {\em Econometrica}, 78(3):1093--1125.

\bibitem[de~B{\'e}zenac et~al., 2020]{de2020normalizing}
de~B{\'e}zenac, E., Rangapuram, S.~S., Benidis, K., Bohlke-Schneider, M.,
  Kurle, R., Stella, L., Hasson, H., Gallinari, P., and Januschowski, T.
  (2020).
\newblock Normalizing kalman filters for multivariate time series analysis.
\newblock In {\em NeurIPS}.

\bibitem[Dheeru and Karra~Taniskidou, 2017]{Dua:2017}
Dheeru, D. and Karra~Taniskidou, E. (2017).
\newblock {UCI} machine learning repository.

\bibitem[Ehrlich et~al., 2021]{ehrlich2021spliced}
Ehrlich, E., Callot, L., and Aubet, F.-X. (2021).
\newblock Spliced binned-pareto distribution for robust modeling of
  heavy-tailed time series.
\newblock {\em arXiv preprint arXiv:2106.10952}.

\bibitem[Eisenach et~al., 2020]{eisenach2020mqtransformer}
Eisenach, C., Patel, Y., and Madeka, D. (2020).
\newblock Mqtransformer: Multi-horizon forecasts with context dependent and
  feedback-aware attention.
\newblock {\em arXiv preprint arXiv:2009.14799}.

\bibitem[Flunkert et~al., 2017]{flunkert2017deepar}
Flunkert, V., Salinas, D., and Gasthaus, J. (2017).
\newblock Deepar: Probabilistic forecasting with autoregressive recurrent
  networks.
\newblock {\em arXiv preprint arXiv:1704.04110}.

\bibitem[Friederichs and Thorarinsdottir, 2012]{friederichs2012forecast}
Friederichs, P. and Thorarinsdottir, T.~L. (2012).
\newblock Forecast verification for extreme value distributions with an
  application to probabilistic peak wind prediction.
\newblock {\em Environmetrics}, 23(7):579--594.

\bibitem[Gasthaus et~al., 2019]{gasthaus19sqf}
Gasthaus, J., Benidis, K., Wang, Y., Rangapuram, S.~S., Salinas, D., Flunkert,
  V., and Januschowski, T. (2019).
\newblock Probabilistic forecasting with spline quantile function rnns.
\newblock In Chaudhuri, K. and Sugiyama, M., editors, {\em Proceedings of the
  Twenty-Second International Conference on Artificial Intelligence and
  Statistics}, volume~89 of {\em Proceedings of Machine Learning Research},
  pages 1901--1910. PMLR.

\bibitem[Gneiting and Raftery, 2007]{gneiting2007strictly}
Gneiting, T. and Raftery, A.~E. (2007).
\newblock Strictly proper scoring rules, prediction, and estimation.
\newblock {\em Journal of the American Statistical Association},
  102(477):359--378.

\bibitem[Hyndman et~al., 2008]{hyndman2008forecasting}
Hyndman, R., Koehler, A.~B., Ord, J.~K., and Snyder, R.~D. (2008).
\newblock {\em Forecasting with exponential smoothing: the state space
  approach}.
\newblock Springer Science \& Business Media.

\bibitem[Jin et~al., 2022]{jin2022domain}
Jin, X., Park, Y., Maddix, D.~C., Wang, H., and Wang, Y. (2022).
\newblock Domain adaptation for time series forecasting via attention sharing.

\bibitem[Kim et~al., 2020]{kim2020optimal}
Kim, J., Park, Y., Fox, J.~D., Boyd, S.~P., and Dally, W. (2020).
\newblock Optimal operation of a plug-in hybrid vehicle with battery thermal
  and degradation model.
\newblock In {\em 2020 American Control Conference (ACC)}, pages 3083--3090.
  IEEE.

\bibitem[Kim et~al., 2021]{kim2021deep}
Kim, T., Fakoor, R., Mueller, J., Smola, A.~J., and Tibshirani, R.~J. (2021).
\newblock Deep quantile aggregation.
\newblock {\em arXiv preprint arXiv:2103.00083}.

\bibitem[Koenker, 2005]{koenker_2005}
Koenker, R. (2005).
\newblock {\em Quantile Regression}.
\newblock Econometric Society Monographs. Cambridge University Press.

\bibitem[Koenker and Bassett~Jr, 1978]{Koenker1978regression}
Koenker, R. and Bassett~Jr, G. (1978).
\newblock Regression quantiles.
\newblock {\em Econometrica: Journal of the Econometric Society}, pages 33--50.

\bibitem[Kuznetsov and Mohri, 2015]{kuznetsov2015learning}
Kuznetsov, V. and Mohri, M. (2015).
\newblock Learning theory and algorithms for forecasting non-stationary time
  series.
\newblock In {\em NIPS}, pages 541--549. Citeseer.

\bibitem[Lai, 2017]{lai_dataset_2017}
Lai (2017).
\newblock Dataset of {Kaggle} {Competition} {Web} {Traffic} {Time} {Series}
  {Forecasting}, {Version} 3.

\bibitem[Liberty et~al., 2020]{liberty2020elastic}
Liberty, E., Karnin, Z., Xiang, B., Rouesnel, L., Coskun, B., Nallapati, R.,
  Delgado, J., Sadoughi, A., Astashonok, Y., Das, P., et~al. (2020).
\newblock Elastic machine learning algorithms in amazon sagemaker.
\newblock In {\em Proceedings of the 2020 ACM SIGMOD International Conference
  on Management of Data}, pages 731--737.

\bibitem[Lim et~al., 2019]{lim2019temporal}
Lim, B., Arik, S.~O., Loeff, N., and Pfister, T. (2019).
\newblock Temporal fusion transformers for interpretable multi-horizon time
  series forecasting.
\newblock {\em arXiv preprint arXiv:1912.09363}.

\bibitem[Liu and Wu, 2009]{liu2009stepwise}
Liu, Y. and Wu, Y. (2009).
\newblock Stepwise multiple quantile regression estimation using non-crossing
  constraints.
\newblock {\em Statistics and its Interface}, 2(3):299--310.

\bibitem[Lu et~al., 2021]{pmlr-v139-lu21d}
Lu, Y., Park, Y., Chen, L., Wang, Y., De~Sa, C., and Foster, D. (2021).
\newblock Variance reduced training with stratified sampling for forecasting
  models.
\newblock In Meila, M. and Zhang, T., editors, {\em Proceedings of the 38th
  International Conference on Machine Learning}, volume 139 of {\em Proceedings
  of Machine Learning Research}, pages 7145--7155. PMLR.

\bibitem[Makridakis et~al., 2018]{makridakisM4concl}
Makridakis, S. et~al. (2018).
\newblock The {M4} competition: Results, findings, conclusion and way forward.
\newblock {\em International Journal of Forecasting}, 34(4):802 -- 808.

\bibitem[Mariet and Kuznetsov, 2019]{mariet2019foundations}
Mariet, Z. and Kuznetsov, V. (2019).
\newblock Foundations of sequence-to-sequence modeling for time series.
\newblock In {\em The 22nd International Conference on Artificial Intelligence
  and Statistics}, pages 408--417. PMLR.

\bibitem[Park et~al., 2019]{park2019linear}
Park, Y., Mahadik, K., Rossi, R.~A., Wu, G., and Zhao, H. (2019).
\newblock Linear quadratic regulator for resource-efficient cloud services.
\newblock In {\em Proceedings of the ACM Symposium on Cloud Computing}, pages
  488--489.

\bibitem[Park et~al., 2020]{park2020structured}
Park, Y., Rossi, R., Wen, Z., Wu, G., and Zhao, H. (2020).
\newblock Structured policy iteration for linear quadratic regulator.
\newblock In {\em International Conference on Machine Learning}, pages
  7521--7531. PMLR.

\bibitem[Petropoulos et~al., 2022]{petropoulos2022forecasting}
Petropoulos, F., Apiletti, D., Assimakopoulos, V., Babai, M.~Z., Barrow, D.~K.,
  Taieb, S.~B., Bergmeir, C., Bessa, R.~J., Bijak, J., Boylan, J.~E., et~al.
  (2022).
\newblock Forecasting: theory and practice.
\newblock {\em International Journal of Forecasting}.

\bibitem[Rangapuram et~al., 2018]{rangapuram2018deep}
Rangapuram, S.~S., Seeger, M.~W., Gasthaus, J., Stella, L., Wang, Y., and
  Januschowski, T. (2018).
\newblock Deep state space models for time series forecasting.
\newblock {\em Advances in neural information processing systems},
  31:7785--7794.

\bibitem[Rasmussen and Williams, 2006]{rasmussen2006}
Rasmussen, C. and Williams, C. (2006).
\newblock {\em Gaussian Processes for Machine Learning}.
\newblock MIT Press.

\bibitem[Salinas et~al., 2019]{salinas2019high}
Salinas, D., Bohlke-Schneider, M., Callot, L., Medico, R., and Gasthaus, J.
  (2019).
\newblock High-dimensional multivariate forecasting with low-rank gaussian
  copula processes.
\newblock {\em Advances in Neural Information Processing Systems},
  32:6827--6837.

\bibitem[Salinas et~al., 2020]{salinas2020deepar}
Salinas, D., Flunkert, V., Gasthaus, J., and Januschowski, T. (2020).
\newblock Deepar: Probabilistic forecasting with autoregressive recurrent
  networks.
\newblock {\em International Journal of Forecasting}, 36(3):1181--1191.

\bibitem[Schmidt and Zhu, 2016]{schmidt2016quantile}
Schmidt, L. and Zhu, Y. (2016).
\newblock Quantile spacings: A simple method for the joint estimation of
  multiple quantiles without crossing.
\newblock {\em Available at SSRN 2220901}.

\bibitem[Wang et~al., 2019]{wang2019deep}
Wang, Y., Smola, A., Maddix, D., Gasthaus, J., Foster, D., and Januschowski, T.
  (2019).
\newblock Deep factors for forecasting.
\newblock In {\em International conference on machine learning}, pages
  6607--6617. PMLR.

\bibitem[Wen and Torkkola, 2019]{wen2019deep}
Wen, R. and Torkkola, K. (2019).
\newblock Deep generative quantile-copula models for probabilistic forecasting.
\newblock {\em arXiv preprint arXiv:1907.10697}.

\bibitem[Wen et~al., 2017]{wen2017multi}
Wen, R., Torkkola, K., and Narayanaswamy, B. (2017).
\newblock A multi-horizon quantile recurrent forecaster.
\newblock {\em arXiv preprint arXiv:1711.11053}.

\bibitem[Zimin and Lampert, 2017]{zimin2017learning}
Zimin, A. and Lampert, C. (2017).
\newblock Learning theory for conditional risk minimization.
\newblock In {\em Artificial Intelligence and Statistics}, pages 213--222.
  PMLR.

\end{thebibliography}
